\documentclass[twoside,leqno,twocolumn]{article}

\usepackage{amsmath, amssymb, amsfonts}
\usepackage{graphicx}
\usepackage{makecell}
\usepackage{mathtools}
\usepackage{multirow}
\usepackage{arydshln}
\usepackage{cellspace}
\usepackage{lipsum}
\usepackage{easyeqn}
\usepackage{cite}
\usepackage{textcomp}
\usepackage{hyperref}
\usepackage{colortbl}
\usepackage[table, dvipsnames]{xcolor}
\usepackage{caption}
\usepackage{subcaption}
\usepackage{bbm}
\usepackage{booktabs}
\usepackage[letterpaper]{geometry}
\usepackage[numbers]{natbib}
\usepackage{ltexpprt}
\usepackage{enumitem}
\usepackage{xcolor}

\begin{document}
\newcommand{\email}[1]{\href{mailto:#1}{\texttt{#1}}}

\title{Learning Confident Classifiers in the Presence of Label Noise}

\author{
Asma Ahmed Hashmi$^{1}$%
\and Aigerim Zhumabayeva$^{2}$%
\and Nikita Kotelevskii$^{2,3}$%
\and Artem Agafonov$^{2}$%
\and Mohammad Yaqub$^{2}$%
\and Maxim Panov$^{2}$%
\and Martin Takáč$^{2}$%
}

\date{}

\maketitle

\footnotetext[1]{Institute of Informatics, LMU Munich (\email{asmah17@gmail.com})}
\footnotetext[2]{MBZUAI, UAE}
\footnotetext[3]{Skoltech, Russia}

\setlength{\marginparwidth }{2cm}

\newcommand{\dummy}[1]{{\color{blue}\lipsum[#1]}}

\newcommand{\mt}[1]{\todo[color=pink]{#1}}
\newcommand{\mti}[1]{\todo[inline,color=pink]{#1}}
\newcommand{\ah}[1]{\todo[color=green]{#1}}

\newcommand{\nk}[1]{\todo[inline,color=red]{#1}}

\newcommand{\R}{\mathbb{R}}
\newcommand{\Y}{\mathcal{Y}}

\newcommand{\inimage}{x_i}
\newcommand{\inimageni}{x} 

\newcommand{\ynoisy}{{\tilde{y}_i^r}}
\newcommand{\ynoisyni}{{\tilde{y}^r}} 
\newcommand{\ygt}{{y_i}}
\newcommand{\ygtni}{{y}} 

\newcommand{\annset}{{S(\inimage)}}
\newcommand{\annsetni}{{S(\inimageni)}} 

\newcommand{\segnetni}{\hat{p}_{\theta}(\inimageni)} 
\newcommand{\segnet}{\hat{p}_{\theta}(\inimage)}

\newcommand{\annetni}{\hat{T}^{r}_{\psi}(\inimageni)} 
\newcommand{\annet}{\hat{T}^{r}_{\psi}(\inimage)}

\definecolor{semitransparentred}{rgb}{1,0,0} 
\colorlet{transred}{semitransparentred!20} 
\newcommand{\indicator}[1]{\mathds{1}\left[#1\right]}

\newcommand{\eqdef}{:=}

\section*{Abstract}

The success of Deep Neural Network (DNN) models significantly depends on the quality of provided annotations. In medical image segmentation, for example, having multiple expert annotations for each data point is standard to minimize subjective annotation bias. Then, the goal of estimation is to filter out the label noise and recover the ground-truth masks, which are not explicitly given. This paper proposes a probabilistic model for noisy observations that allows us to build confident classification and segmentation models. We explicitly model label noise to accomplish this and introduce a new information-based regularization that pushes the network to recover the ground-truth labels. In addition, we adjust the loss function for the segmentation task by prioritizing learning in high-confidence regions where all the annotators agree on labeling. We evaluate the proposed method on a series of classification tasks such as noisy versions of MNIST, CIFAR-10, and Fashion-MNIST datasets, as well as CIFAR-10N, a real-world dataset with noisy human annotations. Additionally, for the segmentation task, we consider several medical imaging datasets, such as LIDC and RIGA, that reflect real-world inter-variability among multiple annotators. Our experiments show that our algorithm outperforms state-of-the-art solutions for the considered classification and segmentation problems.


\section{Introduction}
\label{sec:intro}
  Real-world data are replete with noisy labels. Since the labeling process of large-scale datasets is costly and time-consuming, researchers often resort to less expensive options, such as internet inquiries and crowdsourcing to circumvent this issue~\cite{Yan2014, Veit}. Unfortunately, these methods are viable in producing datasets with incorrect labels. Smaller datasets are also vulnerable to the presence of corrupted labels. In this case, usually, the labeling process is either challenging or the annotators have divergent opinions~\cite{Barkan, Ma}. In medical imaging, for example, it is imperative to procure annotations from clinical experts. However, it is not only expensive to obtain annotated data, but it also suffers from high inter-reader variability among domain's experts~\cite{Lazarus, Litjens2017}, particularly in tasks such as segmentation and classification of images~\cite{Lazarus, joskowicz2019inter}. Furthermore, the quality of the annotated labels became more intrinsic with the ubiquitous use of Deep Learning (DL) models in the medical domain~\cite{schilling, wu, caicedo}, as the decision of the trainable models might have consequences for the health or even the life of a patient. Hence, developing DL algorithms that are robust to the noise in the annotated data is inarguably important. 

  DNNs perform worse when there are noisy labels. Some algorithms have been created to help DNNs keep their performance even with noise. For example, Sample Selection methods that employ a Student-Teacher dual network approach have been developed to handle noisy labels, utilizing ``small loss criteria'' to identify clean instances for peer training, as discussed in works like~\cite{Co-teaching, coteachingplus, Decouple, JoCoR, MentorNet}. However, their effectiveness decreases when true and false-labeled examples' loss distributions significantly overlap.

  The differences in annotations by multiple raters lead to inter-rater variability, which reflects the level of consistency between different raters in measuring the same target~\cite{joskowicz2019inter,schaekermann2019understanding}. This variability arises due to differences in experience, subject-specific expertise, and interpretation among the annotators. Moreover, the annotation task could be complex, time-consuming, and likely to be taxing; this inevitably leads to noise in the annotations, which degrades the model's performance. To counter this, ground-truth annotations are typically obtained through Simultaneous Truth and Performance Level Estimation (STAPLE; \cite{Warfield}) majority voting technique from multiple raters. This is done by selecting one preferred expert, or by using other label fusion approaches~\cite{Kats2019soft, Songbai2005, Asman, Jorge, ZHAO2020107068, myronenko20183d}. However, this process loses important agreement/disagreement information from all experts. In addition, while this technique is simple and easy to adopt, it comes at the expense of ignoring the underlying uncertainty among various experts.

\begin{figure*}[t]
    \centering
    \includegraphics[width=0.7\linewidth]{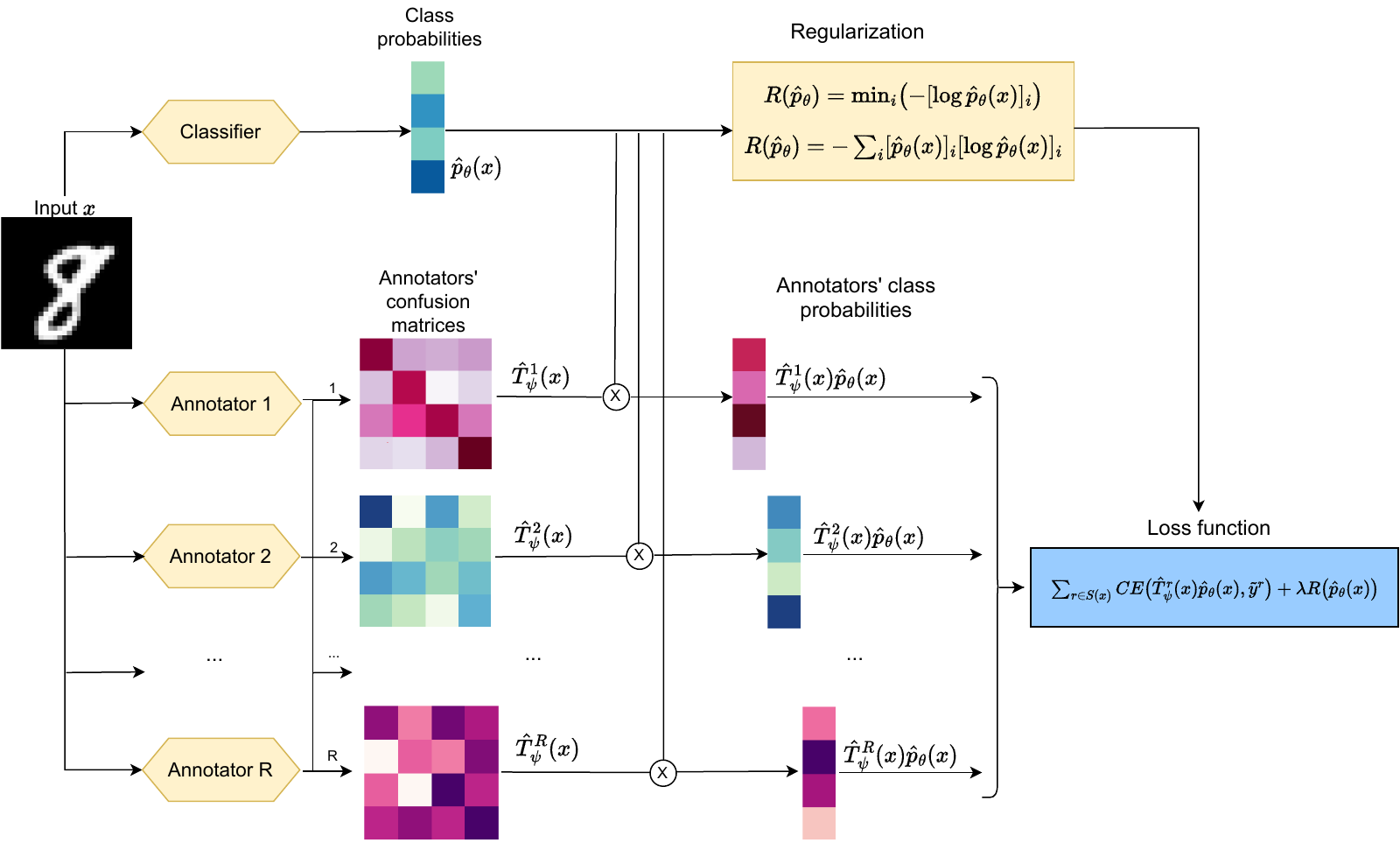}
    \caption{Illustration of the proposed architecture. The input image is fed into two separate networks -- Base (in the particular case of this example -- Classifier Network) and Annotation. Then, their predictions are multiplied to produce the prediction of the noisy (annotators) labels distribution. Finally, the loss function is computed, summing up all the components.}
    \label{fig:Our_NN}
  \end{figure*}

  In this work, we develop a new deep learning approach that achieves high-quality predictions in the presence of multiple annotators, that induce label noise in the dataset. This approach works both, for classification and for segmentation. Inspired by~\cite{Tanno2019,DisentanglingHumanError}, we consider coupling two models: a base classification network that estimates the true labels and an annotation network that estimates confusion matrices for each rater for a given input image. However, unlike other methods, we want to make a base network confident in predicting each class in the presence of the noisy labels alone. To accomplish it, we introduce the new regularization term that forces predictions to be close to the one-hot vector, with \(1\) being for the correct class and \(0\) for the rest; Figure~\ref{fig:Our_NN} gives an illustration of the details of the proposed architecture. This approach assumes low ambiguity in ground truth annotations and attributes existing label uncertainty to the annotator's noise. In addition, for segmentation tasks, we make use of the fact there are regions where all annotators agree and feed that information as input to the additional loss term that forces the confident predictions by the base segmentation network without the involvement of the annotation network; details of the segmentation network is depicted in Figure \ref{fig:Our_segmentation_NN}.

  \noindent We summarize our \textbf{contributions} as follows.
  \begin{enumerate}
    \item We introduce a class of ``confidence'' (information-based) regularizers, that allows us to jointly learn the parameters of both base classification and annotation networks. It explicitly forces the model to learn confident predictions, thus allowing it to filter out the annotator noise in the problems with low ambiguity of ground truth labels. 
    
    \item For the segmentation problem, we introduce a new term in the loss function, which takes into account the ``agreement'' between annotators in the regions where all of them give the same labels. This lets us directly use the base segmentation model in areas where there is little annotation noise.
    
    \item We provide a comprehensive set of experiments on MNIST, CIFAR-10, Fashion-MNIST, and CIFAR-10N (real-world noisy data) datasets for classification. We also conducted a series of experiments on segmentation problems with annotation such as synthetic noisy data (based on MNIST) and real-world medical data (RIGA and LIDC). The experiments show that the proposed method provides state-of-the-art results in the considered problems.
  \end{enumerate}
  Our \href{https://github.com/Singularity-AI-Lab/Confident-models-under-annotator-noise}{code} includes a suite that allows researchers to compare their approach against benchmarks considered in this paper.

\vspace{-5pt}
\section{Related Work}
  Learning with noisy labeled training data has been an active area of research for some time. Various algorithms have been introduced and have shown resistance to noise during training. We highlight the core research being done in this domain.

\vspace{-5pt}
\subsection{Classification}
  
\noindent{\bf Noise Transition Matrix/Loss Correction.} 
  The loss correction approach~\cite{Chen,Bekker} using a noise transition matrix, $T$, is a crucial branch that is used in deep learning systems. The goal of loss correction is for training on noisy labels with the corrected loss to be roughly equivalent to training on clean labels with the original loss. The paper~\cite{Patrini2017} introduced two different approaches for loss correction using a stochastic matrix $T$ that delineates the probability of a class being flipped with another under a certain noise. To address extreme noise, \cite{Hendrycks2018} suggested Gold Loss Correction (GLC) based on Forward Correction~\cite{Sukhbaatar2015}. Other similar works exploring this approach are~\cite{Reed2015, Goldberger2017TrainingDN}.

\noindent{\bf Multi-Network Learning.}
  Multi-network~\cite{Decouple, MentorNet} training frequently employs collaborative learning and co-training. Therefore, the sample selection procedure is governed by the mentor network in the case of joint learning and the peer network in the case of co-training. Co-teaching~\cite{Co-teaching} and Co-teaching+~\cite{coteachingplus} also employ two DNNs, but each DNN selects a certain number of small-loss examples and feeds them to its peer DNN for additional training. In comparison, JoCoR~\cite{JoCoR} reduces the diversity of two networks by means of co-regularization, making the predictions of the two networks more similar. 

\noindent{\bf Robust Regularization.}
  In~\cite{Tanno2019} showcased a method for simultaneously learning the individual annotator model and the underlying true label distribution through the use of trace regularization. The paper~\cite{Menon2020Can} suggests a composite loss-based gradient clipping for label noise robustness, given that excessive trust is not placed in any single sample. 

\noindent{\bf Other Deep Learning/Statistical Methods.}
  The authors of~\cite{PENCIL} proposed a framework that updates label distributions by repeatedly correcting the noisy labels. The paper~\cite{CDR} suggested a robust early-training method to diminish the side effects of noisy labels prior to early stopping. DivideMix~\cite{DM} splits the data into labeled (clean samples) and an unlabelled set (noisy samples) and train it in a semi-supervised approach.

\vspace{-5pt}
\subsection{Segmentation.}
\vspace{-5pt}
  Training deep neural networks on noisy label annotations can overfit the noise in the dataset~\cite{silva2022noise}. Inter-reader variability among annotators gave prominence to the STAPLE; \cite{Warfield}) algorithm that uses Expectation-Maximization to merge segmentation masks from various annotators into estimating a single ground truth. Several algorithms drew their inspiration from the STAPLE framework such as~\cite{Kats2019soft, Songbai2005, Asman}. 

  In~\cite{mirikharaji2019learning}, authors provide a sample re-weighting strategy that considers the expertise level of annotators. This strategy gives greater weights in the loss function for the samples annotated by professionals. To disengage annotator bias, the paper~\cite{DisentanglingHumanError} uses two coupled DNNs. Similar to~\cite{Tanno2019}, the DNN for segmentation estimates the label distribution, while the DNN for annotation is representative of the annotator bias using a confusion matrix. Similar ideas were applied in~\cite{li2021provably}, but with another regularization that led to the recovery of the ground truth confusion matrix and ground truth posterior \( p(y \mid x) \) under the assumption of diagonal dominance for the confusion matrix.

  Recently, various projects have begun to investigate the impact of multi-rater labels using label sampling~\cite{jensen2019improving, Jungo} or multi-head~\cite{guan2018said} techniques. Models trained with multi-rater labels are believed to be better calibrated than those trained with traditional ground-truth labels, such as majority vote, which are prone to overconfidence~\cite{jensen2019improving, Jungo}.
  The paper~\cite{kohl2018probabilistic} proposed learning a distribution of possible segmentations from a given input image. It combines U-Net~\citep{ronneberger2015unet} with conditional Variational Autoencoders (VAE) to generate a wide range of potential segmentation outcomes.  


\section{Methodology}
\label{sec:methodology}

\subsection{Noisy Labels in Deep Learning.}
\label{sec:noisy_labels}
  In this section, we introduce the problem of training a deep learning model in the presence of multiple annotations. Since each of the annotations is received from some expert with his expertise and biases, we consider these annotations as ``noisy''.

  Let $\mathcal{X}$ denote the space that contains a set of input training data $X \eqdef \{x_1, \ldots, x_N\}$. Each of these objects $x_i$ has a corresponding ground-truth label $y_i$ such that $Y \eqdef \{y_1, y_2, \ldots, y_N\} \subseteq \mathcal{Y}$, where $\mathcal{Y}$ is the space of labels. However, in many scenarios, we don't have access to ground-truth labels and only have potentially noisy annotations available. Thus, we assume our training dataset \(\mathcal{D}\) to be a collection of \(N\) input images $x_i$ and their corresponding noisy annotations received from a subset of fixed annotators.
  By \(\annset\), we denote the set of indices of those annotators who provided their annotations for an input image \(\inimage\). We assume that overall we have \(R\) annotators, and their identities are preserved for all the images.

  The dataset therefore has the following structure: \(\mathcal{D} = \bigl\{\inimage, \{ \ynoisy \}_{r \in \annset}\bigr\}_{i=1}^N \), where \(\ynoisy\) corresponds to an annotation of \(r\)-th expert for the input image \(\inimage\). The tilde sign reminds us that the annotation is noisy.

  It is worth noting that the ground truth annotations \(\ygt\) are not available to us but assumed to exist. Thus, our goal is to build a model that will approximate the distribution of targets \(p(\ygt \mid \inimage)\) for a given input solely from the dataset of noisy annotations.

\subsection{Probabilistic Model for Noisy Observations and Architecture.}
\label{sec:model}
  In this section, we build a probabilistic noise model for observed noisy annotations.
  Following~\cite{Tanno2019, DisentanglingHumanError}, we assume statistical independence of annotators.

  For the following derivations, we will omit the object index \(i\) to simplify the notation.
  Taking these assumptions into account, we can write the joint likelihood for all annotations \(\{\ynoisyni\}_{r \in \annsetni}\) of a given input image \(\inimageni\):
  \begin{equation}
  \textstyle 
    p (\{\ynoisyni\}_{r \in \annsetni} \mid \inimageni ) =  {\prod}_{r \in \annsetni} \, p(\ynoisyni \mid \inimageni).
    \label{eq:noisy_ll}
  \end{equation}

  Each of the probabilities \(p(\ynoisyni \mid \inimageni)\) inside the product is an implicit marginalization of the latent ground truth class label \(\ygtni\):
  \begin{equation*}
    \textstyle 
    p(\ynoisyni \mid \inimageni) = \sum _{c = 1}^C p(\ygtni = c \mid \inimageni) \, p(\ynoisyni \mid \ygtni = c, \inimageni),
  \end{equation*}
  where $C$ represents the number of classes for the true labels, $y \in [1, \dots, C]$.
  
  Since the annotators' noise is dependent on the sample $x$, this allows us to model noisy label distribution as $p(\tilde{y}^{r}=j \mid y=k, x) =: u^{r}_{jk}(x)$. These noisy class conditional distributions can be considered as elements of a confusion matrix (CM), effectively modeling the noise introduced by the annotator.

  Now, the marginalization can be rewritten as follows:
  \begin{equation}
    \textstyle 
    p(\ynoisyni \mid \inimageni) = \sum\nolimits_{c = 1}^C u^r_{\ynoisyni c}(x)  \cdot p(y = c \mid x).\ \ \ 
  \label{eq:noisy_marginal}
  \end{equation}
  In this equation, we see that the predictive distribution for a noisy annotation is a mixture of class-conditional distributions of noisy annotations weighed by the ground truth label distributions.

  Inspired by~\cite{DisentanglingHumanError, Tanno2019}, we use a two-component model to approximate the distributions in equation~\eqref{eq:noisy_marginal}. The first component is the base prediction network, which solves the problem of interest. In the case of classification, we call it the classification model, in the case of segmentation -- segmentation model. The other component is an annotation network.
  The base prediction network is a convolutional neural network parameterized by \(\theta\) that aims to approximate the ground truth probability distribution \(p(\ygtni \mid \inimageni)\). 
  Hence, it transforms an input image \(\inimageni\) to a probability map \(\segnetni \in \R^{C}\) (for more details in the case of segmentation see Section~\ref{sec:appendix_segmentation_architecture}). 

  The annotation network, parameterized by \(\psi\), is designed to estimate the confusion matrices associated with each annotator, treating them as functions of the input image. Specifically, for each annotator \(r\), the network produces a matrix 
  $
    \hat{T}^\psi_r(x) \in [0, 1]^{C \times C},
  $
  where \(C\) is the number of classes. Each entry in this matrix, denoted as \(\hat{T}^\psi_r(x)[k, l]\), represents the estimated probability \(p(\ynoisyni = k \mid \ygtni = l, \inimageni)\) that an annotator labels a sample of true class \(l\) as class \(k\), conditioned on the input \(x\).

  The multiplication of these two terms $\hat{p}^{r}_{\theta, \psi}(x) = \annetni \times \segnetni$ is the $r$-th annotator's probability map which is optimized in the loss function (the details are covered in Section~\ref{sec:loss}). The architecture of the resulting model for the case of classification is depicted in Figure~\ref{fig:Our_NN}.

\section{Training and Confidence Regularization}
\label{sec:regularization}
  In this section, we explain in detail the training procedure we use and introduce a new class of regularizers, specially tailored to the problems with label noise.

\subsection{Loss function.} 
\label{sec:loss}
  We introduce a loss function, which allows us to jointly optimize the parameters $\theta$ and $\psi$ of the prediction and annotation networks. 

  For this, we maximize the likelihood of the observed noisy annotations.
  Using~\eqref{eq:noisy_marginal}, we can rewrite the negative log-likelihood for \(N\) training objects in the following form:
  \begin{multline}
  \label{eq:main_loss}
    \textstyle -\log  {\prod_{i=1}^N}  {\prod_{r \in \annset}} \text{Cat}(\ynoisy; p_{\psi, \theta}^{r}(\inimage)) \\
      =
    \textstyle \sum_{i = 1}^N \sum_{r \in \annset} CE\bigl(\annet \, \segnet, \ynoisy\bigr),
     \nonumber
  \end{multline}
  where \(CE\) means the cross-entropy and $\text{Cat}$ is the probability mass function of Categorical distribution.

  Minimizing the above objective function would jointly optimize both the parameters \(\theta\) and \(\psi\). However, minimizing this function alone does not guarantee to disentangle the annotator's noise distribution from the ground truth label distribution. This is because there could be many combinations of \(\annetni \, \segnetni\) that provide good approximations to \(\ynoisyni\). 
  This means that the prediction network's output \(\segnetni\) may not learn to correctly discriminate between noise and true labels. To overcome this, one may consider introducing some regularizer \(R\bigl(\segnetni\bigr)\) that would help the disentanglement. 

  Then, the resulting loss function parameterized by \(\lambda > 0\) becomes
  \vspace{-5pt}
  \begin{equation*}
    \textstyle
    \sum_{i = 1}^{N} {\sum_{r \in \annset}} CE\bigl(\annet \, \segnet, \ynoisy\bigr)
    + \lambda \sum_{i = 1}^{N} R\bigl(\segnet\bigr).
  \end{equation*}
  \vspace{-8pt}
  
  \noindent In the next section, we will discuss the particular choice of the regularizer $R\bigl(\segnetni\bigr)$.

\subsection{Rationale for Confidence Regularization.}
\label{sec:regularizer}  
  In this section, we discuss the choice of the regularizer needed for training the considered model. The motivation for using some regularizer is straightforward -- there are infinitely many solutions for the components of the product \(\annetni \, \segnetni\), which maximizes the likelihood of noisy annotations. 

  \noindent We start by discussing the regularizer from~\cite{Tanno2019,DisentanglingHumanError}:
  \begin{equation*}
  \textstyle
    R_{tr}(x) = \mathrm{trace}\left[\frac{1}{|\annsetni|} \sum_{r \in \annsetni} \annetni\right],
  \end{equation*}
  where \(\frac{1}{|\annsetni|} \sum_{r \in \annsetni} \annetni\) is the confusion matrix averaged over annotators. 
  The motivation of this regularizer is to push the estimated confusion matrix \(\annetni\) to converge to the true confusion matrix of the annotator. It is shown in~\cite{DisentanglingHumanError}, that the minimizer of the trace includes a true confusion matrix as a solution. 

  However, the proof present in~\cite{DisentanglingHumanError} is based on some assumptions. Most importantly, they assume diagonal dominance of the mean confusion matrix \(\frac{1}{|\annsetni|} \sum_{r \in \annsetni} \annetni\). This assumption is not necessarily satisfied in practice. Moreover, the considered regularizer acts in the opposite direction, i.e., it pushes the estimate to be not diagonally dominant. Thus, this regularizer might only spoil the training process because, intuitively, it prevents annotators from being confident.
  
  In this work, we propose a new class of regularizers, that leads to better results without the need for regularizing parameters of the annotation network and the use of architecture modifications.
  Specifically, we push the base prediction network to provide confident predictions.
  Implicitly, we assume that there is no aleatoric noise in the ground-truth distribution of labels, the assumption previously considered in~\cite{DisentanglingHumanError}. This assumption, despite seeming restrictive, holds in many image analysis datasets~\cite{kapoor2022uncertainty} and can be reached in practice, given better instuctions to annotators. Hence, \(p(y \mid x) = \textbf{e}\), where \(\textbf{e}\) is a one-hot vector with one placed to the correct label location. Based on these considerations, we propose a class of regularizes, that encourages base model to provide confident predictions. We consider two instances of this class: the so-called \textbf{confidence} regularizer:
  \begin{equation}
     R\bigl(\segnetni\bigr) = \min\limits_{c \in \overline{1,C}} \bigl(-[\log \segnetni]_c\bigr).
  \label{eq:confidence_regularizer}
  \end{equation}
  This function attains minimum if one of the predicted classes has a predicted probability equal to \(1\). 

  As another instance of this class, we consider predictive \textbf{entropy} regularizer:
  \begin{equation}
    R\bigl(\segnetni\bigr) = -\sum\nolimits_{c = 1}^C 
     \bigl[\segnetni\bigr]_c
     \bigl[\log \segnetni]_c\bigr.
  \label{eq:entropy_regularizer}
  \end{equation}

\section{Experiments}
\label{sec:experiments}

\subsection{Classification.}
  \begin{table*}[t!]  
    \caption{Comparison of test accuracy (\%) (mean $\pm$ st. dev.) for CIFAR-10. \textbf{Best} results are in bold, \underline{second-best} underlined. ``Conf'' stands for confidence regularizer, ``Ent'' for predictive entropy regularizer.}\label{tab:cifar10_data}
    \centering
    \resizebox{0.6\textwidth}{!}{%
    \begin{tabular}{c r r r r r r r}
      \toprule
      Noise rate& Ours-Conf & Ours-Ent & Co-tea & Co-tea+ & JoCoR & Trace & CDR \\
      \midrule
      
      symmetric 20\% & \textbf{84.22}  & \underline{84.00} & 81.82 & 80.42 & 82.12& 82.86 & 81.01  \\ 
                 & $\pm$ 0.44  & $\pm$ 0.41& $\pm$ 0.18& $\pm$ 0.14& $\pm$ 0.07 & $\pm$ 0.59& $\pm$ 0.13 \\ 
       
      symmetric 50\% & \textbf{80.03} & \underline{79.64} & 75.74 & 75.09 & 76.60 & 77.82 & 69.68\\ 
              & $\pm$ 0.22& $\pm$ 0.26 & $\pm$ 0.09& $\pm$ 0.15& $\pm$ 0.09 & $\pm$ 1.17& $\pm$ 0.59\\ 
              
      symmetric 80\% & \textbf{45.76} & \underline{43.21} & 19.84 & 18.01 & 28.32 & 33.61 & 35.38\\ 
              & $\pm$ 0.11& $\pm$ 0.15& $\pm$ 0.10& $\pm$ 0.08& $\pm$ 0.35 & $\pm$ 1.21& $\pm$ 0.40\\ 

      pairflip 20\% & \textbf{84.92} & \underline{84.78} & 81.17 & 80.78 & 81.86 & 83.86 & 82.89\\ 
                  & $\pm$ 0.39& $\pm$ 0.43& $\pm$ 0.13& $\pm$ 0.11& $\pm$ 0.09& $\pm$ 0.42& $\pm$ 0.54\\ 

      pairflip 30\% & \underline{84.36} & \textbf{84.54} & 79.53 & 79.49 & 79.52 & 83.15 & 82.08\\ 
                 & $\pm$ 0.43& $\pm$ 0.39& $\pm$ 0.16& $\pm$ 0.12& $\pm$ 0.25& $\pm$ 0.46& $\pm$ 1.33 \\ 

      pairflip 45\% & \textbf{83.43} & \underline{81.23} & 59.04 & 53.07 & 67.59 & 75.88 & 58.56\\ 
                 & $\pm$ 0.32& $\pm$ 0.36& $\pm$ 0.15& $\pm$ 0.31& $\pm$ 0.36& $\pm$ 2.13& $\pm$ 1.60 \\ 
      \bottomrule
    \end{tabular}
    }
      \caption{Comparison of test accuracy (\%) (mean $\pm$ st. dev.) for CIFAR-10N. \textbf{Best} results are in bold, \underline{second-best} underlined. ``Conf'' stands for confidence regularizer, ``Ent'' for predictive entropy regularizer.}\label{tab:cifar10n_data}
    \centering
    \resizebox{0.7\textwidth}{!}{%
    \begin{tabular}{c @{\hskip 0.07in}c @{\hskip 0.07in}c @{\hskip 0.07in}c @{\hskip 0.07in}c @{\hskip 0.07in}c @{\hskip 0.07in}c @{\hskip 0.07in}c}
      \toprule
      Noise type & Ours-Conf & Ours-Ent & Co-tea & Co-tea+ & JoCoR & Trace & CDR \\
      \midrule
      worst & \underline{73.49 $\pm$ 0.61}& \bf{73.58 $\pm$ 0.96} & 67.97 $\pm$ 1.17 & 65.45 $\pm$ 0.25 & 68.74 $\pm$ 0.21 & 70.23 $\pm$ 0.89 & 59.87 $\pm$ 0.58 \\
      \bottomrule
    \end{tabular}
    }
  \end{table*}
\noindent
In this section, we validate our approach to classification problems. For baselines, we will use the following approaches: Co-teaching~\cite{Co-teaching}, Co-teaching$+$~\cite{coteachingplus}, JoCoR~\cite{JoCoR}, Robust Early-learning (CDR; \cite{CDR}) and Annotator Confusion (Trace; \cite{Tanno2019}). A detailed description of these methods is in the Appendix~B.1.

\noindent{\bf Datasets.}
\label{sec:datasets_classification}
  In this work, we consider a dataset with synthetically generated annotations along with annotations from human experts. For classification problem, we consider the standard benchmark datasets: MNIST~\cite{deng2012mnist}, Fashion-MNIST~\cite{xiao2017fashion}, CIFAR-10~\cite{krizhevsky2009learning} where noises are generated synthetically and CIFAR-10N dataset (real-world human annotations noisy data)~\cite{wei2022learning} to demonstrate the effectiveness of our methodology. The noise types, used in the classification experiments, are as follows.
  \begin{itemize}[nosep,topsep=0pt]
    \item The ``pairflip noise'' involves swapping the labels of two adjacent categories/classes based on a preset ratio~\cite{Liang2022}.
  
    \item The ``symmetric noise'' retains a portion of the original labels and uniformly reassigns the remainder to all other categories~\cite{Ma}.

    \item The ``worst'' noise type (considered exclusively for CIFAR-10N dataset) stands for the highest label noise among human annotations.
  \end{itemize}
  Detailed descriptions of datasets and examples of noise transition matrices are provided in Appendix C.1.

\noindent {\bf Performance Evaluation.}
  We conducted our experiments on four datasets that utilize artificial and real-world noisy labels. In our experiments for the classification task, we synthetically introduced noise to the training data for MNIST, Fashion-MNIST and CIFAR-10 datasets; we chose various noise rates, such as 20$\%$, 30$\%$, 45$\%$, 50$\%$ and 80$\%$ . We used the ``worst'' noise type for CIFAR-10N real-world noisy annotations data; it comprised of 40$\%$ noise (non-synthetic) in the dataset.
  From these noisy labels, we aim to train a model that will be good at predicting the ground truth labels. 
  
  We evaluate the performance of our algorithm in terms of test accuracy for the classifier network. We are particularly interested in the performance of the classifier, as in the evaluation stage this network will be used separately to make predictions. Details on tuning the classification model are in Appendix D.3.

\noindent {\bf Results.}
  In this section, we present an analysis of our proposed approach with other methods for different datasets (see Table~\ref{tab:cifar10_data} for CIFAR-10, Table~\ref{tab:cifar10n_data} for CIFAR-10N and Appendix D.3
  for the rest). For CIFAR-10, CIFAR-10N, and MNIST, our approaches yield the highest performance results across all noise rates and types. Our algorithm has shown robust performance across most baselines for Fashion-MNIST. We see comparable performance among all the algorithms when the noise rate is 20$\%$ and 30$\%$ for both symmetric and pairflip noise types. In general, for higher noise ratios regardless of noise type, which are evidently more challenging cases, our algorithm consistently gives better performance.

  \begin{figure}[t]
  \captionof{table}{Annotator information for three different styles (MNIST).}\label{tab:info}
  \resizebox{\columnwidth}{!}{%
    \begin{tabular}{c c c c}
      \toprule
      Annotators & Original & Thin & Thick \\
      \midrule
      Annotator 1 & symmetric 80\%  & asymmetric 40\% & pairflip 95\%  \\ 
      Annotator 2 & pairflip with permutation 40\%  & symmetric 95\% & asymmetric 70\% \\ 
      Annotator 3 & pairflip 60\%  & pairflip with permutation 40\%  & symmetric 80\%  \\
      \bottomrule
    \end{tabular} }

    \captionof{table}{Comparison of Dice coefficient (\%) for different methods of segmentation task and datasets (mean $\pm$ standard deviation). \textbf{Best} results are in bold, \underline{second-best} underlined.}\label{tab:segment}
    \centering
    \resizebox{\columnwidth}{!}{%
     \begin{tabular}{l @{\hskip 0.12in}c @{\hskip 0.12in}c @{\hskip 0.12in}c @{\hskip 0.12in}c @{\hskip 0.12in}c}
     \toprule
     Datasets & MR-Net & CM-Net & PU-NET & \makecell{Ours\\(no CR)} & Ours \\
     \midrule

     MNIST & \underline{95.81 $\pm 0.45$} & 93.85 $\pm 0.03$ & 89.90 $\pm 0.25$ &  94.47$\pm 0.02$  & \textbf{96.61}$\pm 0.00$  \\ 
     RIGA & \underline{93.21 $\pm 1.12$} & 90.03 $\pm 0.91$ & 91.23 $\pm 1.21$ &  92.06 $\pm 0.58$ & \textbf{93.85 $\pm 0.09$} \\ 
     LIDC & 72.73 $\pm 2.70$ & 72.75 $\pm 0.66$ & 71.83 $\pm 1.50$ &  \underline{75.93 $\pm 0.53$} & \textbf{82.07 $\pm 0.92$} \\ 

     \bottomrule
    \end{tabular}
    }
  \end{figure}

\noindent{\bf Instance dependent noise and Curated Dataset.}\label{sec:curated}
  In this section, we use our proposed method in an important setup of \textit{instance-dependent noise}. For this, we have created a dataset that is based on MNIST. Using Morpho-MNIST software~\cite{castro2019morphomnist} we made morphological transformations of images, resulting in different digit styles. We refer to this dataset as \textbf{Curated MNIST}.

\begin{figure*} 
  \centering
    \resizebox{0.65\textwidth}{!}{
    \renewcommand{\arraystretch}{0.8}
    \begin{tabular}{c c c c}
      Image & Ground truth & \makecell{Ours (m=2, \\ ($\lambda$=0.01)} & \makecell{Ours \\ ($\lambda$=0)} \\
      \includegraphics[width=0.18\linewidth]{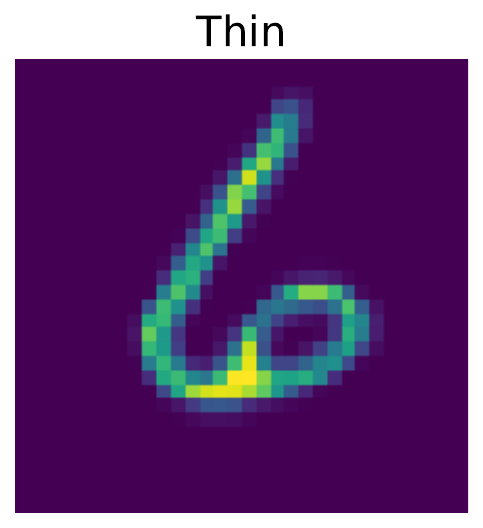} & \includegraphics[width=0.2\linewidth]{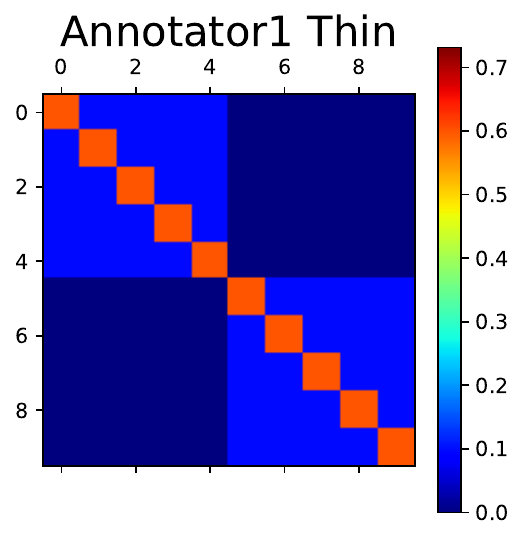} &  \includegraphics[width=0.2\linewidth]{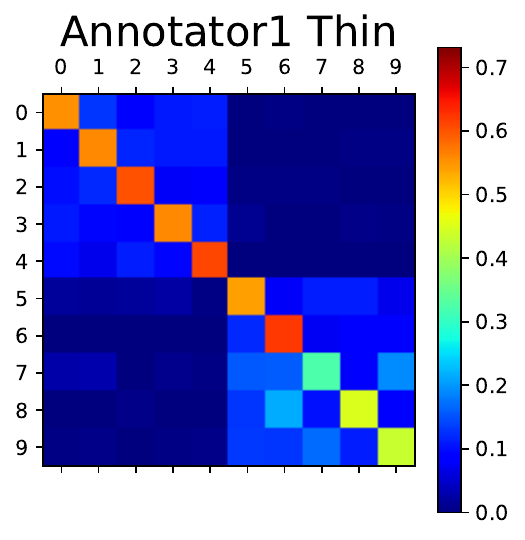}  & \includegraphics[width=0.2\linewidth]{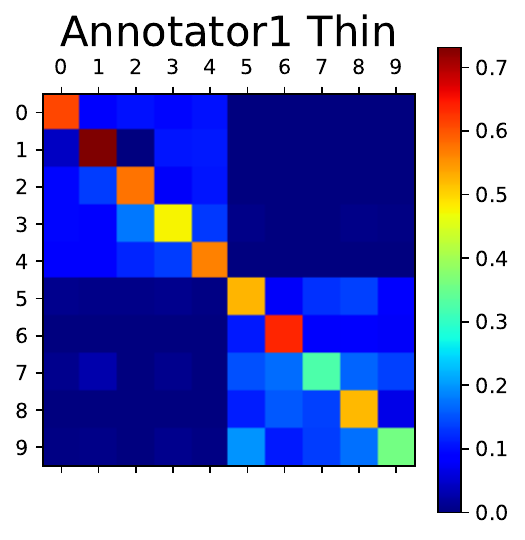}  \\ 
 
      \includegraphics[width=0.18\linewidth]{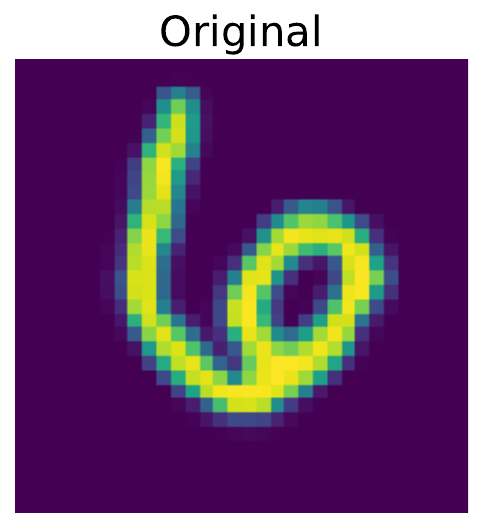} & \includegraphics[width=0.2 \linewidth]{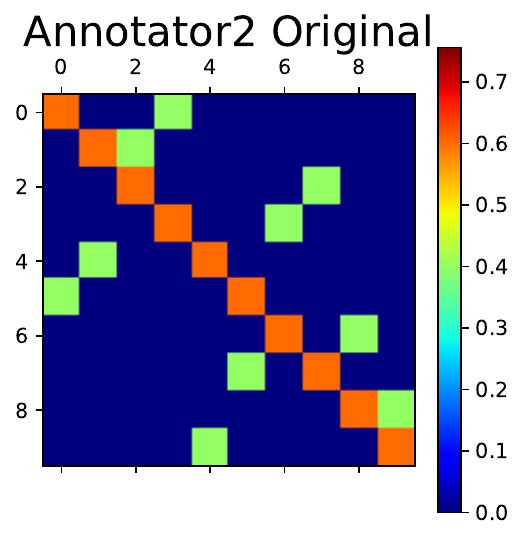} &  \includegraphics[width=0.2 \linewidth]{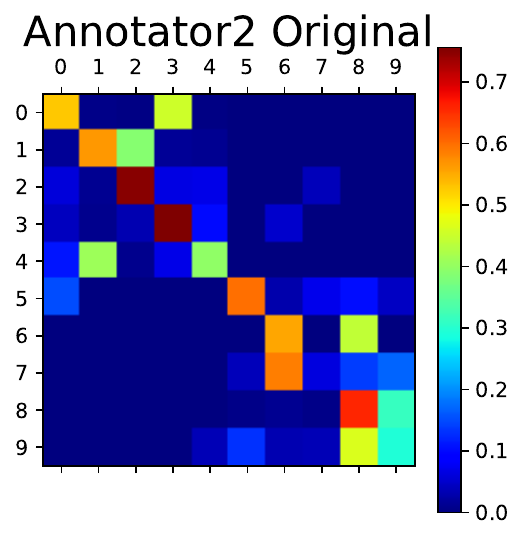}  & \includegraphics[width=0.2 \linewidth]{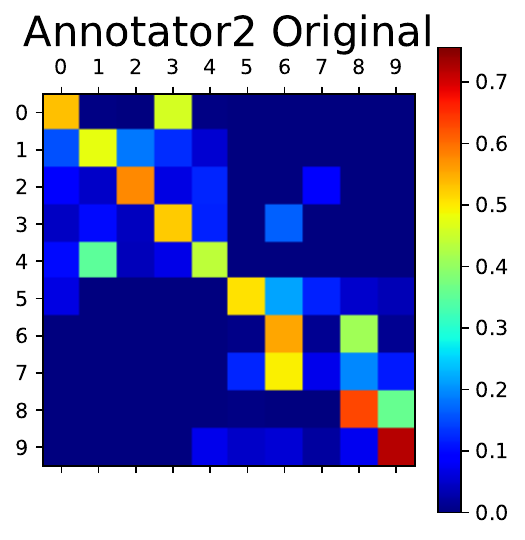}  \\ 

      \includegraphics[width=0.18\linewidth]{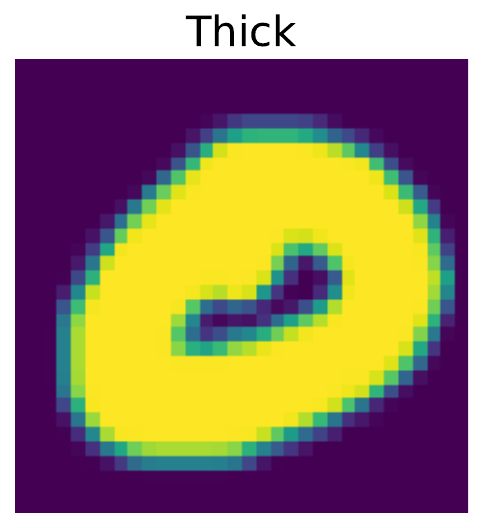} & \includegraphics[width=0.2 \linewidth]{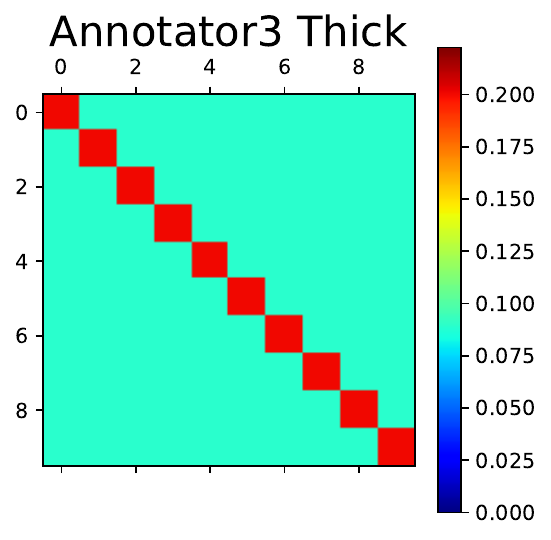} &  \includegraphics[width=0.2 \linewidth]{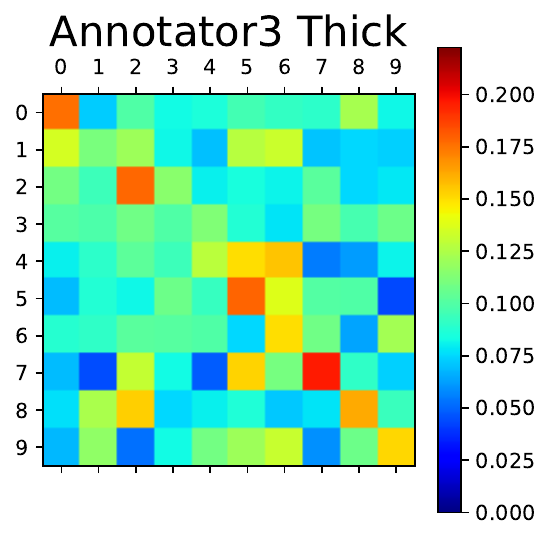}  & \includegraphics[width=0.2 \linewidth]{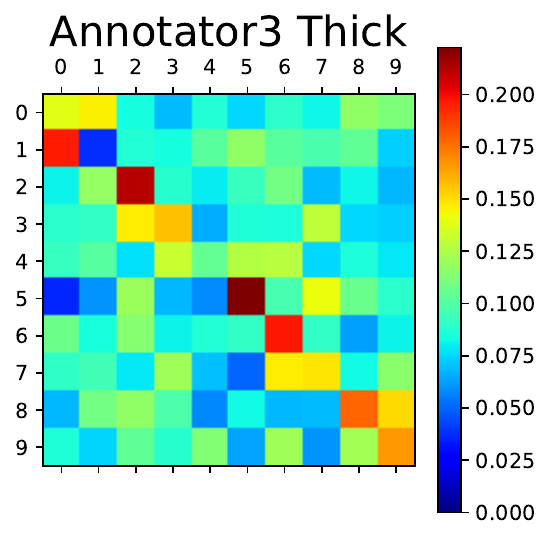}  
    \end{tabular}
    }
    \vspace{-10pt}
  \caption{Ground truth (left column) and predicted confusion matrices for different Annotators using different models: our approach with confidence regularizer ($\lambda=0.01$, m=2) (middle column) and without it ($\lambda=0$) (right column) on \textsl{Curated MNIST.}}
  \label{tab:SMs_main}
    \centering
    \vspace{0pt}
    \setcounter{subfigure}{0}
    \begin{subfigure}{.49\linewidth}
      \centering
      \includegraphics[width=.9\textwidth]{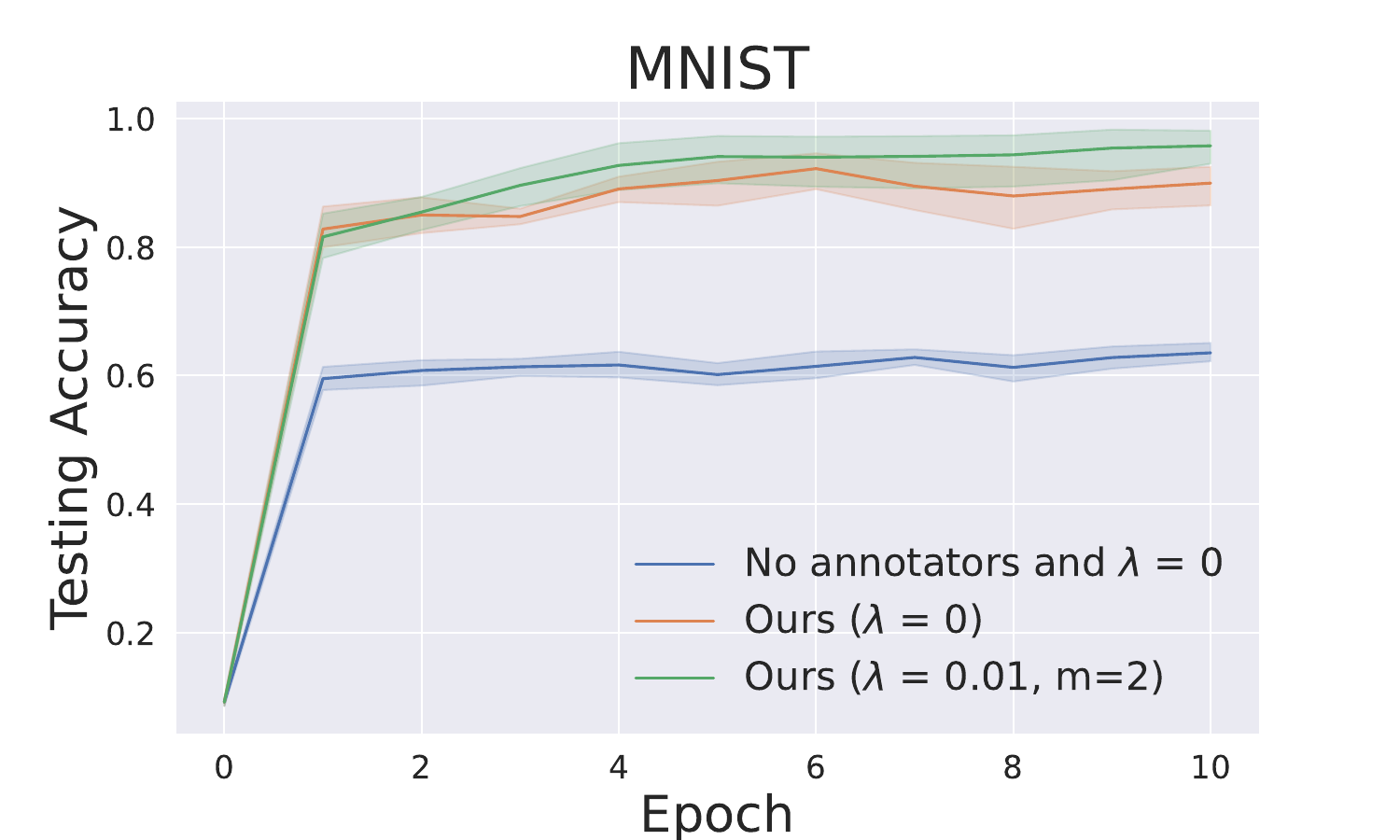}
      \caption{Testing accuracy}
    \end{subfigure}
    \begin{subfigure}{.49\linewidth}
      \centering
      \includegraphics[width=.9\textwidth]{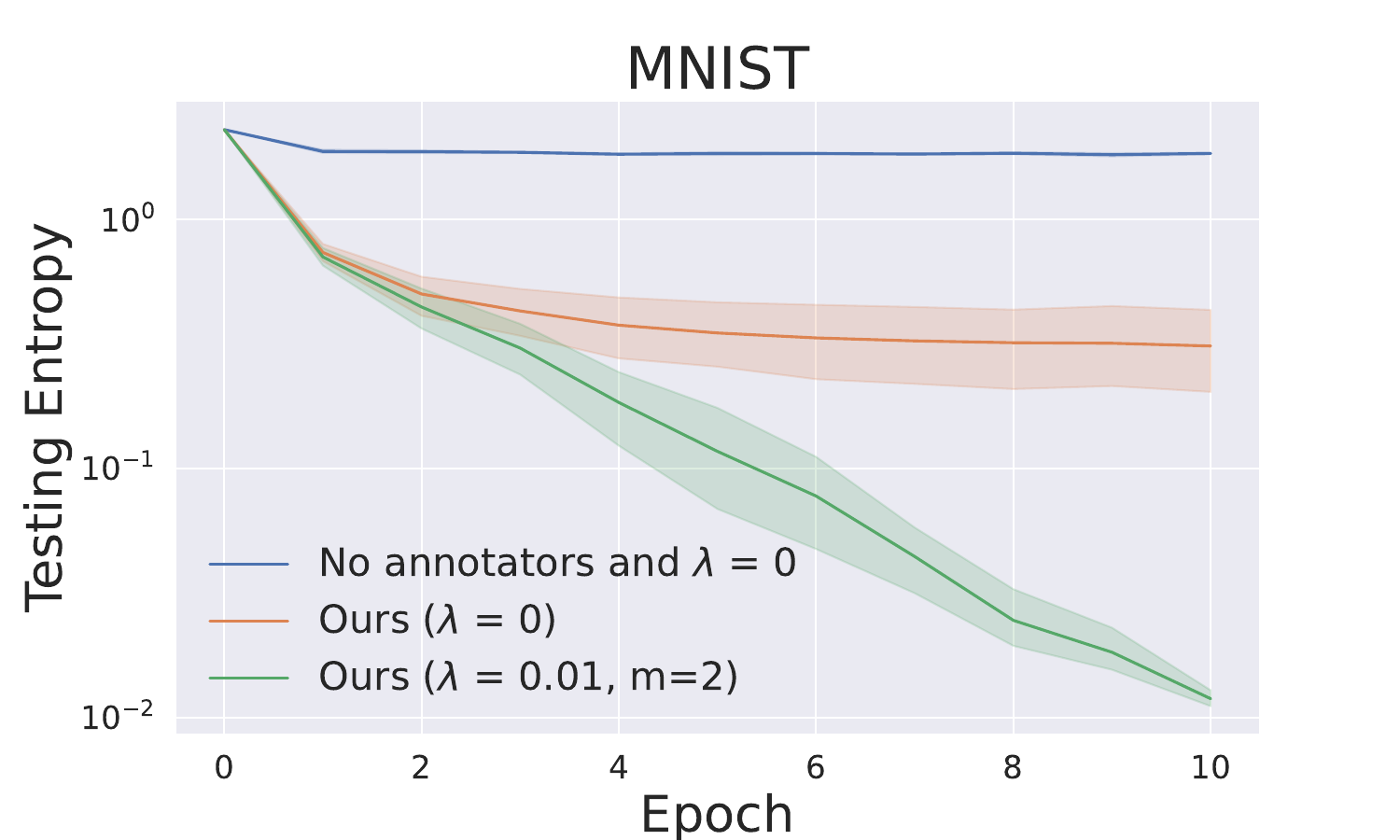}
      \caption{Testing entropy}
    \end{subfigure}
  \vspace{-5pt}
  \caption{Accuracy and Entropy for \textbf{Curated MNIST} test dataset computed over seven different seeds. The bold line is the average, shaded region is std. dev.}
  \label{fig:testing_summary}
  \resizebox{0.65\textwidth}{!}{
  \renewcommand{\arraystretch}{0.8} 
    \begin{tabular}{c c c}
      \includegraphics[width=0.1  \linewidth]{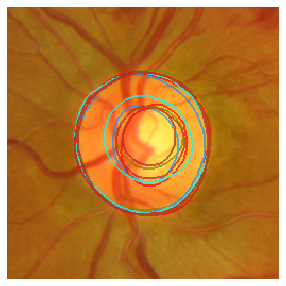} & 
      \includegraphics[width=0.1  \linewidth]{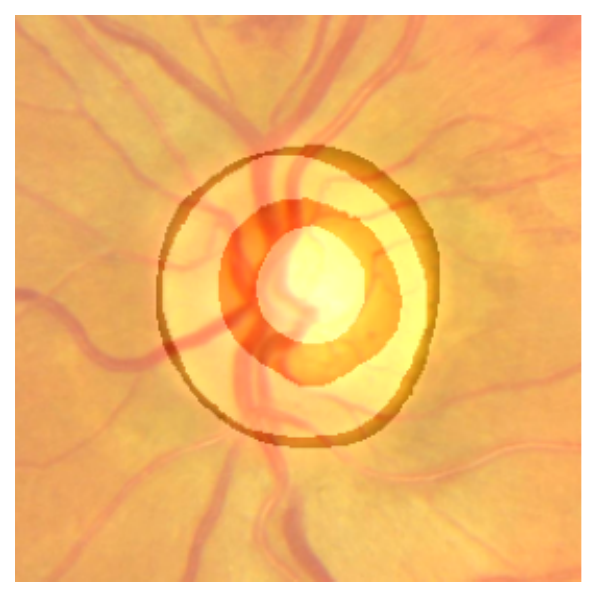} & 
      \includegraphics[width=0.1  \linewidth]{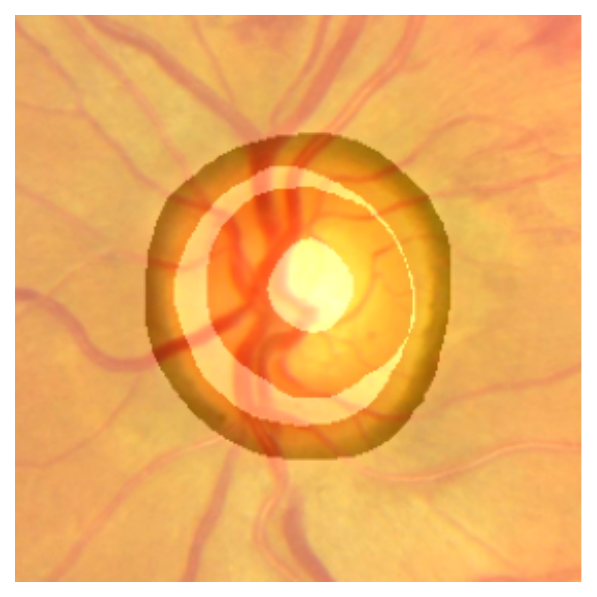} \\

      \includegraphics[width=0.1  \linewidth]{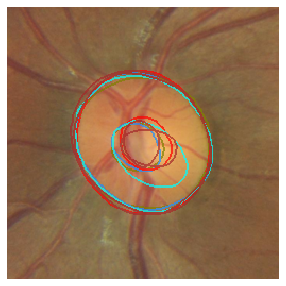} & 
      \includegraphics[width=0.1  \linewidth]{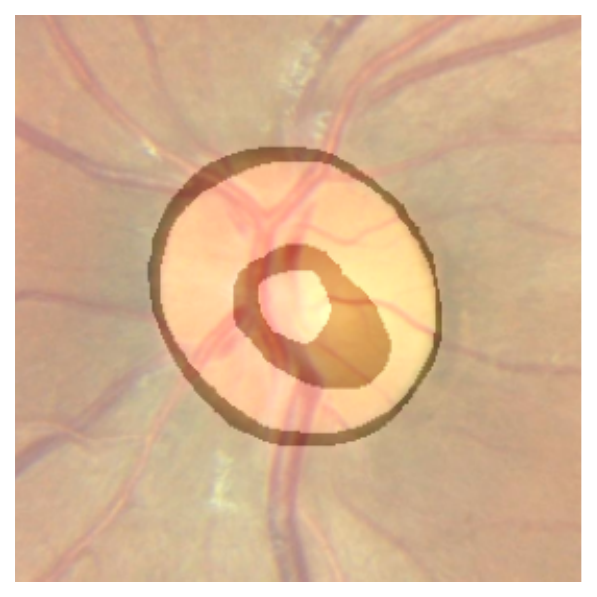} & 
      \includegraphics[width=0.1  \linewidth]{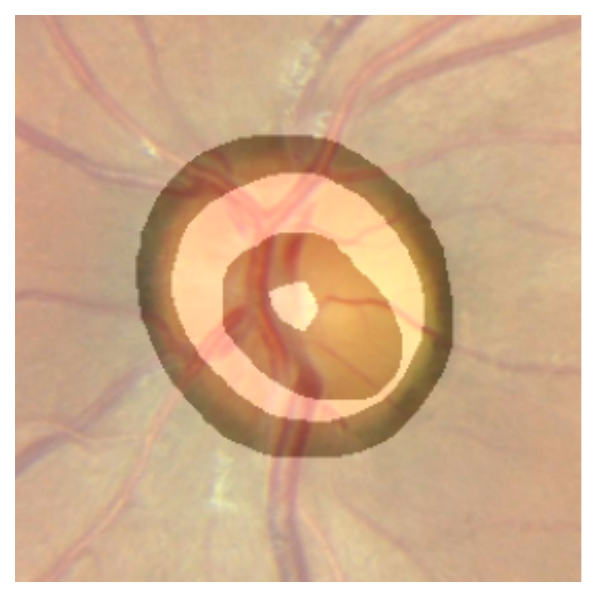} \\
    \end{tabular}
    }
    \vspace{-10pt}
    \captionof{figure}{Visualization of CR with different dilation radius of uncertainty regions (RIGA Optical Cup). \textbf{Left col.} Annotators' boundaries, \textbf{Middle col.} Initial (d=0), \textbf{Right col.} Dilated (d=3). Here, uncertainty regions have natural colors, while CR is whitened.}
    \label{fig:riga_final_new}
  \end{figure*}

  Specifically, each image in the Curated MNIST dataset may be either of three different styles. Namely, Original (no transformations are applied), Thick (the digit's contours become thicker), and Thin (contours become thinner). 
  See Figure~\ref{tab:SMs_main}, leftmost column for the demonstration. 
  We have three different annotators, \textit{who assign different noisy labels, based on the input image style}, making the \textbf{noise instance-dependent}. This is an important experimental setup, that reflects real-world situations. Moreover, the types of noise introduced by annotators are different (see Table~\ref{tab:info}). 

In this experiment, we compare (i) the base classifier model, trained with standard cross-entropy loss (without annotators and regularization), (ii) our approach but without regularization, and (iii) our approach with confidence regularizer (see eq.~\eqref{eq:confidence_regularizer}). Each annotator's model has a similar architecture as the classifier model and takes images as an input.

In Figure~\ref{tab:SMs_main}, we demonstrate ground truth and learned confusion matrices. Note, that for Thick digit style, noise rates are significant (see Table~\ref{tab:info}), and despite it is hardly possible to learn the true annotation matrix (e.g. for Annotator 1 for Thick style noise rate is 95\%), the resulting performance for the proposed method (see below) is still the best.
In Figure~\ref{fig:testing_summary}, we plot the testing accuracy and entropy of the base prediction network. Compared to the usage of a classifier model alone or with no regularizer, we see that our approach with the confidence regularizer is both more accurate and confident. More results are in Appendix D.3.2
 
\vspace{-7pt}
\subsection{Segmentation.}
  In this section, we evaluate our proposed approach to segmentation problems. As a baseline, we use the following approaches: CM-NET~\cite{DisentanglingHumanError}, PU-NET~\cite{kohl2018probabilistic} and MR-NET~\cite{MRNet}. A detailed description of these methods is in the Appendix B.2.

\noindent We refer readers to the Appendix D.2 for additional experiments where we adapt the regularizer from~\cite{li2021provably} to the segmentation problem and compare it to the approaches above.

\noindent{\bf ``Confident regions'' (CR) component in the loss.}
\label{sec:confident_regions}
  In this work, we also propose to directly apply the segmentation network \(\segnet\) for predictions for the pixels with low labeling noise. We construct such confident regions (CR) by a special procedure. First, we find such areas of the annotation masks for which all the annotators assign the same label. 
  Next, we use the heuristic that the pixels near the boundary of this region have a high potential for uncertainty. Because of this, we dilate the uncertainty region (complement of the unanimously segmented region) using the convolution operation with zero bias. The complement of this dilated uncertainty region is now a CR, which we will use in our loss function by adding a special term.
  The process of selecting the CR is illustrated by Figure~\ref{fig:riga_final_new}.

\noindent{\bf Datasets.}
\label{sec:datasets_segmentation}
  For the segmentation problem, we use MNIST~\cite{deng2012mnist} as a representative of the dataset with artificial noises, where synthetic noisy annotations are created by applying morphological transformations simulations on assumed ground truth~\cite{castro2019morphomnist}. As an example of a real-world dataset we utilize retinal fundus images for glaucoma analysis (RIGA; \cite{Almazroa2017AgreementAO}) and the lung image database consortium (LIDC; \cite{kohl2018probabilistic}) datasets. Descriptions of these datasets are in Appendix C.2.

\noindent{\bf Performance Evaluation.}
  To evaluate the segmentation model, we use the Dice coefficient on validation splits and report its mean and standard deviation computed over the last 10 epochs.
  Note the Dice coefficients are calculated for each sample individually, and then the average value is taken.
  Although the Dice coefficient between the prediction \(\segnetni\) and true label \(\ygtni\) is more important, we also consider Dice coefficient among the noisy annotations ($D(y_{a}, y_{a})$), Dice coefficient among predictions of annotations ($D(\tilde{y}_{a},\tilde{y}_{a})$), Dice coefficient between the annotations' predictions and noisy labels ($D(\tilde{y}_{a}, y_{a})$), Dice coefficient between noisy labels and true labels ($D(y_{a},y_{gt})$), and Dice coefficient between the annotations' predictions and true label ($D(\tilde{y}_{a},y_{gt})$). More details are provided in Appendix D.1.
  In the case of medical images, to make the evaluation, we compute the true label by majority voting. We emphasize that they are used only in the evaluation stage, not in the training.
  Along with these baseline models, we also run the model without involving the confidence region in our framework to see the effect of including it.

\noindent{\bf Segmentation Results.}
  In this section, we compare the performance of our proposed method with various competing approaches across multiple datasets. In Table~\ref{tab:segment}, we report the macro-averaged Dice coefficient, calculated as the average of individual class-wise scores.
  From this aggregate measure, it is seen that our method, enhanced with a confidence regions regularization, consistently outperforms the alternatives. 
  Notably, the standard deviation observed for our approach is generally lower than those of our competitors, indicating the superior stability and reliability of our method.

  Detailed tables for each dataset with all six types of Dice coefficients computed separately for each of the classes are provided in Appendix D.1.
  It can be seen that our model with a confidence regularizer outperforms the competitors in terms of class-wise Dice coefficients for the MNIST dataset. When applied to the LIDC dataset, our approach yields the highest performance with a substantial margin for class 1 (lungs) and the second-best result for class 0 (background). Our models show robust performance for classes 0 (background) and 1 (optical disc) on the RIGA dataset and achieve the highest score for class 2 (optical cup). We have included additional visualizations of the final segmentations achieved on each dataset in the Appendix D.1.
  Additional experiments, which study the effect of the regularizer, can be found in Appendix D.2.


\vspace{-6pt}
\section{Discussion}
  Our experiments demonstrate that the proposed methodology perform at least as competing approaches, frequently surpassing them in terms of both the average Dice coefficient (as shown in Table~\ref{tab:segment}) and class-specific Dice scores for segmentation tasks and in terms of test accuracy for classification task. A notable observation that appears consistently across all datasets is the superior performance of our model when it uses the proposed confidence or entropy regularizers. Moreover, for segmentation, the regularized model shows consistently higher Dice scores than the model without regularization across all classes and datasets. This consistency of the best performance, regardless of the diversity and complexity of datasets, shows the generalization capability of our model and proposed regularizers, which reassures its potential as a valuable tool for classification and image segmentation tasks of medical data.

\vspace{-6pt}
\section{Conclusion}
  In this paper, we have introduced a new approach for training the classification model with noisy labels and a segmentation model under the scenario of multiple noisy annotators. Our key contribution is the introduction of a class of regularizers that, unlike previous methods, exclusively penalizes the parameters of the base prediction network, thereby incentivizing the model to make more confident predictions. This regularizers allow us to improve the results over the competing approaches.

  Moreover, we introduce the concept of confident regions for the segmentation problem, identified as regions where all annotators unanimously agree on their predictions, into our loss function. This allows us to push the segmentation model explicitly to predict the class selected by all the annotators. We observed empirically in both, synthetic and real world noisy data, that this consistently helps to improve the results.

\vspace{-6pt}
\section*{Acknowledgements}
The research was partially supported by the RSF grant 20-71-10135. Asma Ahmed Hashmi is supported by the DAAD programme Konrad Zuse Schools of Excellence in Artificial Intelligence, sponsored by the German Federal Ministry of Education and Research.

\bibliographystyle{plain}
\bibliography{./siam_bibliography}


\clearpage
\setcounter{page}{1}

\appendix
\definecolor{semitransparentred}{rgb}{1,0,0} 
\colorlet{transred}{semitransparentred!20}

\section{Proposed architecture for Segmentation}
\label{sec:appendix_segmentation_architecture}
\begin{figure*}
 \centering
    \includegraphics[width=1.0\linewidth]{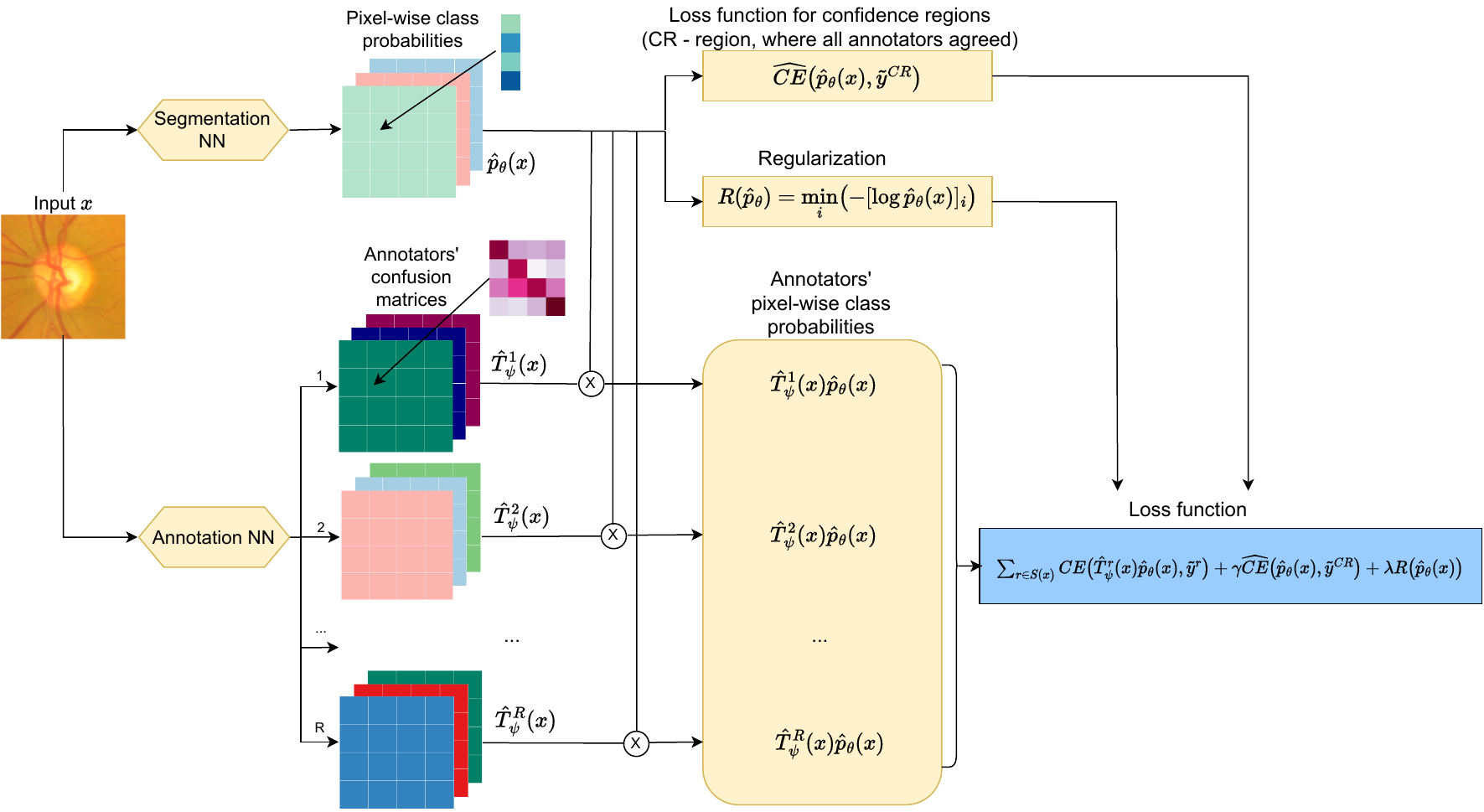}
    \caption{Illustration of the proposed architecture. The input image is fed into two separate networks -- Segmentation and Annotation. Then, their predictions are multiplied to produce the prediction of the noisy (annotators) segmentation masks. Finally, the loss function is computed, summing up all the components.}
    \label{fig:Our_segmentation_NN}
\end{figure*}

  In this section, we introduce the problem of training a segmentation model in the presence of multiple annotations. Since each of the annotations is received from some expert with his expertise and biases, we consider these annotations as ``noisy''.

  We assume our training dataset \(\mathcal{D}\) is a collection of \(N\) input images and their corresponding noisy annotations received from a subset of fixed annotators.
  By \(\annset\), we denote the set of indices of those annotators who provided their annotations for an input image \(\inimage\). We assume that overall we have \(R\) annotators, and their identities are preserved for all the images.

  The dataset therefore has the following structure: \(\mathcal{D} = \bigl\{\inimage, \{ \ynoisy \}_{r \in \annset}\bigr\}_{i=1}^N \), where \(\ynoisy\) corresponds to an annotation of \(r\)-th expert for the input image \(\inimage\). The tilde sign reminds us that the annotation is noisy.
  Each image and corresponding annotations have the following shapes: \( \inimage \in \R^{W \times H \times C} \) and \(\ynoisy \in \Y^{W \times H}\), where \(W, H, C\) are width, height and a number of channels correspondingly, and \(\Y = [1, 2, \ldots, L] \) is a set of possible class labels.

  It is worth noting that the ground truth segmentation annotations \(\ygt\) are not available to us but assumed to exist. Thus, our goal is to build a model that will approximate the true distribution of class labels \(p(\ygt \mid \inimage)\) for a given input solely from the dataset of noisy annotations.

   The segmentation architecture combines a U-Net \cite{ronneberger2015unet} based segmentation network with a confusion matrix network to handle noisy labels from multiple annotators. It uses multiple encoder-decoder blocks with skip connections, allowing it to capture both fine and coarse details of the input, while the confusion matrix layers model the noise in the annotations. This is a flexible, highly customizable architecture designed for segmentation tasks, especially in scenarios where annotator label noise is a concern. For comparison with other baselines methods (described in Appendix \ref{sec:appendix_segmentation_baselines_description}, we used the same segmentation network.

   \subsection{Probabilistic Model for Noisy Observations and Architecture}
  In this section, we build a probabilistic noise model for observed noisy annotations.
  Following~\cite{Tanno2019, DisentanglingHumanError}, we assume statistical independence of annotators and independence of annotations of different pixels for a given input image \(\inimage\).

  For the following derivations, we will omit the object index \(i\) to simplify the notation.
  Taking these assumptions into account, we can write the joint likelihood for all annotations \(\{\ynoisyni\}_{r \in \annsetni}\) of a given input image \(\inimageni\):
  \begin{align}
    p (\{\ynoisyni\}_{r \in \annsetni} \mid \inimageni ) &= \textstyle{\prod}_{{r \in \annsetni}} p(\ynoisyni \mid \inimageni) 
    \\
    &= \textstyle{\prod}_{r \in \annsetni} \textstyle{\prod}_{w = 1}^W \textstyle{\prod}_{h = 1}^H p(\ynoisyni_{wh} \mid \inimageni),
  \nonumber
  \end{align}
  where \(\ynoisyni_{wh}\) is a class label of a pixel in a specific location.

  Each of the probabilities \(p(\ynoisyni_{wh} \mid \inimageni)\) inside the product is an implicit marginalization of the latent ground truth class label \(\ygtni_{wh}\):
  \begin{equation}
    p(\ynoisyni_{wh} \mid \inimageni) = \textstyle{\sum}_{l = 1}^L  p(\ygtni_{wh} = l \mid \inimageni) \, p(\ynoisyni_{wh} \mid \ygtni_{wh} = l, \inimageni).
  \label{eq:noisy_marginal_seg}
  \end{equation}
  In this equation, we see that the predictive distribution for a noisy annotation is a mixture of class-conditional distributions of noisy annotations weighed by the ground truth label distributions.
  These noisy class conditional distributions can be considered as elements of a confusion matrix (CM), effectively modeling the noise introduced by the annotator.

  Inspired by~\cite{DisentanglingHumanError, Tanno2019}, we use a two-component model to approximate the densities in equation~\eqref{eq:noisy_marginal}. The first component is the segmentation network, and the other is an annotation network.
  The segmentation network is a convolutional neural network parameterized by \(\theta\) that aims to approximate the ground truth probability distribution \(p(\ygtni_{wh} \mid \inimageni)\). 
  Hence, it transforms an input image \(\inimageni\) to a probability map \(\segnetni \in \R^{W \times H \times L}\), where each \((wh)\)-th element approximates \(p(\ygtni_{wh} \mid \inimageni)\). 

  The annotation network, parameterized by \(\psi\), provides pixel-wise estimates of the CMs of the corresponding annotators as a function of the input image.
  Hence, this model, trained separately for each of the annotators, produces the tensor \(\annetni \in [0,1]^{W \times H \times L \times L}\), \(r \in \annsetni\). The element of this tensor, indexed by \(whkl\), equals to \(p(\ynoisyni_{wh} = k \mid \ygtni_{wh} = l, \inimageni)\).

  The multiplication of these two terms $\hat{p}^{r}_{\theta, \psi}(x) = \annetni \times \segnetni$ resulting in the tensor of shape \(W \times H \times L\), is the $r$-th annotator's probability segmentation map which is optimized in the loss function (the details are covered in Section~\ref{sec:loss}). The architecture of the resulting model is depicted in Figure~\ref{fig:Our_segmentation_NN}.


\section{Baselines description}

\subsection{Classification Task Baselines}
\label{sec:appendix_classification_baselines_description}
  In this section, we discuss baselines used in comparison in \textbf{classification} problems.
  \begin{itemize}[nosep,topsep=0pt]
    \item Co-teaching~\cite{Co-teaching}, which simultaneously trains two DNN models, with each network selecting the batch of data for the other, based on the instances with a small loss.
    
    \item Co-teaching$+$~\cite{coteachingplus}, also employs samples with small loss, but with disagreement about predictions. This is the selection criteria for the networks to pick data for each other.

    \item JoCoR~\cite{JoCoR}, extends the idea of~\cite{Co-teaching,coteachingplus} by using co-regularization to minimize the diversity of the two networks, thus bringing the predictions of the two networks closer together.

    \item Robust Early-learning (CDR; \cite{CDR}), categorizes the critical and non-critical parameters for clean and noisy label fitting, respectively. Different update rules are applied to update these parameters.

    \item Annotator Confusion (Trace; \cite{Tanno2019}) is a regularized approach that assumes the existence of various annotators to simultaneously learn the individual annotator model and the underlying true label distribution, using only noisy observations.
  \end{itemize}
\subsection{Segmentation Task Baselines}
\label{sec:appendix_segmentation_baselines_description}
  In this section, we discuss baselines used in comparison in \textbf{segmentation} problems.
  \begin{itemize}[nosep,topsep=0pt]
    \item CM-NET~\cite{DisentanglingHumanError} uses two coupled CNN models that jointly estimate the true segmentation label distribution, as well as the reliability/confusion of the annotators from noisy labels. The strategy helps with disentangling human bias from the GT.

    \item PU-NET~\cite{kohl2018probabilistic} incorporates a probabilistic model at each pixel in the output segmentation map. Instead of predicting a single label for each pixel, the network predicts a probability distribution over all possible labels.

    \item MR-NET~\cite{MRNet} proposes a method that uses a deep learning model to learn from multiple annotations of the same image and generates a consensus segmentation that is calibrated to match the annotations.
  \end{itemize}


\section{Datasets} 

\subsection{Classification Datasets}
\label{sec:appendix_ClassificationDatasets}
  In this section, we describe the data and types of noises we use to train our model for the classification task.
  \paragraph{MNIST.} The dataset comprises 60,000 samples for training and 10,000 data samples reserved for testing. The number of classes in the dataset is 10. 

  \paragraph{CIFAR-10.} The CIFAR-10 dataset contains 60,000 color images in 10 classifications, with 6000 images in each class. There are 50,000 training and 10,000 test images. The dataset is divided into five training batches and one test batch, each containing 10,000 images. The test batch is a collection of exactly 1,000 data samples randomly selected from each class. The training batches comprise the remaining images in a random order, however, certain training batches may likely have more images from one class than another. 

  \paragraph{CIFAR-10N.} CIFAR-10N is prepared by augmenting the training datasets of CIFAR-10 using Amazon Mechanical Turk-collected human-annotated real-world noisy labels. It has the same number of classes and training and test data points as the CIFAR-10 dataset.

  \paragraph{FMNIST.} Fashion-MNIST is a dataset of article images from Zalando. It consists of a training set of 60,000 instances and a test set of 10,000 instances. Each instance is a $28\times 28$ grayscale image with a label from one of ten classes. 
\subsection{Segmentation Datasets} 
\label{sec:appendix_SegmentationDatasets}
  In this section, we describe and illustrate the data we use to train our model for the segmentation task. 
  
\begin{figure*}[!t]
    \centering
    \begin{tabular}{c c c c c }
      Input & Thin & Thick & Fractured 
      & GT \\
      \includegraphics[width=0.17\textwidth]{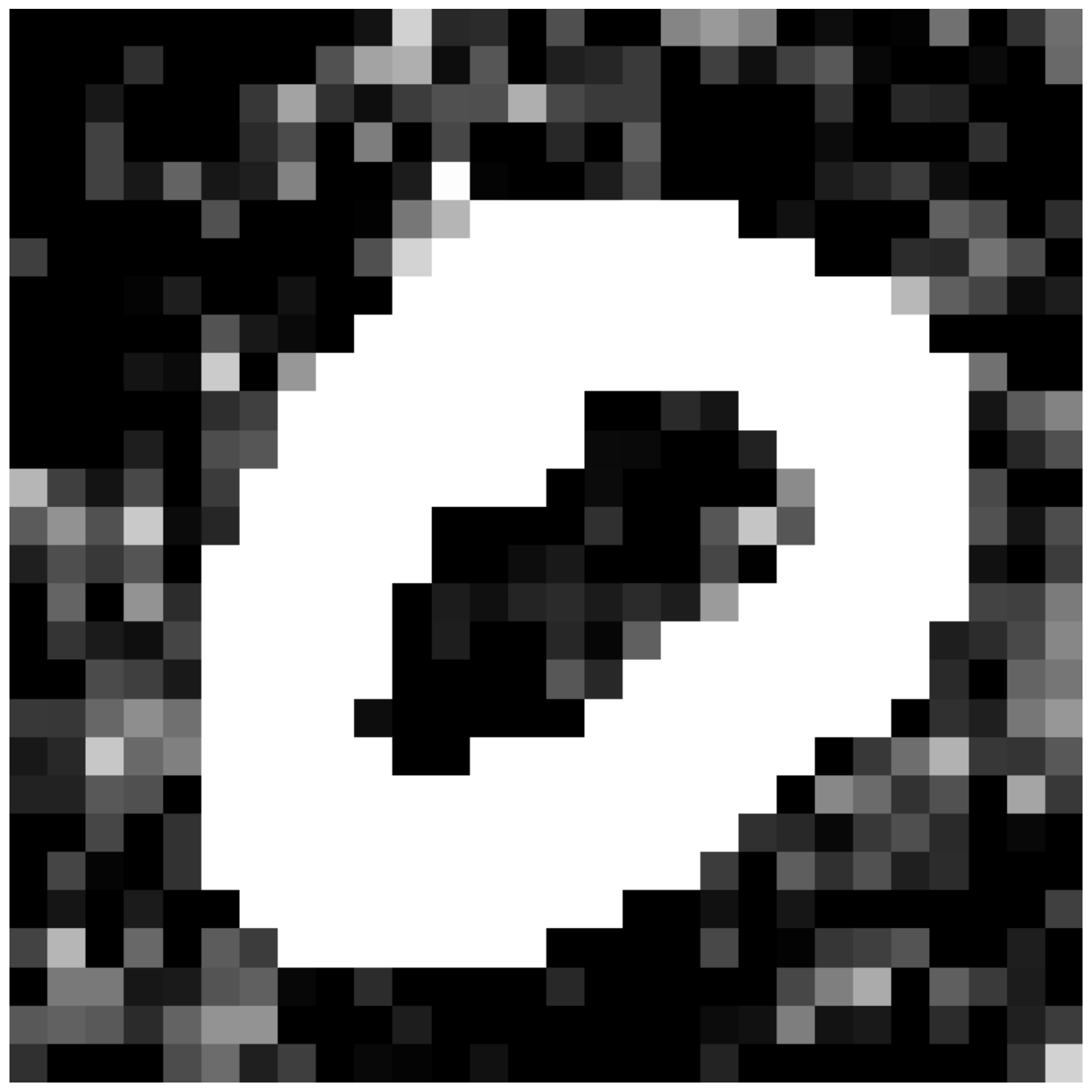} & 
      \includegraphics[width=0.17\textwidth]{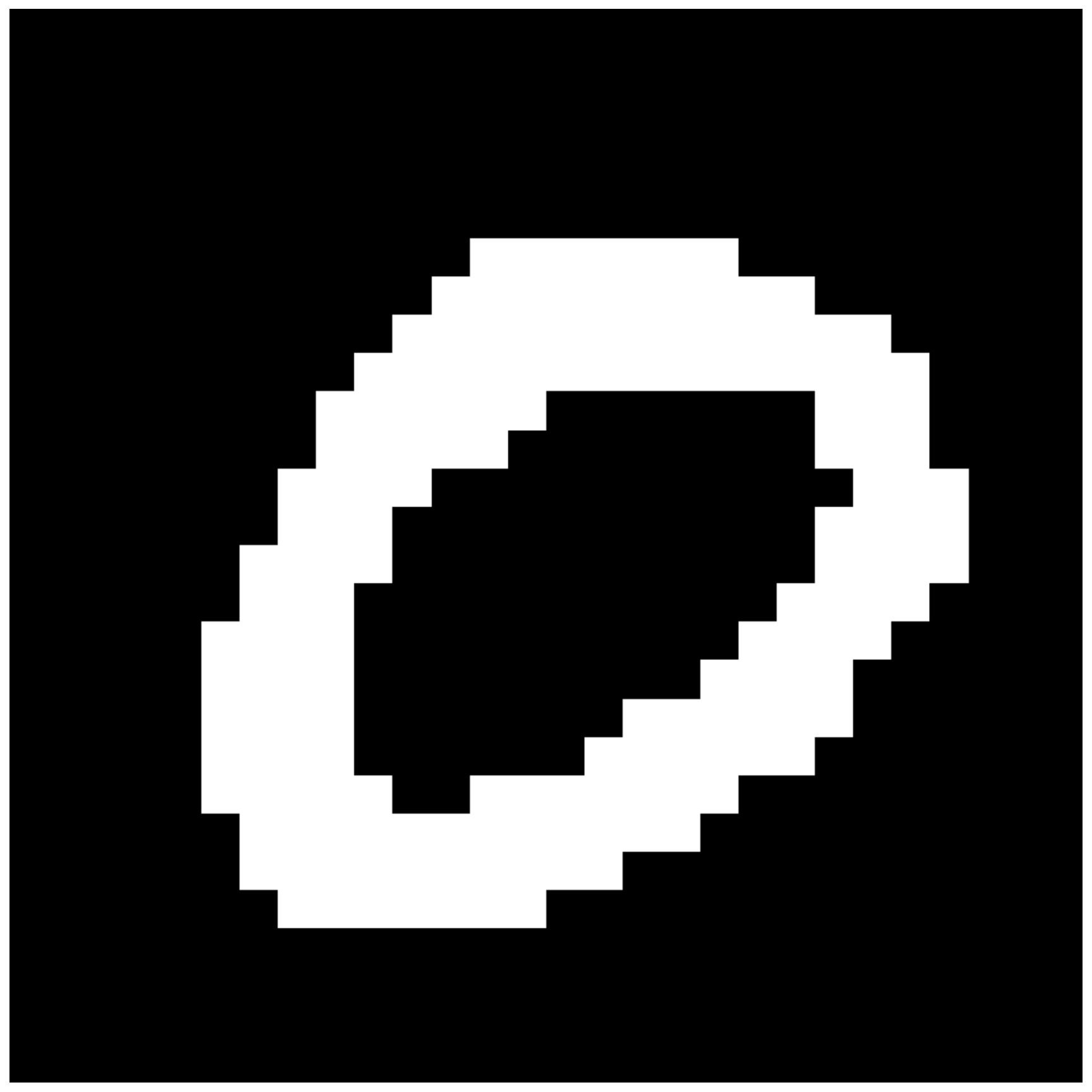} &
      \includegraphics[width=0.17\textwidth]{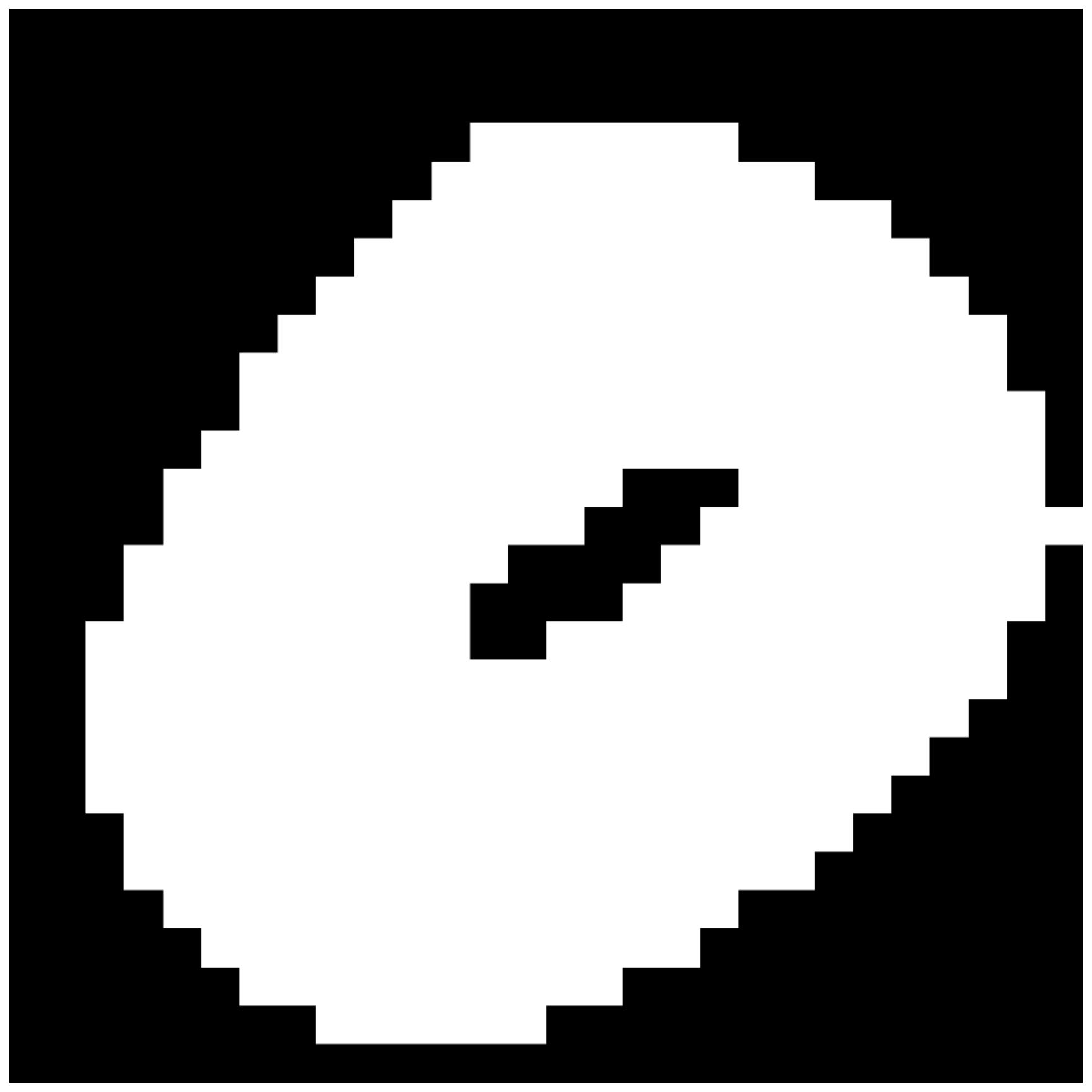} &
      \includegraphics[width=0.17\textwidth]{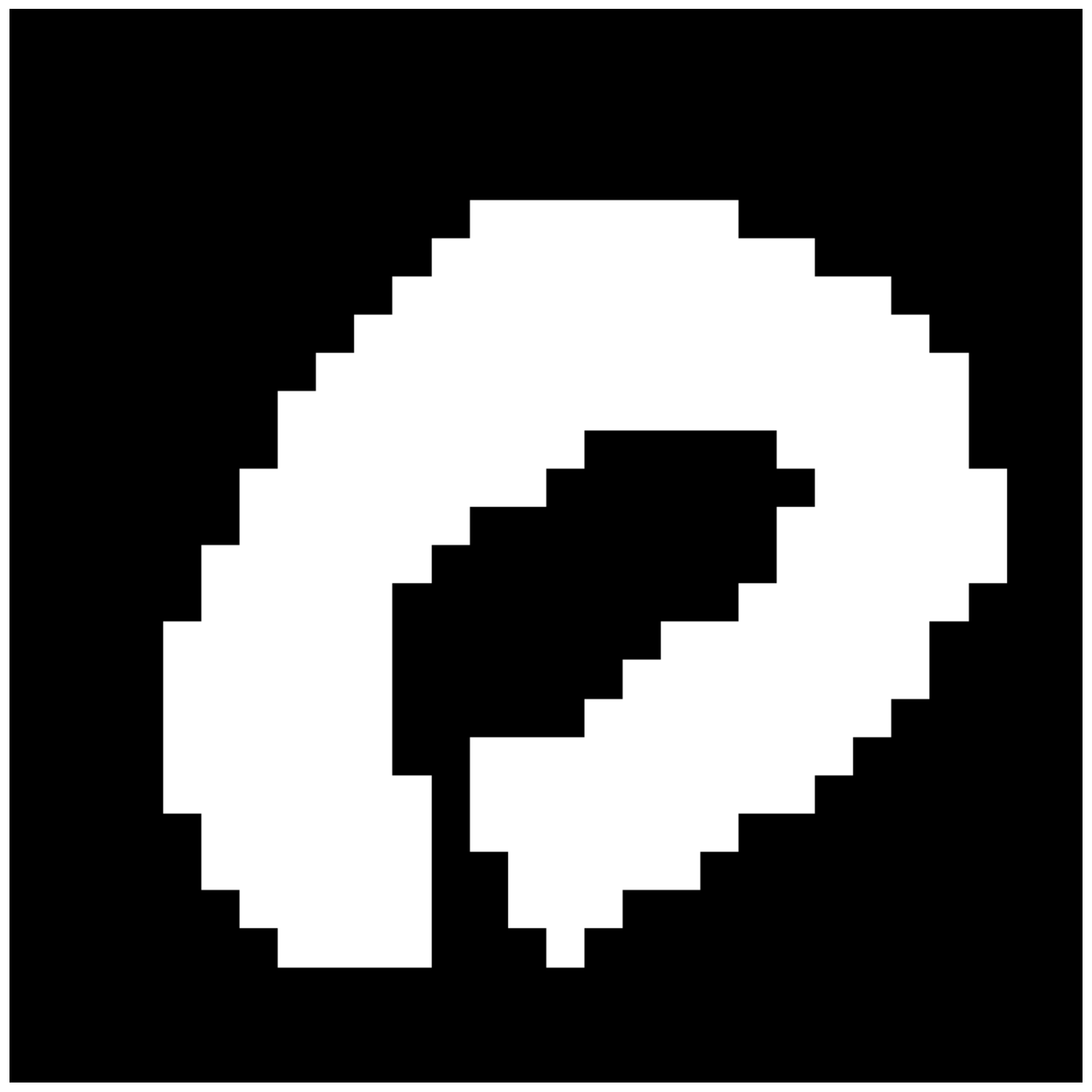} &
      \includegraphics[width=0.17\textwidth]{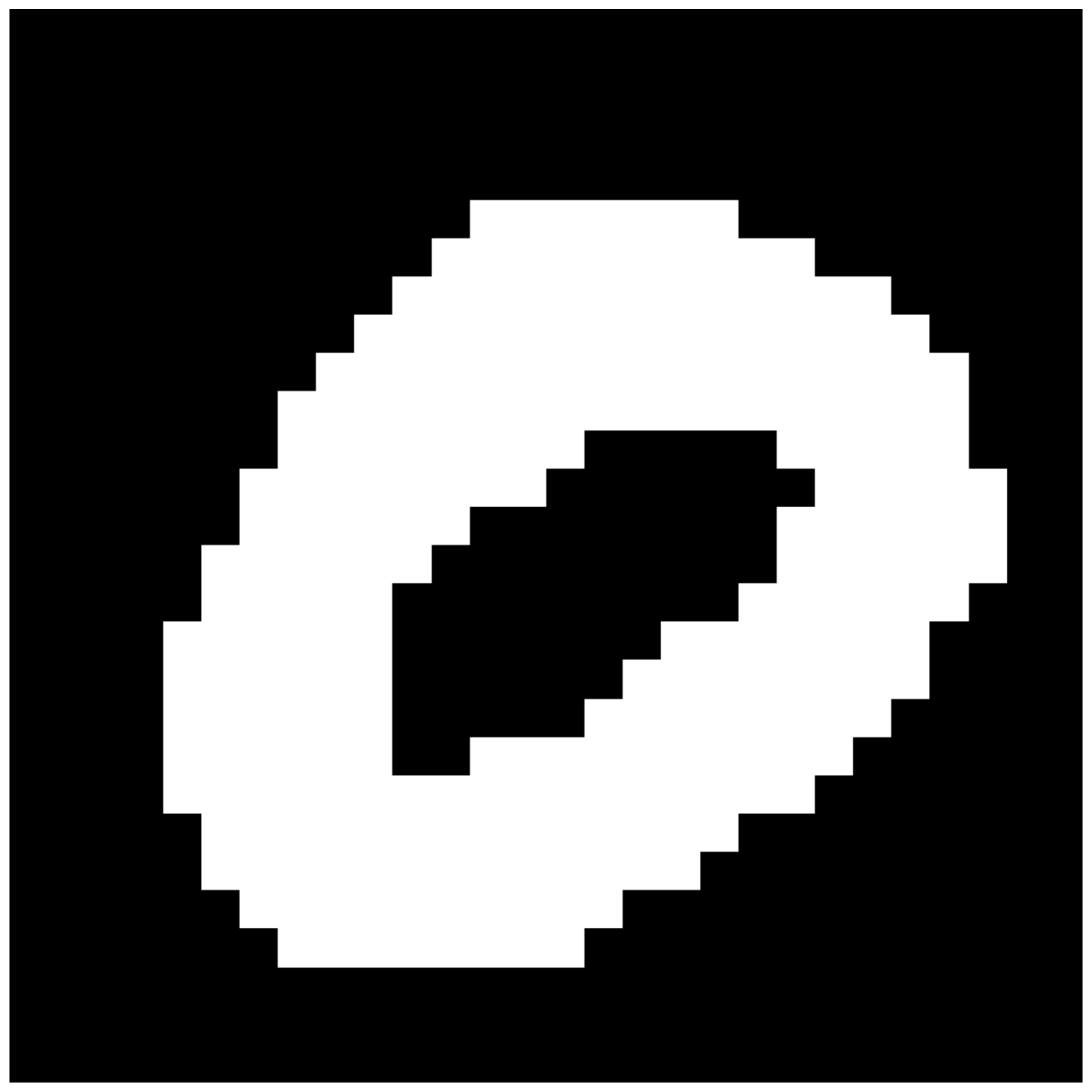} \\

      \includegraphics[width=0.17\textwidth]{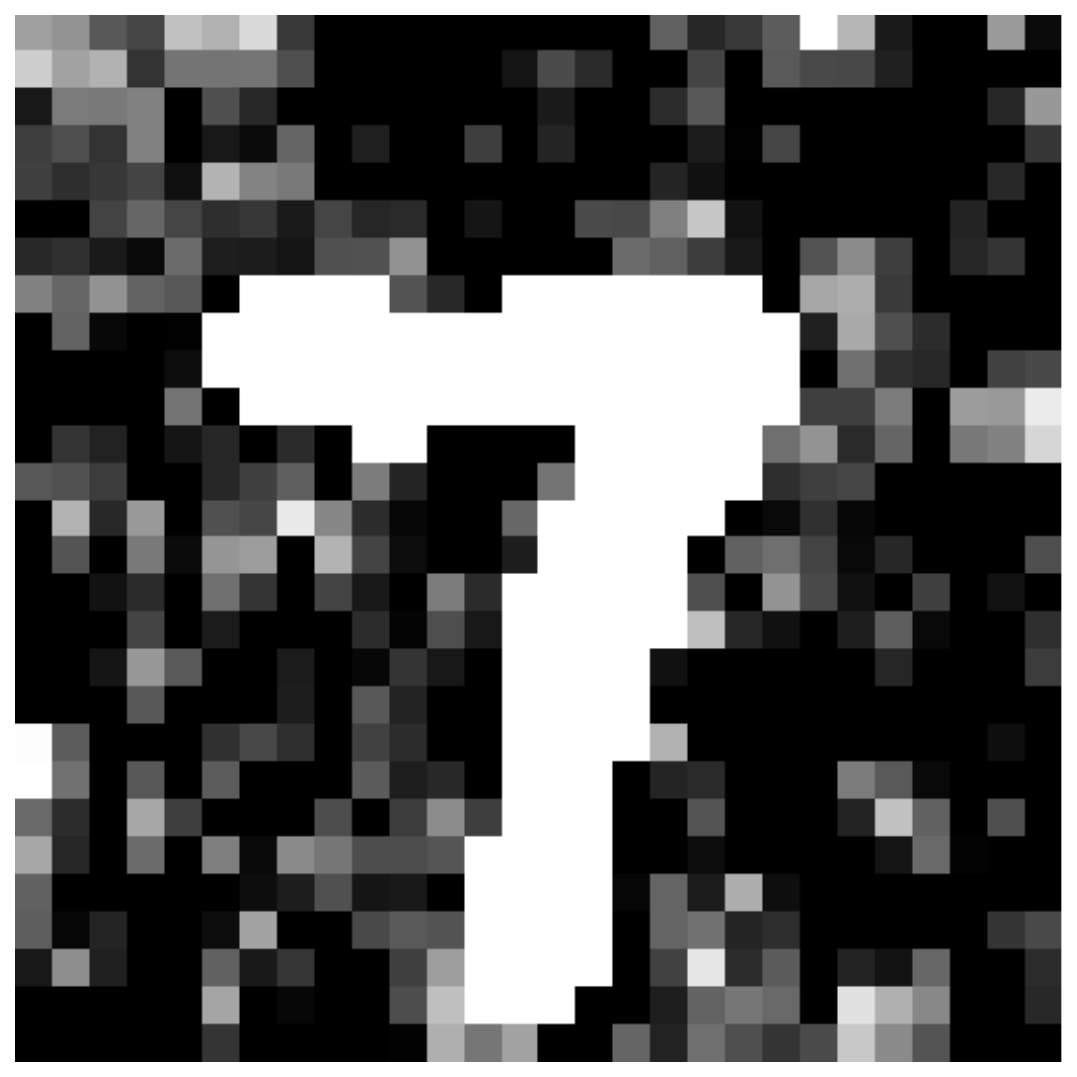} & 
      \includegraphics[width=0.17\textwidth]{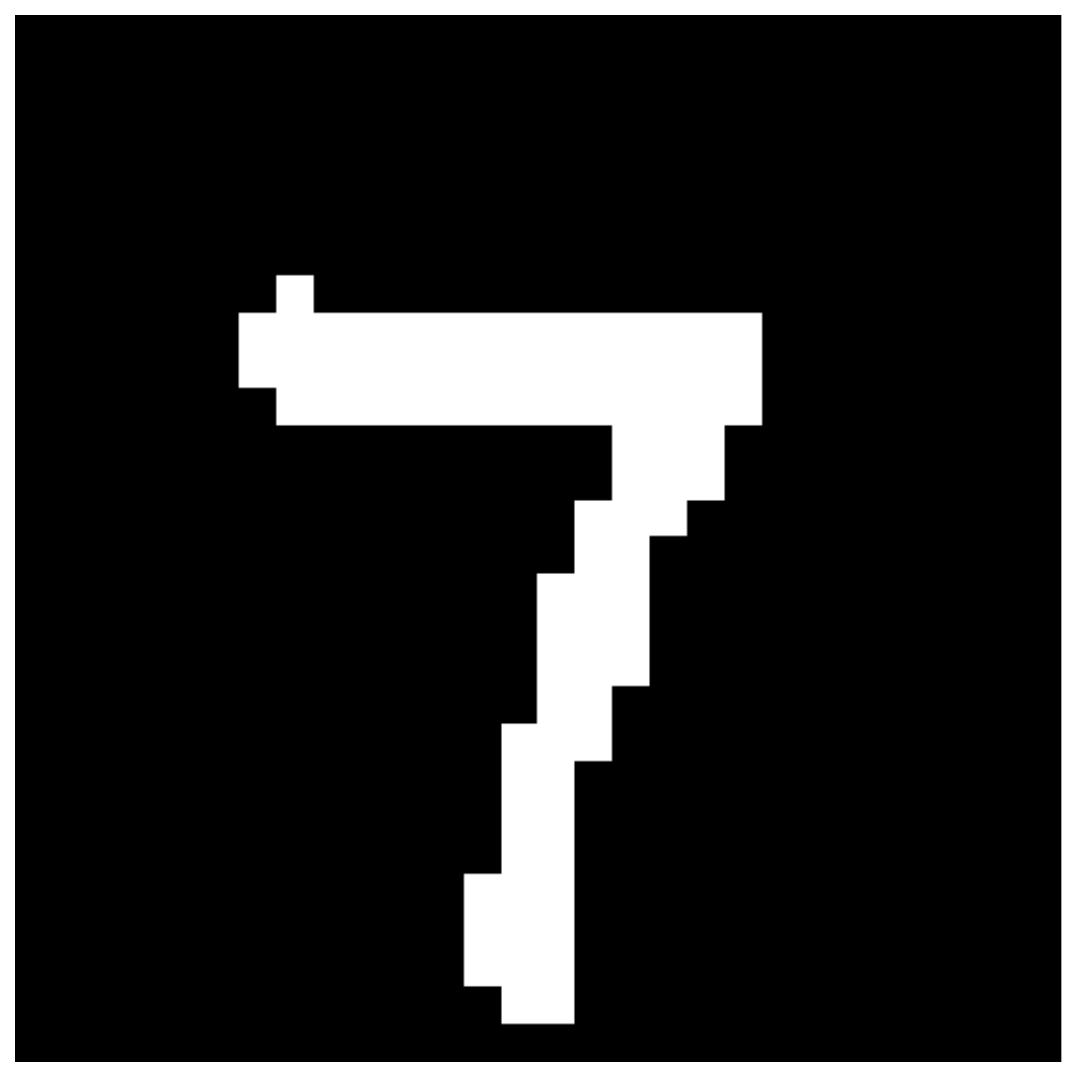} &
      \includegraphics[width=0.17\textwidth]{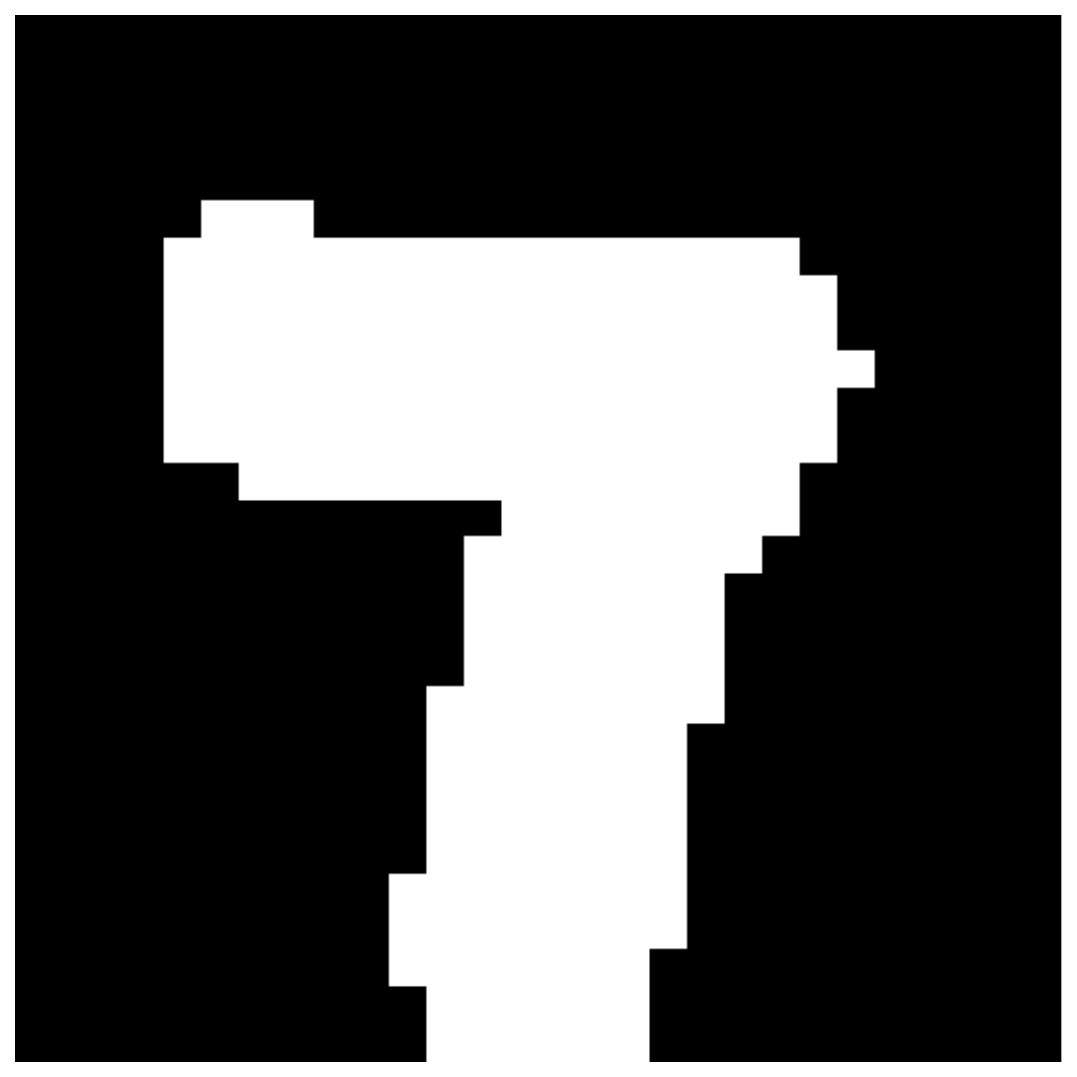} &
      \includegraphics[width=0.17\textwidth]{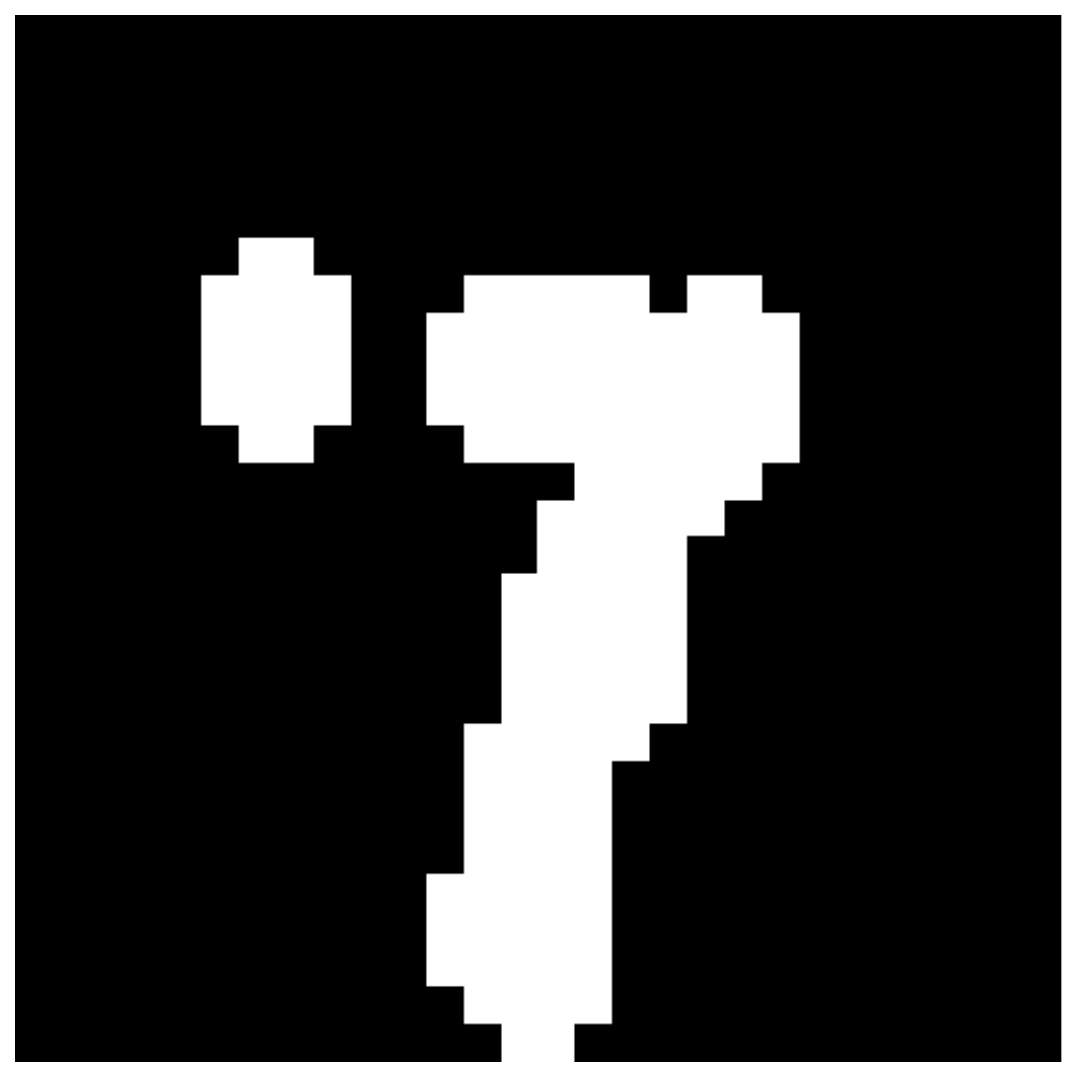} &
      \includegraphics[width=0.17\textwidth]{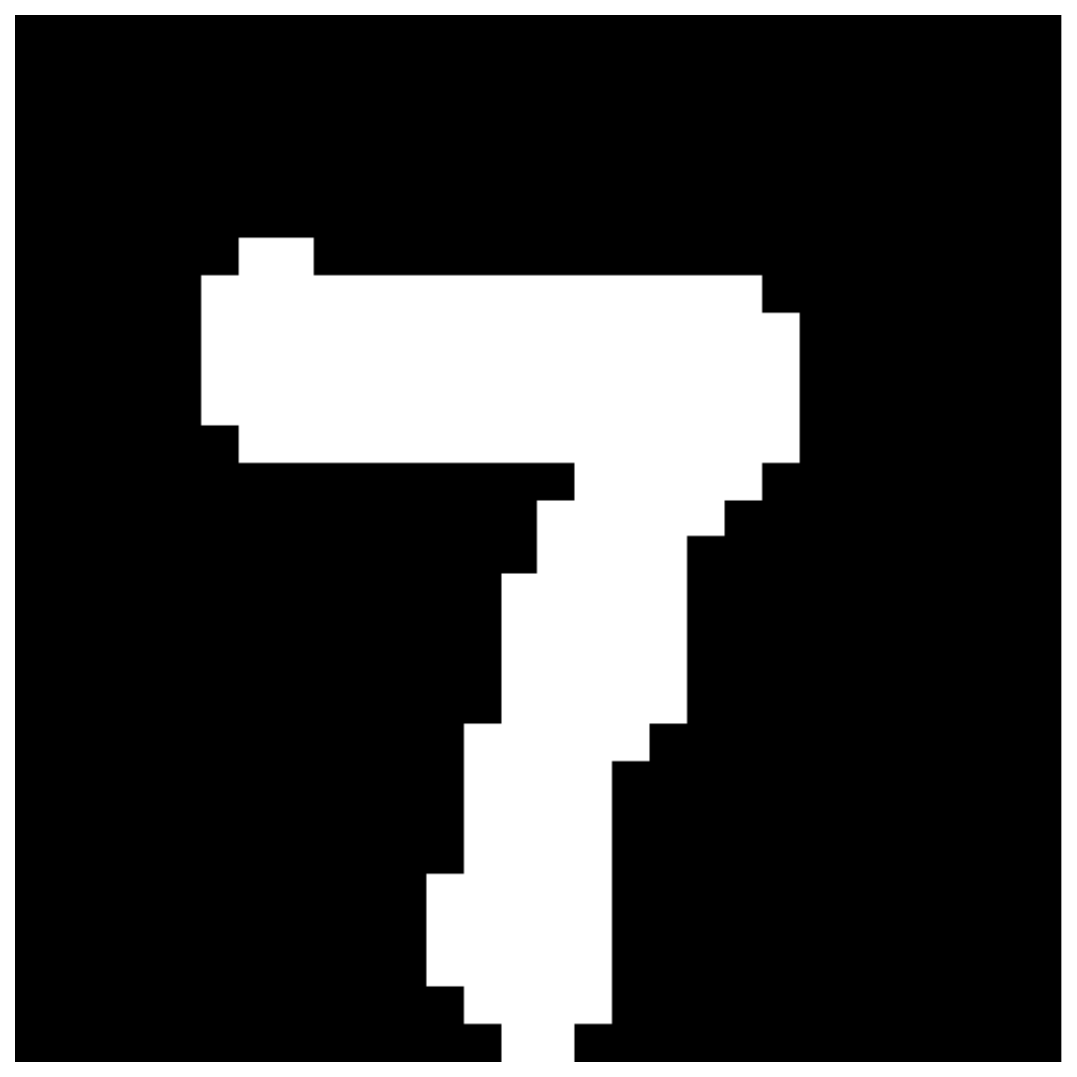} \\

      \includegraphics[width=0.17\textwidth]{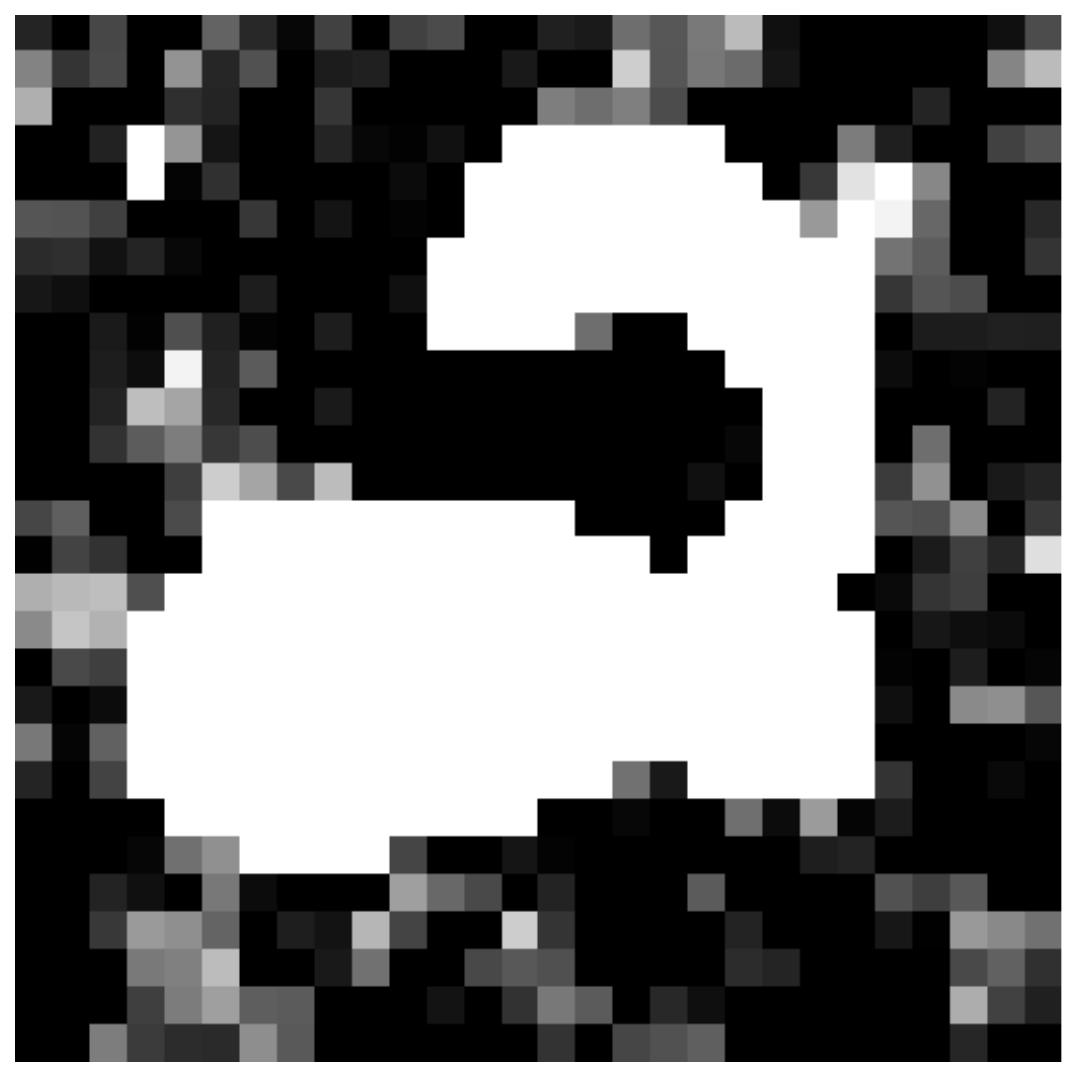} & 
      \includegraphics[width=0.17\textwidth]{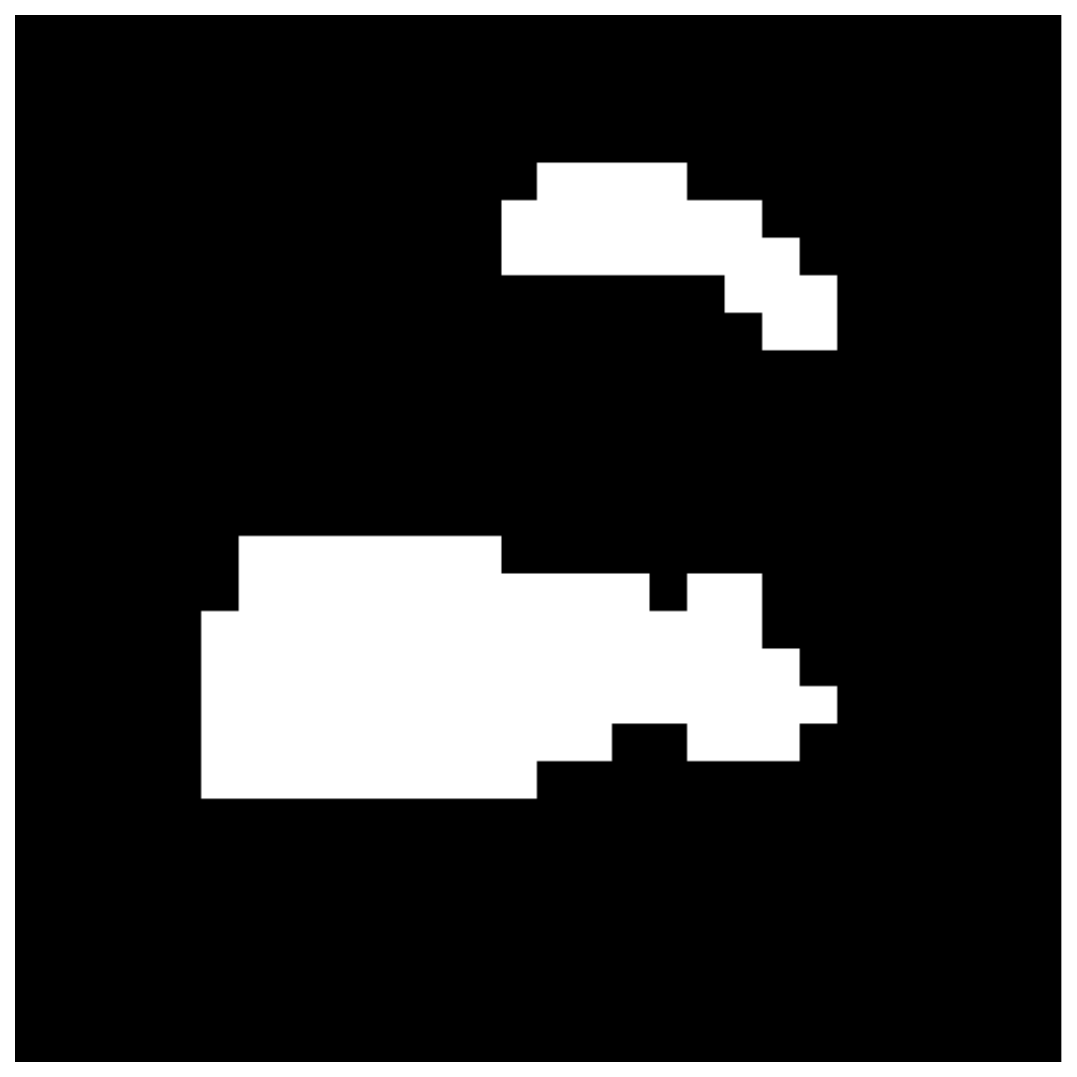} &
      \includegraphics[width=0.17\textwidth]{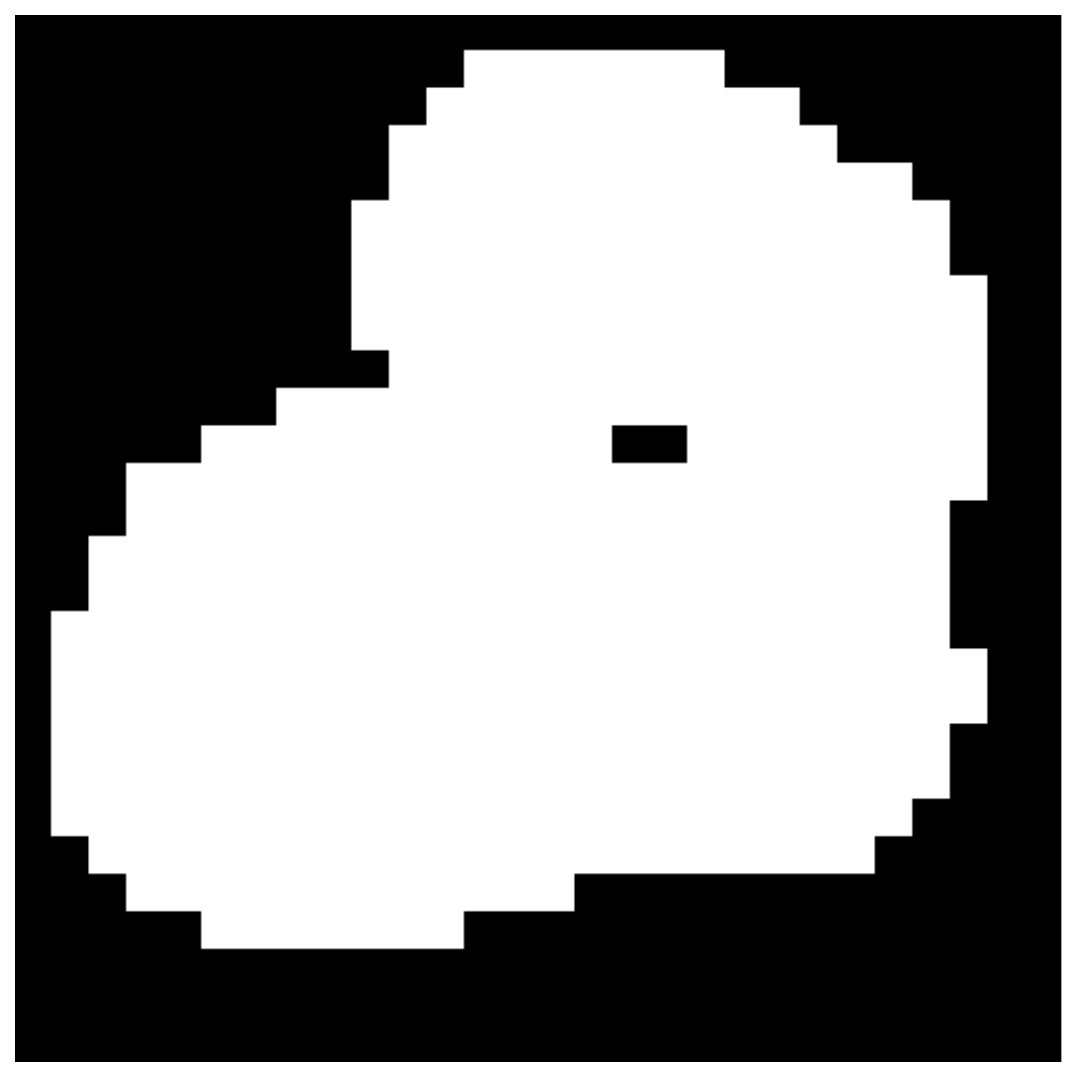} &
      \includegraphics[width=0.17\textwidth]{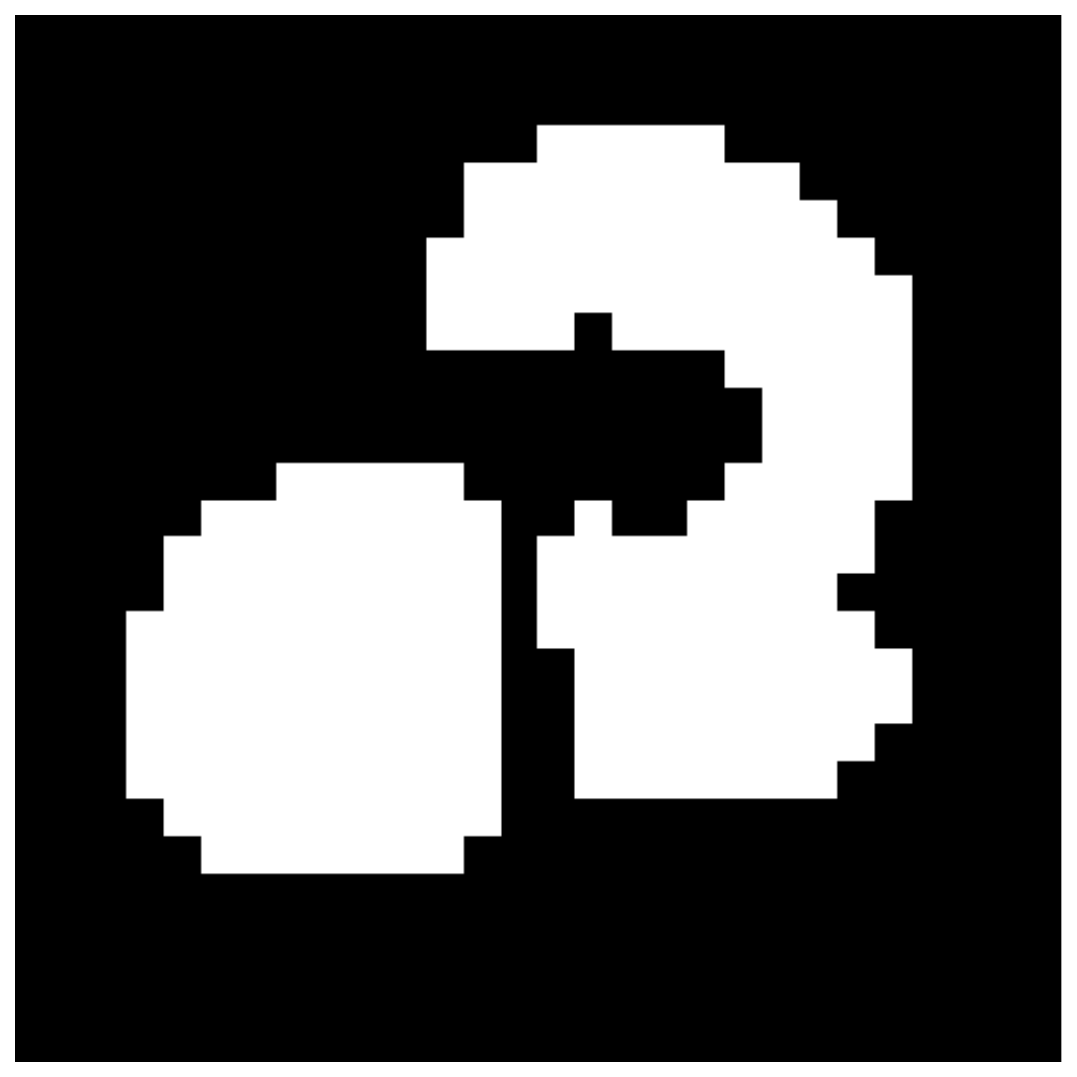} &
      \includegraphics[width=0.17\textwidth]{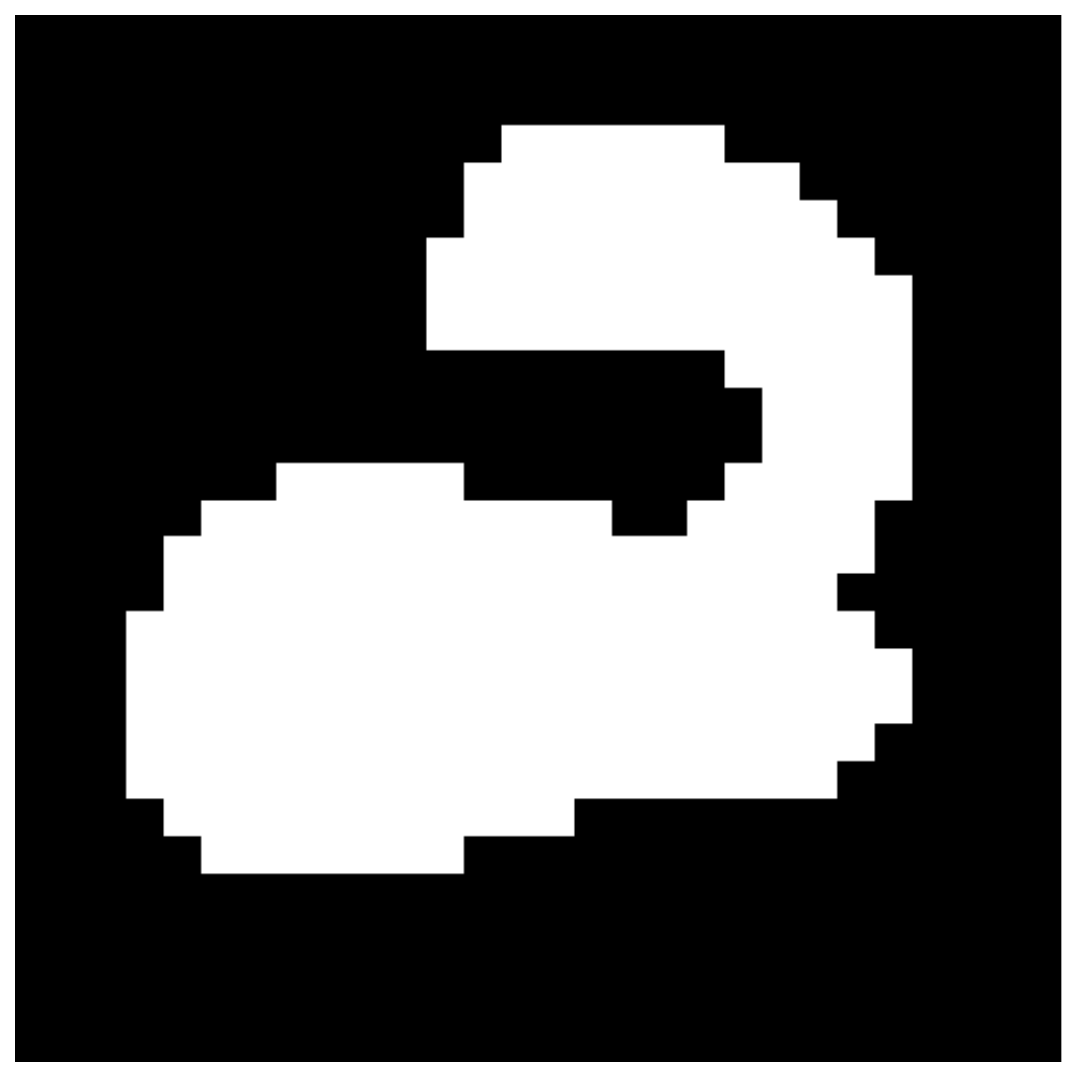}
    \end{tabular}
    \caption{Visualization of the MNIST dataset input image with Gaussian noise, three annotations (thin, thick, fracture), and ground truth.}
  \label{fig:MNIST_data_examples}
  \end{figure*}

  \paragraph{MNIST.} It is the database of handwritten digits that is used as an example of a dataset with artificial noises. This dataset is used to show the performance of the model in cases where ground truth is known. Synthetic annotations are created by applying morphological transformation simulations on assumed ground truth~\cite{castro2019morphomnist}.
  The types of annotators from the simulation are thin (under-segmentation), thick (over-segmentation), fracture (fracture-segmentation), and good (ground truth). For our purpose, we only use the first three annotations. In addition, Gaussian noise was added to the original images to use them as inputs for the model.  There are 60,000 samples for training and 10,000 data samples reserved for testing. In Figure~\ref{fig:MNIST_data_examples}, we show different augmentations for the MNIST dataset, artificially creating several annotators.

\begin{figure*}[!htb]
    \centering
    \begin{tabular}{c c c c c}
      Image & Annotation 1 & Annotation 2 & Annotation 3 & Annotation 4 \\
      \includegraphics[width=0.18\textwidth]{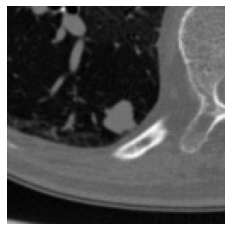} & 
      \includegraphics[width=0.18\textwidth]{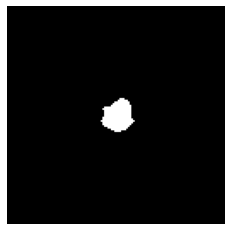} & 
      \includegraphics[width=0.18\textwidth]{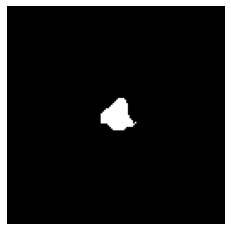} & 
      \includegraphics[width=0.18\textwidth]{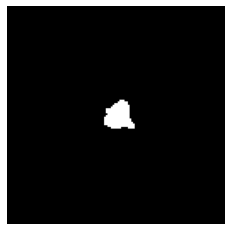} & 
      \includegraphics[width=0.18\textwidth]{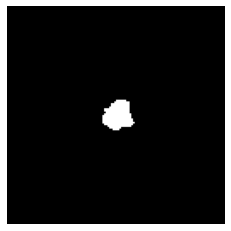} \\

      \includegraphics[width=0.18\textwidth]{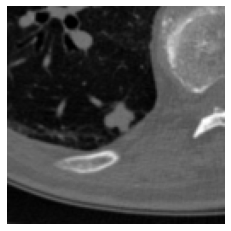} & 
      \includegraphics[width=0.18\textwidth]{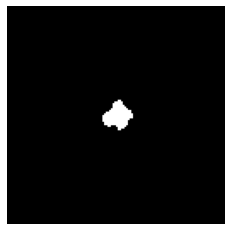} & 
      \includegraphics[width=0.18\textwidth]{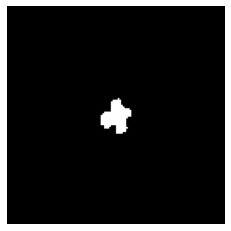} & 
      \includegraphics[width=0.18\textwidth]{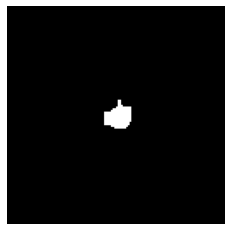} & 
      \includegraphics[width=0.18\textwidth]{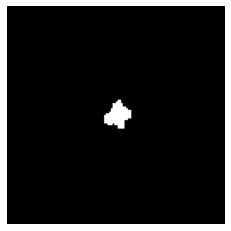} \\

      \includegraphics[width=0.18\textwidth]{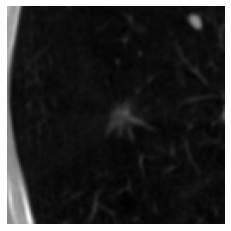} & 
      \includegraphics[width=0.18\textwidth]{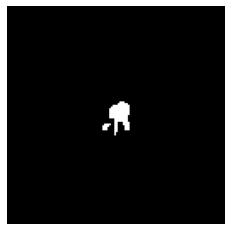} & 
      \includegraphics[width=0.18\textwidth]{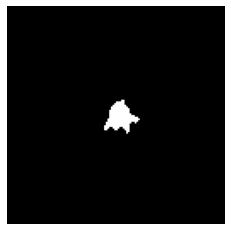} & 
      \includegraphics[width=0.18\textwidth]{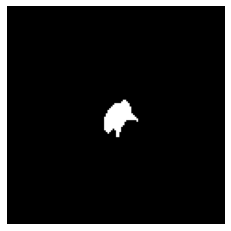} & 
      \includegraphics[width=0.18\textwidth]{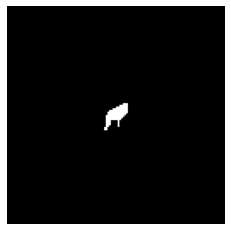} \\

    \end{tabular}
    \caption{Visualization of LIDC dataset input image and its all annotations.}
  \label{fig:lidc_samples}
  \end{figure*}

  \paragraph{LIDC-IDRI.} The lung image database consortium (LIDC) and image database resource initiative (IDRI) is a completed reference database of lung nodules on CT scans~\cite{Armato}. It consists of 1,018 3D thorax CT scans with 4 annotations from four radiologists. For pre-processing, we used the same method in~\cite{kohl2018probabilistic}, which provides 15,096 2D slices, and used the 60-20-20 dataset split as done in~\cite{PHiSeg2019Baumgartner}. In Figure~\ref{fig:lidc_samples}, we illustrate the input image with annotations, as well as our predicted results for the LIDC dataset.

\begin{figure*}[!htb]
    \centering
    \begin{tabular}{c c c c }
      Input fundus image & Annotation 2 & Optic cup annotations & Optic disc annotations \\
      \includegraphics[width=0.23\textwidth]{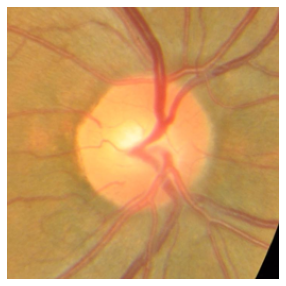} & 
      \includegraphics[width=0.23\textwidth]{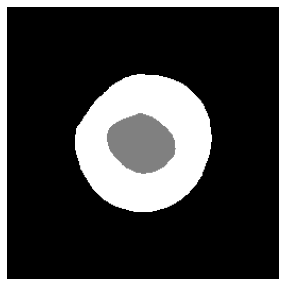} &
      \includegraphics[width=0.23\textwidth]{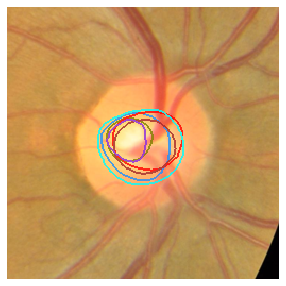} &
      \includegraphics[width=0.23\textwidth]{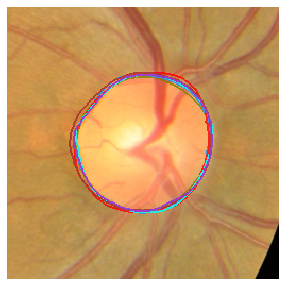} \\

      \includegraphics[width=0.23\textwidth]{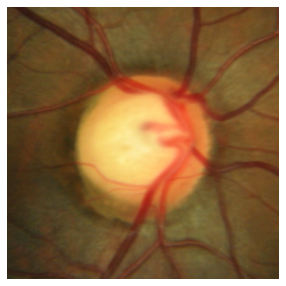} & 
      \includegraphics[width=0.23\textwidth]{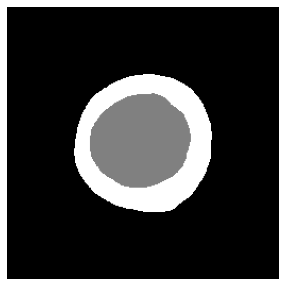} &
      \includegraphics[width=0.23\textwidth]{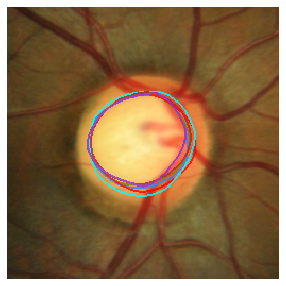} &
      \includegraphics[width=0.23\textwidth]{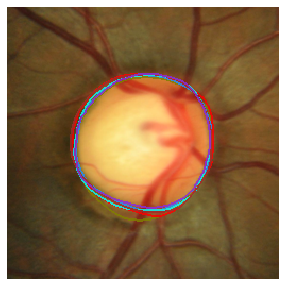}\\

      \includegraphics[width=0.23\textwidth]{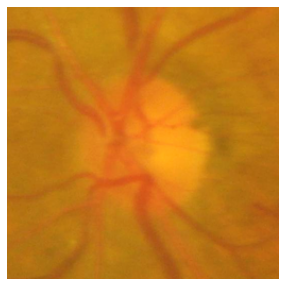} & 
      \includegraphics[width=0.23\textwidth]{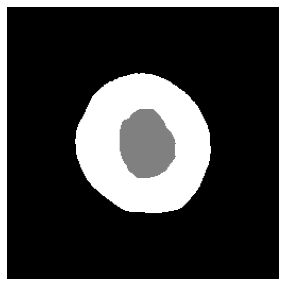} &
      \includegraphics[width=0.23\textwidth]{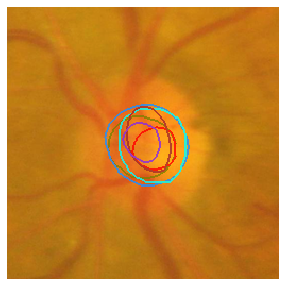} &
      \includegraphics[width=0.23\textwidth]{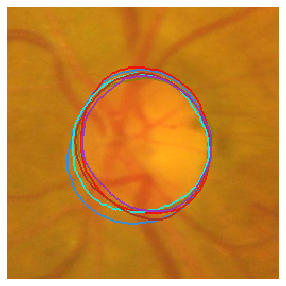}
    \end{tabular}
    \caption{Visualization of the RIGA dataset from left to right: input image, one of the annotations, boundaries of all six annotations for cup, and boundaries of all six annotations for the disc.}
  \label{fig:RIGA_disc_cup_examples}
  \end{figure*}

  \paragraph{RIGA.} Retinal fundus images for glaucoma analysis (RIGA) dataset contains 750  original images and 4,500 annotations and was collected from three sources: BinRushed (195 images), MESSIDOR (460 images), and Magrabia (95 images)~\cite{Almazroa2017AgreementAO}. 4,500 annotations came from six experienced ophthalmologists, who manually labeled the optic cup/disc boundaries for each image. For model training, we used samples from BinRushed and MESSIDOR and used samples from Magrabia as a test set. For the image preprocessing,  we followed the settings in~\cite{MRNet} by resizing to \(256 \times 256\), and then we normalized each pixel by subtracting the mean and dividing by the standard deviation, which is calculated on training data. In Figure~\ref{fig:RIGA_disc_cup_examples}, we illustrate the input image, annotator masks, and boundaries of different annotators for the RIGA dataset.

\subsubsection{Types of Noise} 
\begin{figure*}[!htb]
    \centering
    \vspace*{10pt}
    \begin{tabular}{c c }
      \includegraphics[width=0.51\textwidth]{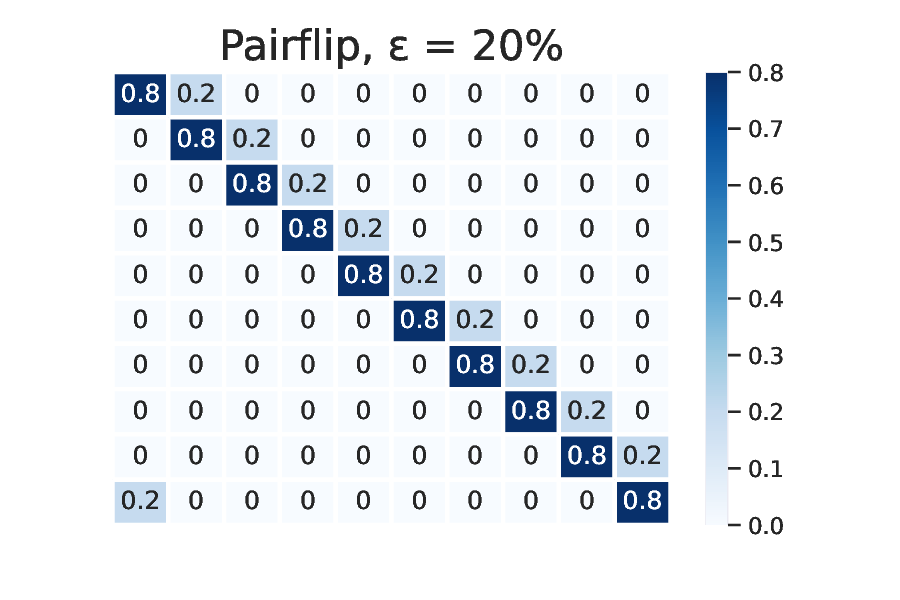} & 
      \includegraphics[width=0.51\textwidth]{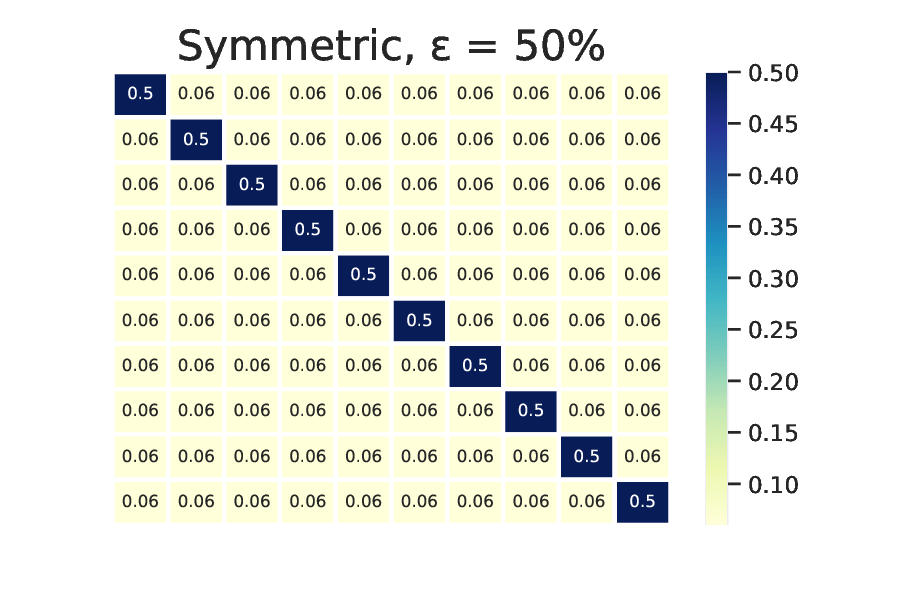}
    \end{tabular}
    \caption{Noise Transition matrices for Pairflip and Symmetric noise.}
  \label{fig:noise_transition_matrix}
  \end{figure*}

\begin{figure*}[!htb]
    \centering
    \begin{tabular}{c c }
      \includegraphics[width=0.485\textwidth]{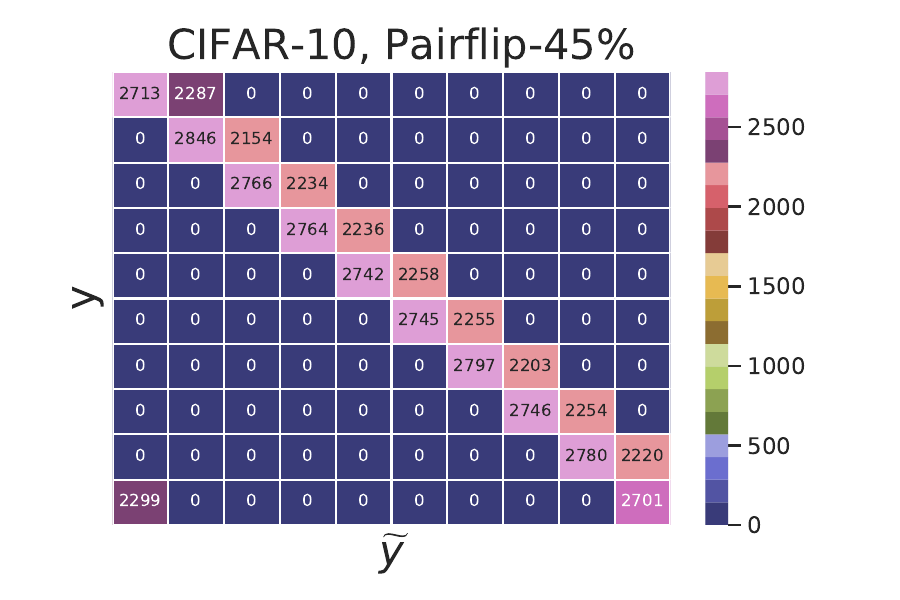} & 
      \includegraphics[width=0.485\textwidth]{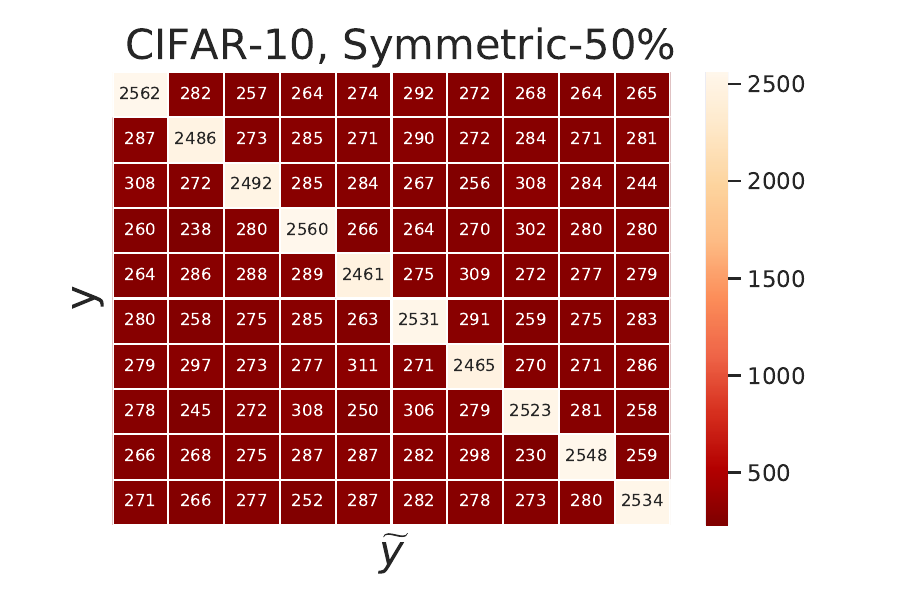}
    \end{tabular}
    \caption{Confusion matrix between clean (y) and noisy labels ($\tilde{y}$) of CIFAR-10 dataset for (a) Pairflip-45$\%$ and (b) Symmetric-50$\%$ noise.}
  \label{fig:noise_transition_matrix_TrueVsConfused_labels}
  \end{figure*}

  We simulated the noise in our datasets by confusing each class with another class and assigning the probability proportion for corrupting the labels. Figure~\ref{fig:noise_transition_matrix} shows the examples of noise transition matrices for pairflip 20$\%$ and symmetric 50$\%$ noise types. In addition, Figure~\ref{fig:noise_transition_matrix_TrueVsConfused_labels} signifies the noise labels distributions for the CIFAR-10 dataset for pairflip 45$\%$ and symmetric 50$\%$ noise types; this distribution of the label noise is used in the training process. 

\section{Supplementary Results}
  In this section, we provide supplementary visualizations for trained segmentation models. In the following subsections, we discuss the results for each dataset in more detail, computing the Dice coefficients separately for each of the classes.
 
\subsection{Segmentation} 
\label{sec:SegmentationSupplementaryResults}

  Along the Dice coefficient between the prediction \(\segnetni\) and true label \(\ygtni\), we also consider following Dice coefficients:
  \begin{itemize}
    \item $D(y_{a}, y_{a})$ is the average Dice coefficient among the noisy annotations $y_{a}$. This shows how raters are in agreement among themselves. First, the Dice coefficient between $i^{th}$ and $j^{th}$ ($i \neq j$) annotations are calculated, and the average value is taken from all possible combinations. Similarly, the Dice coefficient among predictions of annotations $D(\tilde{y}_{a},\tilde{y}_{a})$ calculated to show the agreement/disagreement level among the predictions. If these two measures are close, we did a good job of predicting annotations.
    \item $D(\tilde{y}_{a}, y_{a})$ is a Dice coefficient between the annotations' predictions and noisy labels. The Dice coefficient between the annotations $y_a^i$ and its corresponding prediction $\tilde{y}_a^i$ ($i=1, \dots, R$) is calculated first and then averaged. This measures how correctly the annotations were modeled.
    \item $D(y_{a},y_{gt})$ is a Dice coefficient between noisy labels and true labels. First, it was calculated for the rater and averaged over all raters. This shows each rater's agreement/disagreement with the true label. Similarly, the Dice coefficient between the annotations' predictions and true label $D(\tilde{y}_{a},y_{gt})$ is calculated. The expectation is that these values will be close. 
    \item $D(\tilde{y},y_{gt})$ if the Dice coefficient between the main output (prediction of the true label) and true label.
  \end{itemize}

  We tuned relative weight hyperparameters such as learning rate, $\lambda, \gamma$ for the segmentation model using an automatic tool (wandb.ai) for the results in the main text and ablation study in Section~\ref{sec:reg_ablation}. Then, we picked the best values in terms of the Dice score between predicted and true labels. 
  
\begin{figure*}[tbh!]
    \centering
    \resizebox{0.9\textwidth}{!}{
    \begin{tabular}{c c c}
      Annotations & Annotations with Prediction & Prediction and Ground Truth \\
      \includegraphics[width=0.3\textwidth]{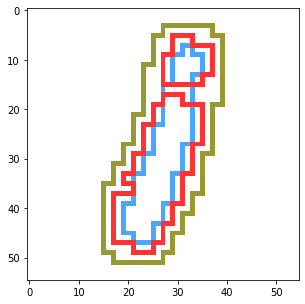} & 
      \includegraphics[width=0.3\textwidth]{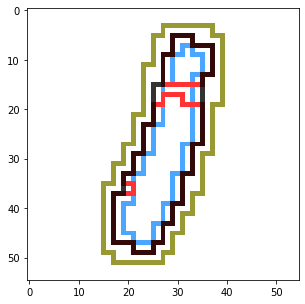} & 
      \includegraphics[width=0.3\textwidth]{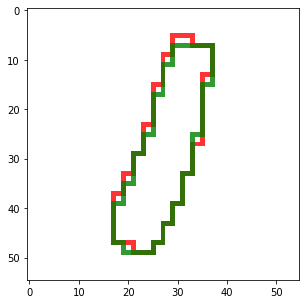} \\

      \includegraphics[width=0.3\textwidth]{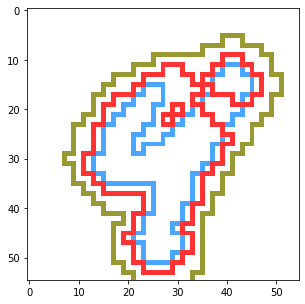} & 
      \includegraphics[width=0.3\textwidth]{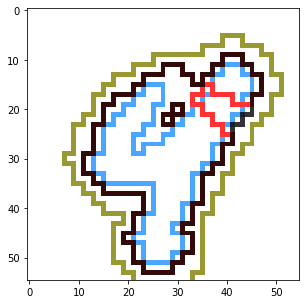} & 
      \includegraphics[width=0.3\textwidth]{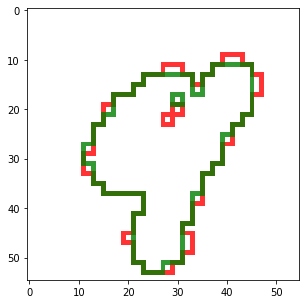} \\

      \includegraphics[width=0.3\textwidth]{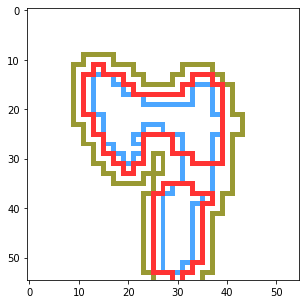} & 
      \includegraphics[width=0.3\textwidth]{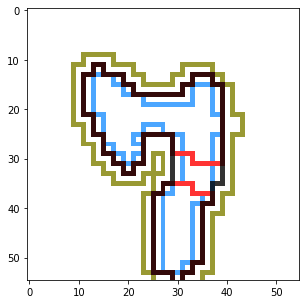} & 
      \includegraphics[width=0.3\textwidth]{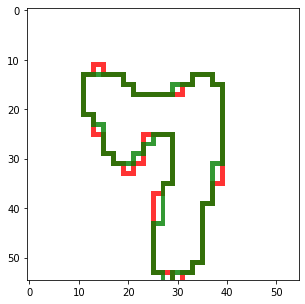} \\

      \includegraphics[width=0.3\textwidth]{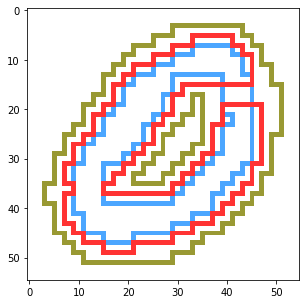} & 
      \includegraphics[width=0.3\textwidth]{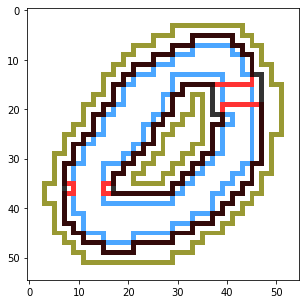} & 
      \includegraphics[width=0.3\textwidth]{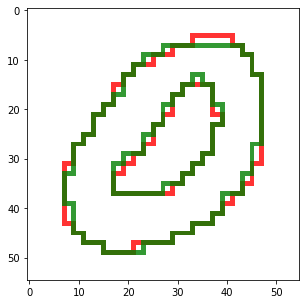} \\
    \end{tabular}
    }
    \caption[Visualization of segmentation results along with annotators(MNIST).]{Visualization of segmentation results along with annotators (MNIST): left: red for fractured, dodgerblue for thin and olive for thick; middle: same colors for annotators and black for prediction; right: red for true label and green for prediction.}
  \label{fig:mnist_final}
  \end{figure*}

  \paragraph{MNIST.} The Dice coefficients for all six classes of the MNIST dataset are depicted in Table~\ref{tab:MNIST_results}. In this context, class 0 denotes the background, while class 1 signifies the numbers. We observe that our approach, augmented with a confidence regularizer, consistently yields the highest or second-best performance in terms of class-wise Dice coefficients when applied to the MNIST dataset. To further illustrate these results, we have included additional visualizations of the final segmentations achieved on the MNIST dataset as depicted in Figure~\ref{fig:mnist_final}.

    \begin{table*}
    \caption{Comparison of segmentation Dice coefficient (\%) (mean $\pm$ standard deviation) for MNIST, computed separately for each class. \textbf{Best} results are in bold, \underline{second-best} underlined. The red column is the most interesting -- the Dice coefficient between approximated and ``real'' ground-truth masks.}
  \label{tab:MNIST_results}
    \centering
    \resizebox{.9\textwidth}{!}{
    \begin{tabular}{@{}ScScScScScScSc Sc} 
      \toprule
      & & $D(\tilde{y}_{a},\tilde{y}_{a})$ & $D(y_{a},y_{a})$ & $D(\tilde{y}_{a},y_{a})$ & $D(\tilde{y}_{a},y_{gt})$ & $D(y_{a},y_{gt})$ & $D(\tilde{y},y_{gt})$\\ \midrule

      \multirow{4}{*}{\rotatebox{90}{\begin{tabular}[c]{@{}c@{}}\textbf{MR-Net}\\ \tiny{\cite{MRNet}}\end{tabular}}} & \multirow{2}{*}{class 0} & 91.08 & 90.90 & 98.34 &  95.03 & 95.34 & \underline{98.20} \\ 
      & & $\pm 0.49$ & $\pm 0.00$ & $\pm 0.04$ & $\pm 0.23$ & $\pm 0.00$ & $\pm 0.19$ \\ \cdashline{2-8}

      & \multirow{2}*{class 1} & 72.48 & 71.80 & 94.15 &  84.16 & 85.25 & \underline{93.43} \\ 
      & & $\pm 0.89$ & $\pm 0.00$ & $\pm 0.11$ &  $\pm 0.60$ & $\pm 0.00$ & $\pm 0.72$ \\ \midrule

      \multirow{4}{*}{\rotatebox{90}{\begin{tabular}[c]{@{}c@{}}\textbf{PU-Net}\\ \tiny{\cite{kohl2018probabilistic}}\end{tabular}}} & \multirow{2}{*}{class 0} & 93.78 & 90.90 & 95.22 &  95.02 & 95.34 & 94.92 \\ 
      & & $\pm 0.43$ & $\pm 0.00$ & $\pm 0.26$ & $\pm 0.43$ & $\pm 0.00$ & $\pm 0.88$ \\ \cdashline{2-8}

      & \multirow{2}*{class 1} & 78.68 & 71.80 & 94.35 &  82.99 & 82.25 & 82.31 \\ 
      & & $\pm 1.14$ & $\pm 0.00$ & $\pm 0.19$ &  $\pm 0.71$ & $\pm 0.00$ & $\pm 0.85$ \\ \midrule

      \multirow{4}{*}{\rotatebox{90}{\begin{tabular}[c]{@{}c@{}}\textbf{CM-Net}\\ \tiny{\cite{DisentanglingHumanError}}\end{tabular}}} & \multirow{2}{*}{class 0} & 91.10 & 90.90 & 98.54 &  95.11 & 95.34 & 97.17 \\ 
      & & $\pm 0.12$ & $\pm 0.00$ & $\pm 0.00$ & $\pm 0.06$ & $\pm 0.00$ & $\pm 0.01$ \\ \cdashline{2-8}

      & \multirow{2}{*}{class 1} & 72.45 & 71.80 & 94.88 &  84.44 & 85.25 & 90.52 \\ 
      & & $\pm 0.04$ & $\pm 0.00$ & $\pm 0.01$ &  $\pm 0.09$ & $\pm 0.00$ & $\pm 0.03$ \\ \midrule

      \multirow{4}{*}{\rotatebox{90}{\begin{tabular}[c]{@{}c@{}}\textbf{Ours} \end{tabular}}} & \multirow{2}{*}{class 0} & 91.00 & 90.90 & 98.47 &  95.05 & 95.34 & \textbf{98.49} \\ 
      & & $\pm 0.10$ & $\pm 0.00$ & $\pm 0.01$ & $\pm 0.05$ & $\pm 0.00$ & $\pm 0.02$ \\ \cdashline{2-8}

      & \multirow{2}{*}{class 1} & 72.30 & 71.80 & 94.69 &  84.32 & 85.25 & \textbf{94.72} \\ 
      & & $\pm 0.31$ & $\pm 0.00$ & $\pm 0.02$ &  $\pm 0.16$ & $\pm 0.00$ & $\pm 0.06$ \\ \midrule

      \multirow{4}{*}{\rotatebox{90}{\begin{tabular}[c]{@{}c@{}}\textbf{Ours (No}\\\textbf{Conf.)}\end{tabular}}} & \multirow{2}{*}{class 0} & 91.01 & 90.90 & 98.47 &  95.05 & 95.34 & 97.39 \\ 
      & & $\pm 0.08$ & $\pm 0.00$ & $\pm 0.01$ & $\pm 0.04$ & $\pm 0.00$ & $\pm 0.01$ \\ \cdashline{2-8}

      & \multirow{2}*{class 1} & 72.29 & 71.80 & 94.69 &  84.31 & 85.25 & 91.56 \\ 
      & & $\pm 0.13$ & $\pm 0.00$ & $\pm 0.02$ &  $\pm 0.16$ & $\pm 0.00$ & $\pm 0.03$ \\ \midrule

    \end{tabular}
    }
  \end{table*}

\begin{figure*}
    \begin{tabular}{c c c c c}
      Annotation 1 & Annotation 2 & Annotation 3 & Annotation 4 & Prediction and GT \\
      \includegraphics[width=0.18\textwidth]{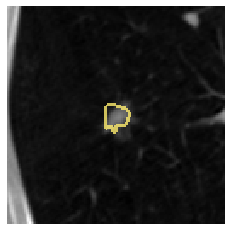} & 
      \includegraphics[width=0.18\textwidth]{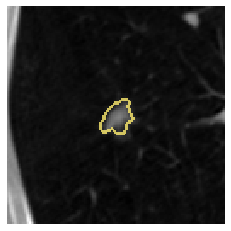} & 
      \includegraphics[width=0.18\textwidth]{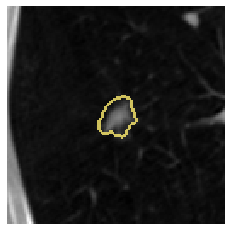} & 
      \includegraphics[width=0.18\textwidth]{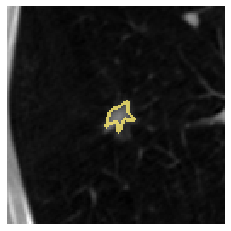} & 
      \includegraphics[width=0.18\textwidth]{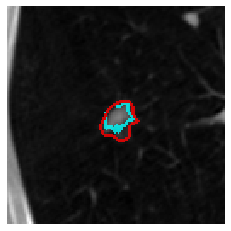} \\

      \includegraphics[width=0.18\textwidth]{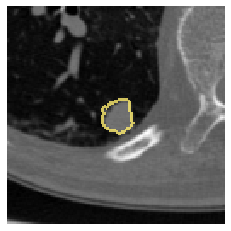} & 
      \includegraphics[width=0.18\textwidth]{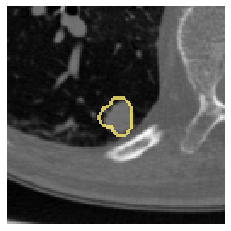} & 
      \includegraphics[width=0.18\textwidth]{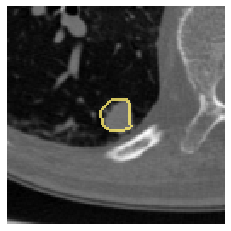} & 
      \includegraphics[width=0.18\textwidth]{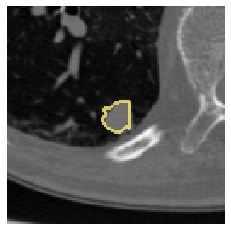} & 
      \includegraphics[width=0.18\textwidth]{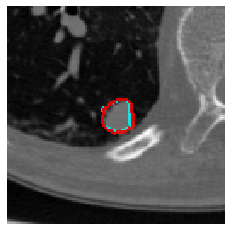} \\

      \includegraphics[width=0.18\textwidth]{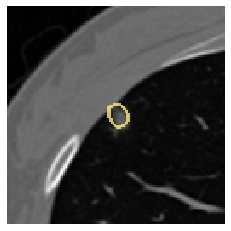} & 
      \includegraphics[width=0.18\textwidth]{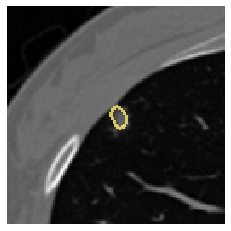} & 
      \includegraphics[width=0.18\textwidth]{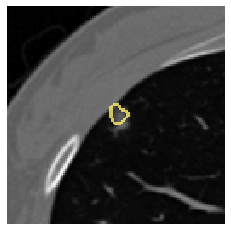} & 
      \includegraphics[width=0.18\textwidth]{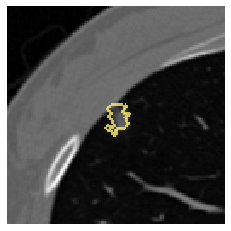} & 
      \includegraphics[width=0.18\textwidth]{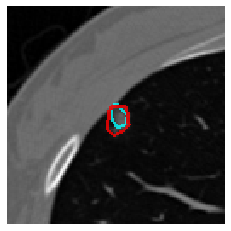} \\

      \includegraphics[width=0.18\textwidth]{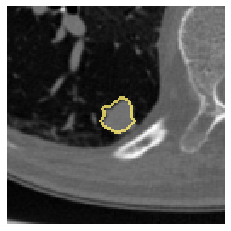} & 
      \includegraphics[width=0.18\textwidth]{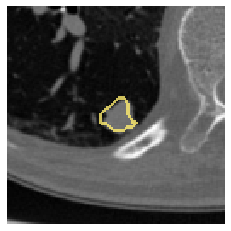} & 
      \includegraphics[width=0.18\textwidth]{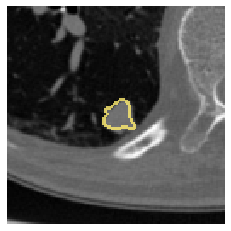} & 
      \includegraphics[width=0.18\textwidth]{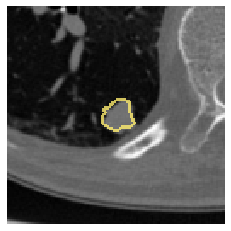} & 
      \includegraphics[width=0.18\textwidth]{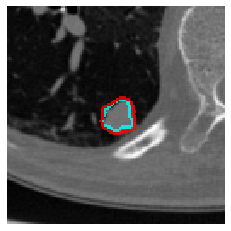} \\
    \end{tabular}
    \caption{Visualization of all annotations and segmentation results for LIDC dataset: in the last column, prediction is in the red and true label (maj) in skyblue.}
  \label{fig:lidc_final}
  \end{figure*}

  \paragraph{LIDC.} The Dice coefficients for all six classes within the LIDC dataset are presented in Table~\ref{tab:LIDC_results}. Within this dataset, class 0 corresponds to the background, while class 1 denotes the lung. As demonstrated by the results, our model with the confidence regularizer typically outperforms the competitors. Interestingly, even in the absence of the confidence regularizer, our method outperforms other models by a substantial margin for class 1. To provide a more detailed comparison, we have included examples of predictions made by our model and true label (taken as majority vote) in Figure~\ref{fig:lidc_final}.
 
  \begin{table*}
    \caption{Comparison of segmentation Dice coefficient (\%) (mean $\pm$ standard deviation) for LIDC, computed separately for each class. \textbf{Best} results are in bold, \underline{second-best} underlined. The red column is the most interesting -- the Dice coefficient between approximated and ``real'' ground-truth masks.}
  \label{tab:LIDC_results}
    \centering
    \resizebox{0.9\textwidth}{!}{
    \begin{tabular}{@{}ScScScScScScSc Sc} 
      \toprule
      & & $D(\tilde{y}_{a},\tilde{y}_{a})$ & $D(y_{a},y_{a})$ & $D(\tilde{y}_{a},y_{a})$ & $D(\tilde{y}_{a},y_{gt})$ & $D(y_{a},y_{gt})$ & $D(\tilde{y},y_{gt})$\\ \midrule

      \multirow{4}{*}{\rotatebox{90}{\begin{tabular}[c]{@{}c@{}}\textbf{MR-Net}\\ \tiny{\cite{MRNet}}\end{tabular}}} & \multirow{2}{*}{class 0} & 99.99 & 99.83 & 99.84 &  99.59 & 99.86 & \textbf{99.85} \\ 
      & & $\pm 0.002$ & $\pm 0.00$ & $\pm 0.01$ & $\pm 0.01$ & $\pm 0.00$ & $\pm 0.02$ \\ \cdashline{2-8}

      & \multirow{2}*{class 1} & 92.08 & 53.02 & 46.16 &  23.18 & 53.39 & 45.62 \\ 
      & & $\pm 1.45$ & $\pm 0.00$ & $\pm 1.48$ &  $\pm .43$ & $\pm 0.00$ & $\pm 9.75$ \\ \midrule


     \multirow{4}{*}{\rotatebox{90}{\begin{tabular}[c]{@{}c@{}}\textbf{PU-Net}\\ \tiny{\cite{kohl2018probabilistic}}\end{tabular}}} & \multirow{2}{*}{class 0} & 99.96 & 99.83 & 99.53 &  99.79 & 99.86 & 99.73 \\ 
      & & $\pm 0.007$ & $\pm 0.00$ & $\pm 0.04$ & $\pm 0.06$ & $\pm 0.00$ & $\pm 0.09$ \\ \cdashline{2-8}

      & \multirow{2}*{class 1} & 91.48 & 53.02 & 28.89 &  43.39 & 53.39 & 43.92 \\ 
      & & $\pm 1.65$ & $\pm 0.00$ & $\pm 1.53$ &  $\pm 1.54$ & $\pm 0.00$ & $\pm 1.25$ \\ \midrule

      \multirow{4}{*}{\rotatebox{90}{\begin{tabular}[c]{@{}c@{}}\textbf{CM-Net}\\ \tiny{\cite{DisentanglingHumanError}}\end{tabular}}} & \multirow{2}{*}{class 0} & 99.96 & 99.83 & 99.42 &  99.50 & 99.86 & 99.38 \\ 
      & & $\pm 0.01$ & $\pm 0.00$ & $\pm 0.03$ & $\pm 0.03$ & $\pm 0.00$ & $\pm 0.04$ \\ \cdashline{2-8}

      & \multirow{2}{*}{class 1} & 99.07 & 53.02 & 32.09 &  47.17 & 53.39 & 46.12 \\ 
      & & $\pm 0.20$ & $\pm 0.00$ & $\pm 0.75$ &  $\pm 1.24$ & $\pm 0.00$ & $\pm 1.30$ \\ \midrule

      \multirow{4}{*}{\rotatebox{90}{\begin{tabular}[c]{@{}c@{}}\textbf{Ours} \end{tabular}}} & \multirow{2}{*}{class 0} & 99.89 & 99.83 & 99.44 &  99.52 & 99.86 & \underline{99.82} \\ 
      & & $\pm 0.05$ & $\pm 0.00$ & $\pm 0.09$ & $\pm 0.09$ & $\pm 0.00$ & $\pm 0.02$ \\ \cdashline{2-8}

      & \multirow{2}{*}{class 1} & 99.18 & 53.02 & 31.43 &  46.21 & 53.39 & \textbf{64.33} \\ 
      & & $\pm 2.28$ & $\pm 0.00$ & $\pm 1.78$ &  $\pm 3.05$ & $\pm 0.00$ & $\pm 1.82$ \\ \midrule

      \multirow{4}{*}{\rotatebox{90}{\begin{tabular}[c]{@{}c@{}}\textbf{Ours (No}\\\textbf{Conf.)}\end{tabular}}} & \multirow{2}{*}{class 0} & 100.00 & 99.83 & 99.57 &  99.65 & 99.86 & 99.65 \\ 
      & & $\pm 0.00$ & $\pm 0.00$ & $\pm 0.02$ & $\pm 0.02$ & $\pm 0.00$ & $\pm 0.02$ \\ \cdashline{2-8}

      & \multirow{2}*{class 1} & 100.00 & 53.02 & 35.09 &  52.20 & 53.39 & \underline{52.20} \\ 
      & & $\pm 0.00$ & $\pm 0.00$ & $\pm 0.70$ &  $\pm 1.05$ & $\pm 0.00$ & $\pm 1.05$ \\ \midrule

    \end{tabular}
    }
  \end{table*}

\begin{figure*}[!htb]
    \centering
    \resizebox{.9\textwidth}{!}{
    \begin{tabular}{c c c}
      Annotations with GT & Annotations with Prediction & Prediction and Ground Truth \\
      \includegraphics[width=0.3\textwidth]{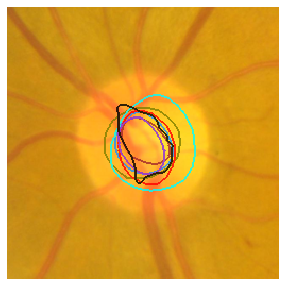} & 
      \includegraphics[width=0.3\textwidth]{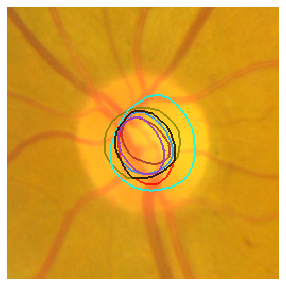} & 
      \includegraphics[width=0.3\textwidth]{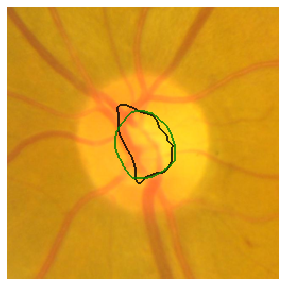} \\

      \includegraphics[width=0.3\textwidth]{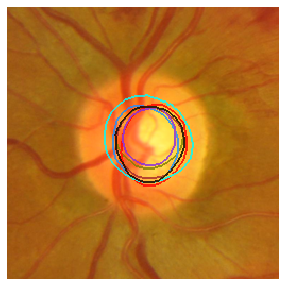} & 
      \includegraphics[width=0.3\textwidth]{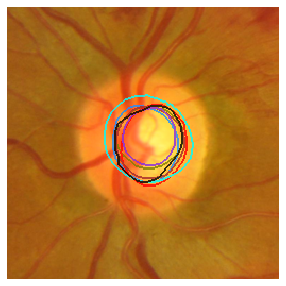} & 
      \includegraphics[width=0.3\textwidth]{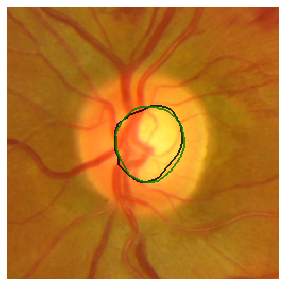} \\

      \includegraphics[width=0.3\textwidth]{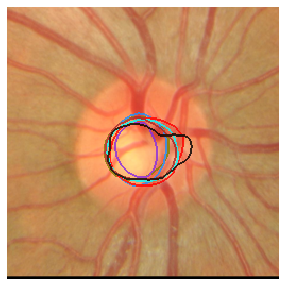} & 
      \includegraphics[width=0.3\textwidth]{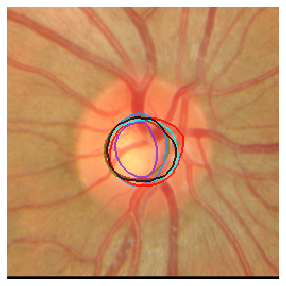} & 
      \includegraphics[width=0.3\textwidth]{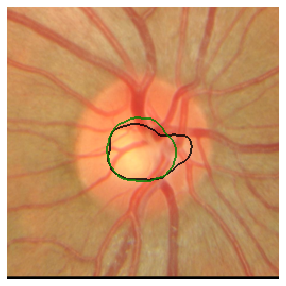} \\

      \includegraphics[width=0.3\textwidth]{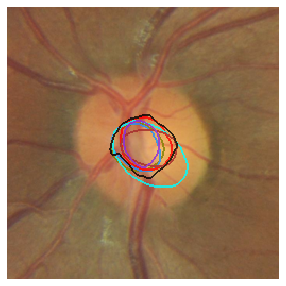} & 
      \includegraphics[width=0.3\textwidth]{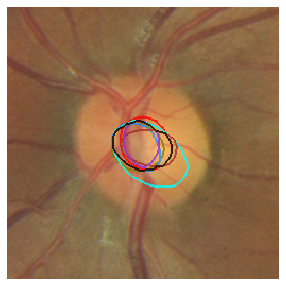} & 
      \includegraphics[width=0.3\textwidth]{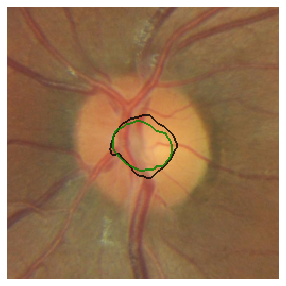} \\
    \end{tabular}
    }
    \caption[Visualization of segmentation results along with annotators (RIGA Optical Cup).]{Visualization of segmentation results along with annotators (RIGA Optical Cup): left: all annotations and prediction (in black); middle: all annotations and true label (in black); right: prediction (in black) and true label (in green).}
  \label{fig:riga_final_cup}
  \end{figure*}

\begin{figure*}
    \resizebox{.9\textwidth}{!}{
    \begin{tabular}{c c c}
      Annotations with GT & Annotations with Prediction & Prediction and Ground Truth \\
      \includegraphics[width=0.3\textwidth]{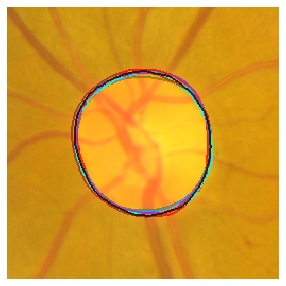} & 
      \includegraphics[width=0.3\textwidth]{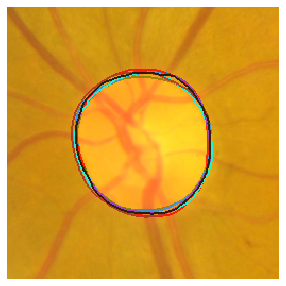} & 
      \includegraphics[width=0.3\textwidth]{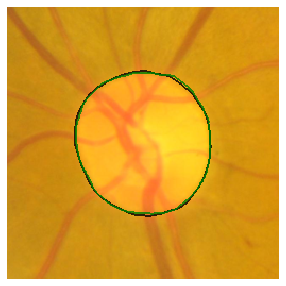} \\

      \includegraphics[width=0.3\textwidth]{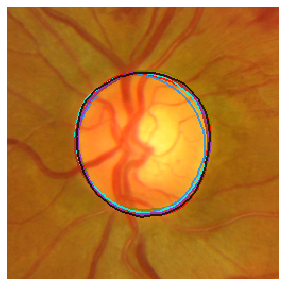} & 
      \includegraphics[width=0.3\textwidth]{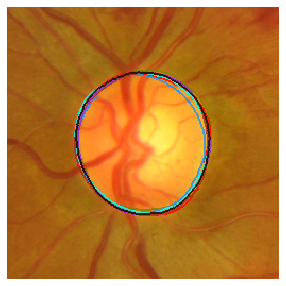} & 
      \includegraphics[width=0.3\textwidth]{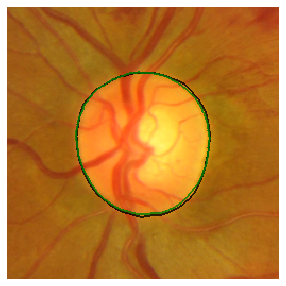} \\

      \includegraphics[width=0.3\textwidth]{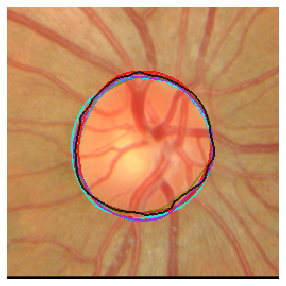} & 
      \includegraphics[width=0.3\textwidth]{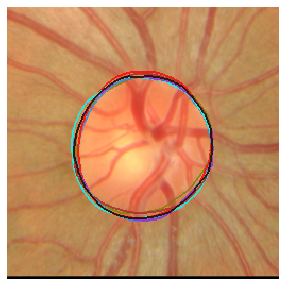} & 
      \includegraphics[width=0.3\textwidth]{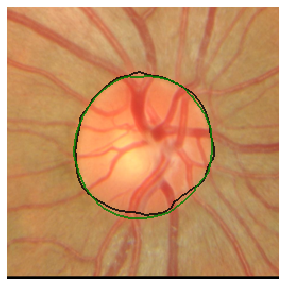} \\

      \includegraphics[width=0.3\textwidth]{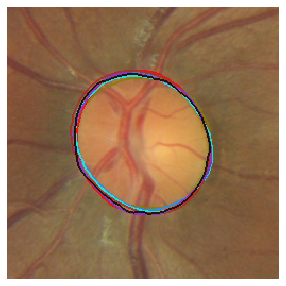} & 
      \includegraphics[width=0.3\textwidth]{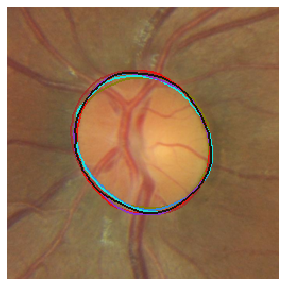} & 
      \includegraphics[width=0.3\textwidth]{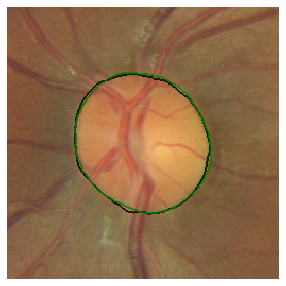} \\
    \end{tabular}
    }
    \caption[Visualization of segmentation results along with annotators (RIGA Optical Disc).]{Visualization of segmentation results along with annotators (RIGA Optical Disc): left: all annotations and prediction (in black); middle: all annotations and true label (in black); right:  prediction (in black) and true label (in green).}
  \label{fig:riga_final_disc}
  \end{figure*}

  \paragraph{RIGA.} The performance metrics for the RIGA dataset are presented in Table~\ref{tab:RIGA_results}. In this context, class 0 corresponds to the background, class 1 is the optical disc, and class 2 is the optical cup. Our models show robust performance for classes 0 and 1 on this dataset and achieve the highest score for class 2. Interestingly, the application of confidence regularization consistently yields superior results. Intriguingly, in this particular instance, the variance of the Dice coefficient for our model is notably smaller in comparison to other methods, underscoring the stability of our approach. For a more detailed visual insight, please refer to Figure~\ref{fig:riga_final_cup} and Figure~\ref{fig:riga_final_disc}.

  \begin{table*} 
    \caption{Comparison of segmentation Dice coefficient (\%) (mean $\pm$ standard deviation) for RIGA, computed separately for each class. \textbf{Best} results are in bold, \underline{second-best} underlined. The red column is the most interesting -- the Dice coefficient between approximated and ``real'' ground-truth masks.}\label{tab:RIGA_results}
    \centering
    \resizebox{0.85\textwidth}{!}{
    \begin{tabular}{@{}ScScScScScScSc Sc} 
      \toprule
      & & $D(\tilde{y}_{a},\tilde{y}_{a})$ & $D(y_{a},y_{a})$ & $D(\tilde{y}_{a},y_{a})$ & $D(\tilde{y}_{a},y_{gt})$ & $D(y_{a},y_{gt})$ & $D(\tilde{y},y_{gt})$\\ \midrule
      \multirow{6}*{\rotatebox{90}{\begin{tabular}[c]{@{}c@{}}\textbf{MR-Net}\\ \tiny{\cite{MRNet}}\end{tabular}}} & \multirow{2}*{class 0} & 99.70 & 98.96 & 99.19 &  99.47 & 99.35 & 99.50 \\ 
      & & $\pm 0.01$ & $\pm 0.00$ & $\pm 0.04$ & $\pm 0.40$ & $\pm 0.00$ & $\pm 0.02$ \\ \cdashline{2-8}

      & \multirow{2}*{class 1} & 95.75 & 89.04 & 91.54 &  93.26 & 93.22 & \textbf{93.97} \\ 
      & & $\pm 0.76$ & $\pm 0.00$ & $\pm 0.26$ &  $\pm 0.25$ & $\pm 0.00$ & $\pm 0.50$ \\  \cdashline{2-8}

      & \multirow{2}*{class 2} & 85.16 & 78.03 & 81.06 &  81.82 & 86.22 & \underline{86.16} \\ 
      & & $\pm 2.01$ & $\pm 0.00$ & $\pm 2.54$ &  $\pm 2.62$ & $\pm 0.00$ & $\pm 2.92$ \\ \midrule


      \multirow{6}*{\rotatebox{90}{\begin{tabular}[c]{@{}c@{}}\textbf{PU-Net}\\ \tiny{\cite{kohl2018probabilistic}}\end{tabular}}} & \multirow{2}*{class 0} & 99.88 & 98.96 & 99.19 &  99.52 & 99.35 & \bf{99.60} \\ 
      & & $\pm 0.05$ & $\pm 0.00$ & $\pm 0.18$ & $\pm 0.89$ & $\pm 0.00$ & $\pm 0.23$ \\ \cdashline{2-8}

      & \multirow{2}*{class 1} & 91.73 & 89.04 & 90.59 &  90.96 & 93.22 & 90.90 \\ 
      & & $\pm 0.89$ & $\pm 0.00$ & $\pm 0.45$ &  $\pm 0.35$ & $\pm 0.00$ & $\pm 0.83$ \\  \cdashline{2-8}

      & \multirow{2}*{class 2} & 78.67 & 78.03 & 80.68 &  83.36 & 86.22 & 80.55 \\ 
      & & $\pm 1.09$ & $\pm 0.00$ & $\pm 1.48$ &  $\pm 1.59$ & $\pm 0.00$ & $\pm 0.87$ \\ \midrule

      \multirow{6}*{\rotatebox{90}{\begin{tabular}[c]{@{}c@{}}\textbf{CM-Net}\\ \tiny{\cite{DisentanglingHumanError}}\end{tabular}}} & \multirow{2}*{class 0} & 99.58 & 98.96 & 98.93 &  99.10 & 99.35 & 99.07 \\ 
      & & $\pm 0.07$ & $\pm 0.00$ & $\pm 0.06$ & $\pm 0.06$ & $\pm 0.00$ & $\pm 0.10$ \\  \cdashline{2-8}

      & \multirow{2}*{class 1} & 93.78 & 89.04 & 89.29 &  90.34 & 93.22 & 90.21 \\ 
      & & $\pm 0.51$ & $\pm 0.00$ & $\pm 0.47$ &  $\pm 0.47$ & $\pm 0.00$ & $\pm 0.78$ \\  \cdashline{2-8}

      & \multirow{2}*{class 2} & 84.47 & 78.03 & 78.87 &  79.84 & 86.22 & 80.80 \\ 
      & & $\pm 2.42$ & $\pm 0.00$ & $\pm 0.97$ &  $\pm 1.23$ & $\pm 0.00$ & $\pm 1.36$ \\ \midrule

      \multirow{6}*{\rotatebox{90}{\textbf{Ours}}} & \multirow{2}*{class 0} & 99.54 & 98.96 & 99.07 &  99.24 & 99.35 & \underline{99.33} \\ 
      & & $\pm 0.05$ & $\pm 0.00$ & $\pm 0.02$ & $\pm 0.30$ & $\pm 0.00$ & $\pm 0.01$ \\  \cdashline{2-8}

      & \multirow{2}*{class 1} & 94.04 & 89.04 & 90.76 &  91.80 & 93.22 & \underline{93.63} \\ 
      & & $\pm 0.40$ & $\pm 0.00$ & $\pm 0.19$ &  $\pm 0.25$ & $\pm 0.00$ & $\pm 0.07$ \\  \cdashline{2-8}

      & \multirow{2}*{class 2} & 84.05 & 78.03 & 81.52 &  82.21 & 86.22 & \textbf{88.58} \\ 
      & & $\pm 1.63$ & $\pm 0.00$ & $\pm 0.61$ &  $\pm 0.96$ & $\pm 0.00$ & $\pm 0.25$ \\ \midrule

      \multirow{6}*{\rotatebox{90}{\begin{tabular}[c]{@{}c@{}}\textbf{Ours (No}\\\textbf{Conf.)}\end{tabular}}} & \multirow{2}*{class 0} & 99.62 & 98.96 & 99.05 &  99.24 & 99.35 & 99.26 \\ 
      & & $\pm 0.05$ & $\pm 0.00$ & $\pm 0.01$ & $\pm 0.02$ & $\pm 0.00$ & $\pm 0.03$ \\  \cdashline{2-8}

      & \multirow{2}*{class 1} & 94.02 & 89.04 & 90.56 &  91.62 & 93.22 & 92.18 \\ 
      & & $\pm 0.42$ & $\pm 0.00$ & $\pm 0.15$ &  $\pm 0.27$ & $\pm 0.00$ & $\pm 0.49$ \\  \cdashline{2-8}

      & \multirow{2}*{class 2} & 83.78 & 78.03 & 82.09 &  82.56 & 86.22 & 84.75 \\ 
      & & $\pm 2.15$ & $\pm 0.00$ & $\pm 0.66$ &  $\pm 1.23$ & $\pm 0.00$ & $\pm 1.29$ \\ \midrule

    \end{tabular}
    }
  \end{table*}

\subsection{Comparative Analysis of Various Regularizers for Segmentation.}
\label{sec:reg_ablation}
  This section is dedicated to a comparative analysis of various regularizers applied to the same architecture type, namely segmentation and annotation networks; and the dilation procedure (see Table~\ref{tab:reg_ablation} below).
  The goal of this ablation study is to assess the individual and collective impact of these regularizers on real-world image segmentation datasets, thereby emphasizing the importance of the regularizer choice in such models. We observe that having both loss terms we proposed is critical to achieving high-quality results. Interestingly, we see that for the LIDC dataset, one can slightly improve the results further by tuning the dilation procedure.

  The results of our analysis, which are summarized in Table~\ref{tab:reg_ablation}, consider several regularizers.
  Among these are the CM-Net~\cite{DisentanglingHumanError} and the Logdet~\cite{li2021provably}, in addition to other combinations.
  We intended to compare these well-established regularizers against the one we propose in this study.

  Notably, the table reveals that our proposed confidence regularizer, which includes an additional term for confident regions (CR), exceeds the performance of the other options by a considerable margin. 
  This superior performance may be attributed to the fact that our proposed regularizer does not impose any additional constraints on the parameters. For instance, it does not enforce diagonal dominance as in the Logdet~\cite{li2021provably}, a requirement that is explicitly stated. Neither does it employ the same constraints as those theoretically proven in CM-Net~\cite{DisentanglingHumanError}, whose practical implementation in the algorithm and code remains somewhat ambiguous.

  \begin{table*}
  \caption{In this table we present additional ablation results. Average Dice score with standard deviation over 10 last epochs, computed for different combinations of regularizers. Reg-(Y/N) means whether we apply (Y) our confidence regularizer or not (N). CR-(Y/N) means the same for the confidence region (CR) term on the loss function. The symbol ``R'' represents the count of dilation iterations. In the last two columns, we keep Reg-Y and CR-Y, while varying the number of dilation iterations. The \textbf{bold} results highlight the top performances, while the \underline{underlined} ones signify the next best outcomes.}
  \label{tab:reg_ablation}
  \centering
  \resizebox{0.99\textwidth}{!}{
  \begin{tabular}{cc|c|c|c|c|c|c|c|c|c|}
  \cline{3-11}
                                              &      & \makecell{Ours (R=4),\\ Reg-Y, CR-Y} & \makecell{Ours (R=4),\\ Reg-Y, CR-N} & \makecell{Ours (R=4),\\ Reg-N, CR-Y} & \makecell{Ours (R=4),\\ Reg-N, CR-N} & CM-Net & LogDet & \makecell{LogDet\\ + Ours} & Ours: R=3 & Ours: R=5 \\ \hline
  \multicolumn{1}{|c|}{\multirow{2}{*}{RIGA}} & mean & \textbf{93.85} & 92.06 & 91.43 & 75.8 & 90.03 & \underline{93.31} & 92.99 & 92.14 & 91.17  \\ \cline{2-11} 
  \multicolumn{1}{|c|}{}                      & std  & $\pm 0.09$ & $\pm 0.58$ & $\pm 0.25$ & $\pm 0.47$ & $\pm 0.91$ & $\pm 0.11$ & $\pm 0.26$ & $\pm 0.12$ & $\pm 0.58$  \\ \hline 

  \multicolumn{1}{|l|}{}                      & & & & & & & &    \\ 

  \multicolumn{1}{|l|}{\textbf{}}             &      & \makecell{Ours (R=2),\\ Reg-Y, CR-Y} & \makecell{Ours (R=2),\\ Reg-Y, CR-N} & \makecell{Ours (R=2),\\ Reg-N, CR-Y} & \makecell{Ours (R=2),\\ Reg-N, CR-N} & CM-Net & LogDet & \makecell{LogDet\\ + Ours} & Ours: R=1 & Ours: R=3  \\ \hline
  \multicolumn{1}{|l|}{\multirow{2}{*}{LIDC}} & mean & \underline{82.07} & 75.93  & 79.48 & 75.31 & 72.75 & 77.38 & 77.58 & \textbf{82.82} & 80.92  \\ \cline{2-11} 
  \multicolumn{1}{|l|}{}                      & std  & $\pm 0.92$ & $\pm 0.53$ & $\pm 0.38$ & $\pm 1.14$ & $\pm 0.66$ & $\pm 0.69$ & $\pm 0.69$ & $\pm 0.21$ & $\pm 0.41$ \\ \hline
  \end{tabular}
  }
  \end{table*}

  In conclusion, these results highlight the potential benefits of our proposed regularizer within the context of medical image segmentation.

\subsection{Classiffication} 
\label{sec:appendix_ClassifficationSupplementaryResults}

  \paragraph{MNIST.} We used a LeNet model as a classifier for our backbone network. The LeNet5 architecture is a convolutional neural network for image classification, featuring a feature extractor and a classifier. The feature extractor consists of three convolutional layers with 6, 16, and 120 filters, each followed by ReLU activation and average pooling. The output is flattened and passed to the classifier, which has two fully connected layers: the first with 84 nodes and the second with number of classes nodes, ending with a softmax for class probabilities. For the annotator network, we have a linear layer of size $C \times C$, $C$ denotes the number of classes. This linear layer represents our annotator confusion matrices, and we apply a softmax layer to it to make it a stochastic matrix along a certain dimension. We fine-tuned our model for a combination of learning rates, $\alpha$ = [0.01, 0.001, 0.0001, 0.000001, 0.0016, 0.008, 0.0064, 0.005], and about 50 different lambda values, $\lambda$, for our regularizer hyper-parameter. We started with a very small value of $\lambda$ = 0.001506746, and slowly increased it exponentially (geometric progression) per epoch with a rate, r = 1.18; we trained the model for 70 epochs and used Adam as an optimizer. In addition, the experiments were regulated to assess the performance of the model when the confusion matrix was initialized as an identity. These fine-tunings are done across all 2 different types of noise described in Section~\ref{sec:datasets_classification} with the respective noise rate that is associated with each of the noise types.

  Tables~\ref{tab:mnist_data} and \ref{tab:fmnist} show the performance comparison of our algorithm with the other methods. The results of our methodology with entropy and confidence regularizers are superior compared to other methods, especially for more challenging cases of higher noise rates.

  \begin{table*}[!h]
  \caption{Comparison of test accuracy (\%) (mean $\pm$ st. dev.) for MNIST. \textbf{Best} results are in bold, \underline{second-best} underlined.}\label{tab:mnist_data}
  \centering
  \resizebox{0.85\textwidth}{!}{
  \begin{tabular}{c c c c c c c c}
    \toprule
    Noise rate& Ours-Inf & Ours-Ent & Co-tea. & Co-tea.+ & JoCoR & Trace & CDR\\
    \midrule
    symmetric 20\% & \underline{99.20} & \bf{99.48} & 99.01 & 98.88 & 98.82 & 99.16 & 98.97\\ 
                    & $\pm$ 0.20 & $\pm$ 0.23 & $\pm$ 0.02& $\pm$ 0.02 & $\pm$ 0.03 & $\pm$ 0.21 & $\pm$ 0.02 \\
                    
    symmetric 30\% & \textbf{99.11} & \underline{99.09} & 98.78 & 98.38 & 98.40 & 99.01 & 98.75\\
                    & $\pm$ 0.34 & $\pm$ 0.38 & $\pm$ 0.03& $\pm$ 0.02 & $\pm$ 0.02 & $\pm$ 0.43 & $\pm$ 0.02 \\
                    
    symmetric 50\% & \textbf{98.94} & \underline{98.93} & 92.24 &  95.26 & 96.83 & 98.87 & 97.72\\
                    & $\pm{0.77}$ & $\pm{0.81}$ & $\pm{0.02}$ & $\pm{0.03}$ & $\pm{0.04}$ & $\pm{0.80}$ & $\pm{0.02}$ \\
                    
    pairflip 20\% & 99.08 & \textbf{99.55} & 98.84 & 98.59 & 98.89 & \underline{99.13} & 98.88\\ 
                    & $\pm{0.10}$ & $\pm{0.15}$ & $\pm{0.02}$ & $\pm{0.01}$ & $\pm{0.04}$ & $\pm{0.19}$ & $\pm{0.01}$ \\
                    
    pairflip 30\% & 98.94 & \textbf{99.54} & 98.57 & 97.95 & 98.56 & \underline{99.08} & 98.50\\ 
                    & $\pm{0.17}$ & $\pm{0.20}$ & $\pm{0.02}$ & $\pm{0.01}$ & $\pm{0.04}$ & $\pm{0.13}$ & $\pm{0.01}$ \\
                    
    pairflip 45\% & \underline{98.77} & \textbf{99.10} & 87.63 & 71.36 & 85.86 & 97.95  & 87.04\\
                & $\pm{0.39}$ & $\pm{0.35}$ & $\pm{0.04}$ & $\pm{0.06}$ & $\pm{0.05}$ & $\pm{1.09}$ & $\pm{0.63}$ \\
    \bottomrule 
  \end{tabular}
  }

    \caption{Comparison of test accuracy (\%) (mean $\pm$ st.dev.) for Fashion-MNIST. \textbf{Best} results are in bold, \underline{second-best} underlined.}\label{tab:fmnist}
    \centering
    \begin{tabular}{c c c c c c c c}
      \toprule
      Noise rate& Ours-Inf & Ours-Ent & Co-tea & Co-tea+ & JoCoR & Trace & CDR \\
      \midrule
      symmetric 20\% & 90.67 & \underline{90.79} & 90.48 & 88.69 & \textbf{91.88} & 90.61 & 88.69\\ 
                      & $\pm$ 0.21 & $\pm$ 0.19 & $\pm$ 0.06 & $\pm$ 0.09 & $\pm$ 0.06 & $\pm$ 0.24 & $\pm$ 0.09 \\
                      
      symmetric 30\% & \textbf{91.35} &  90.34 & 90.36 & 88.50 & \underline{91.33} & 89.64  & 87.38\\
                      & $\pm$ 0.25 & $\pm$ 0.26 & $\pm$ 0.13 & $\pm$ 0.09 & $\pm$ 0.08 & $\pm$ 0.30 & $\pm$ 0.09 \\
                      
      symmetric 50\% & \textbf{89.51} & \underline{89.49} & 89.37 & 77.96 & 89.21 & 88.94 & 85.36\\
                       & $\pm$ 0.55 & $\pm$ 0.58 & $\pm$ 0.08 & $\pm$ 0.33 & $\pm$ 0.06 & $\pm$ 0.53 & $\pm$ 0.09 \\
                     
      pairflip 20\% & \underline{90.90} & 90.77 & 90.68 & 89.12 & \textbf{91.37} & 90.40 & 90.01\\
                    & $\pm$ 0.33 & $\pm$ 0.29 & $\pm$ 0.05 & $\pm$ 0.08 & $\pm$ 0.09 & $\pm$ 0.35 & $\pm$ 0.11 \\
                    
      pairflip 30\% & \underline{90.38} & \textbf{90.65} & 90.11 & 89.06 & 89.67 & 90.33 & 88.78 \\ 
                    & $\pm$ 0.41 & $\pm$ 0.45 & $\pm$ 0.08 & $\pm$ 0.10 & $\pm$  0.07& $\pm$ 0.46 & $\pm$ 0.14 \\
                    
      pairflip 45\% & \textbf{89.37} & 89.02 & 78.86 & 52.61 & 88.10 & \underline{89.08} & 64.63\\
                    & $\pm$ 0.68 & $\pm$ 0.71 & $\pm$ 0.11 & $\pm$ 0.43 & $\pm$ 0.39 & $\pm$ 0.72 & $\pm$ 0.17 \\
      \bottomrule
    \end{tabular}
  
  \end{table*}

\begin{figure*}
    \begin{tabular}{c c c}
      Annotations with GT & Annotations with Prediction & Prediction and Ground Truth \\
    
      \multicolumn{3}{|c|}{ \multirow{1}{*}{ \fbox{\includegraphics[trim={1.9cm 10 .5cm 14cm},clip,width=0.95\textwidth]{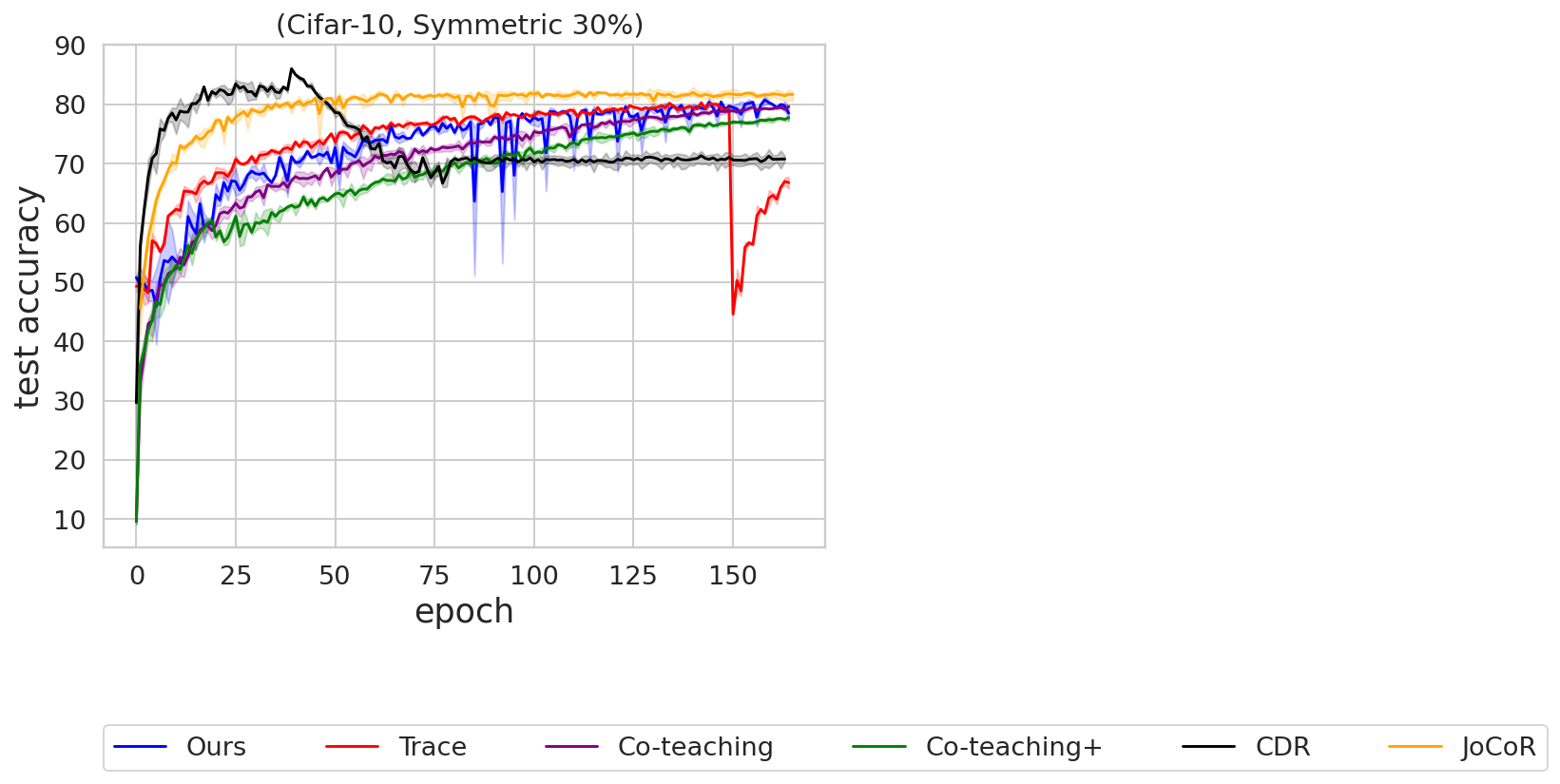}} } } \\

      \includegraphics[width=0.3\textwidth]{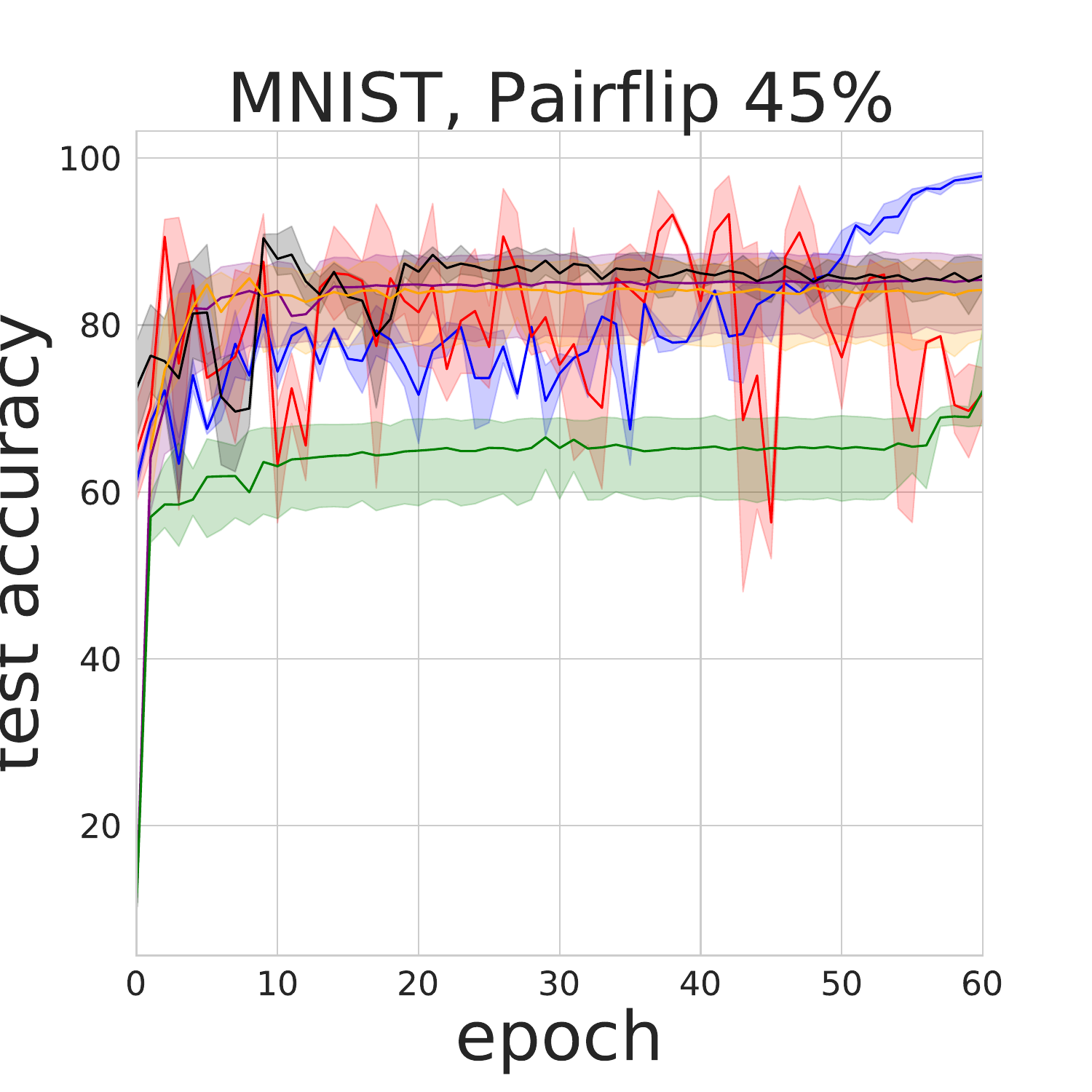} & 
      \includegraphics[width=0.3\textwidth]{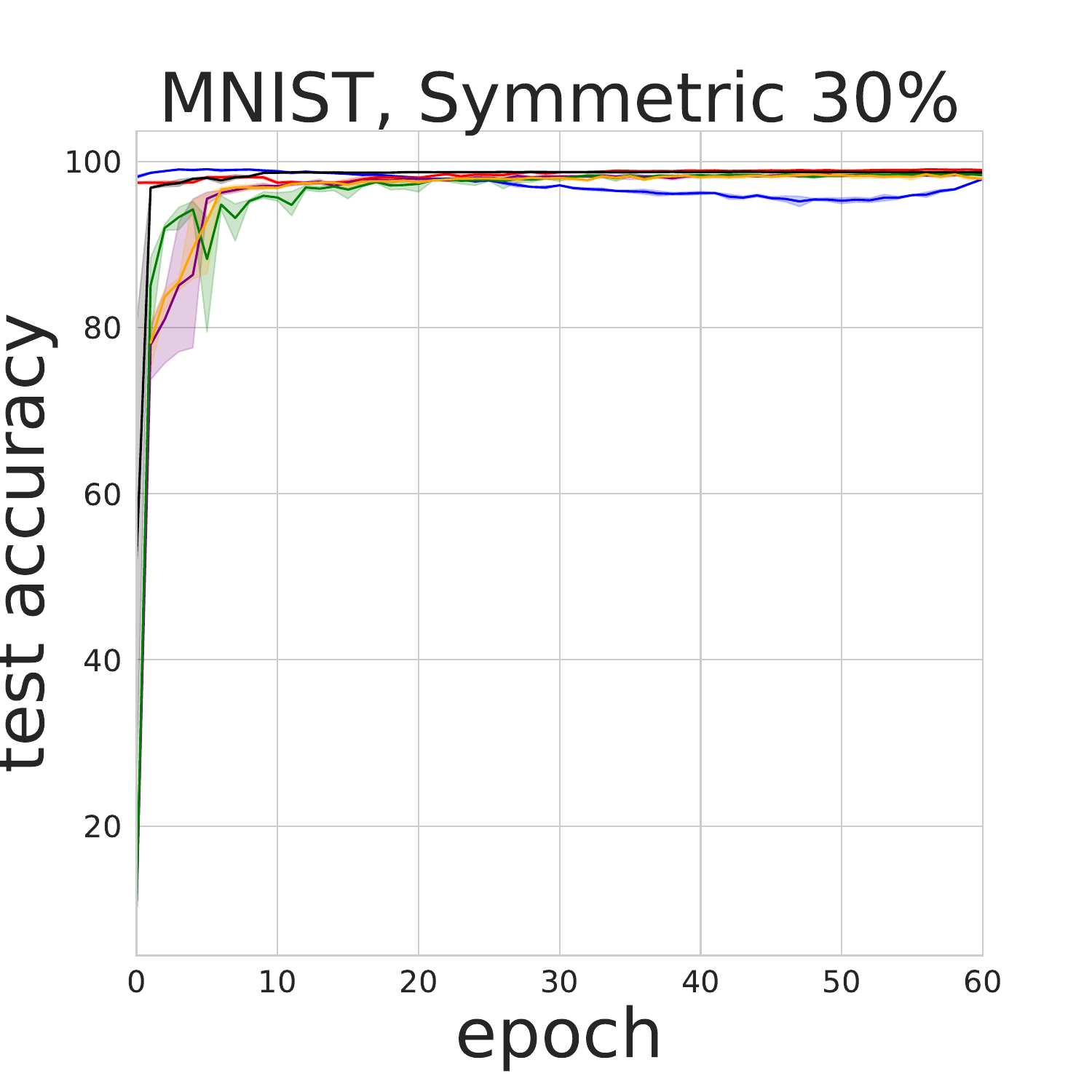} & 
      \includegraphics[width=0.3\textwidth]{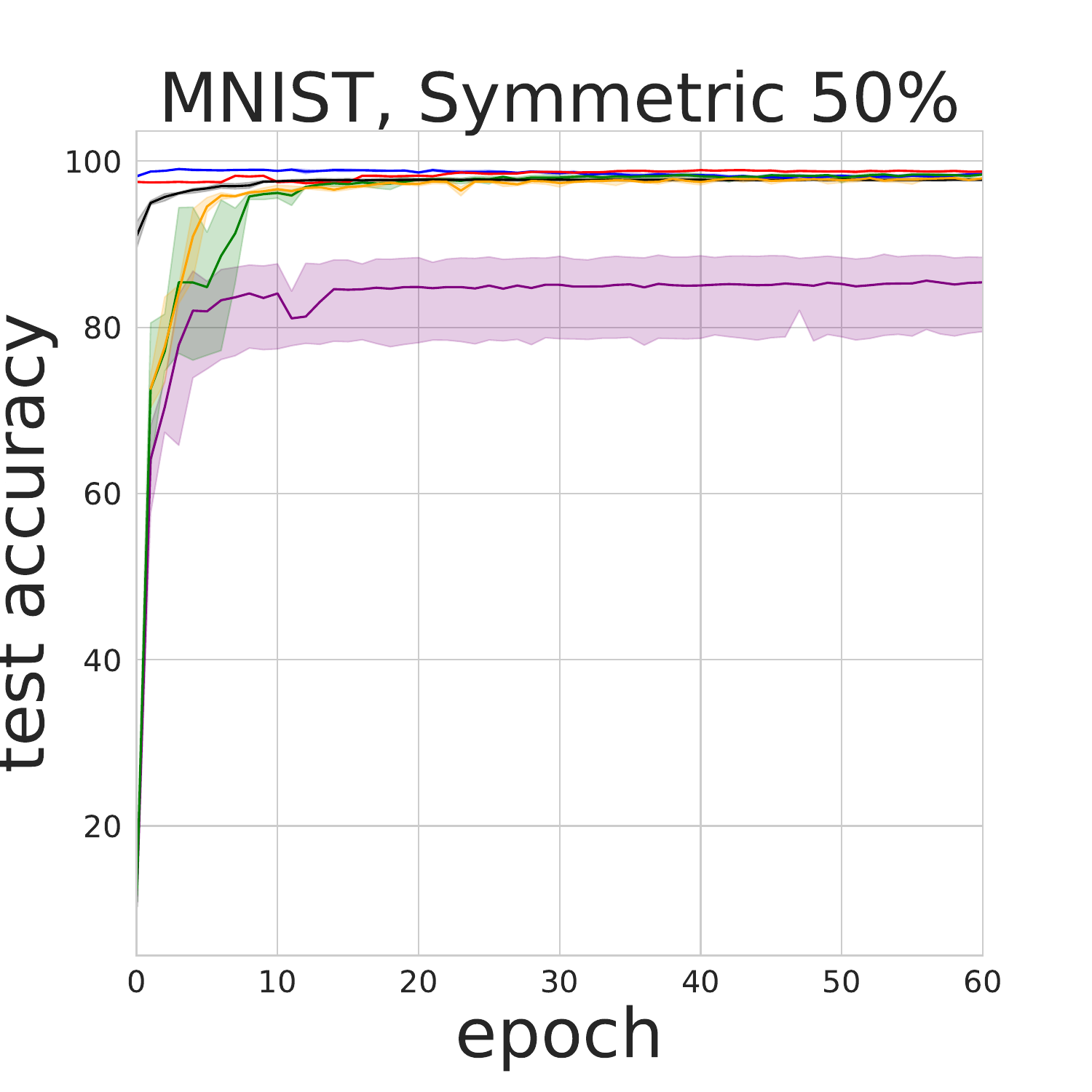} \\
    \end{tabular}
    \caption{Test accuracy (\%) vs. number of epochs on MNIST.}
  \label{fig:mnist_performance}

  \begin{tabular}{c c c}
      \includegraphics[width=0.3\textwidth]{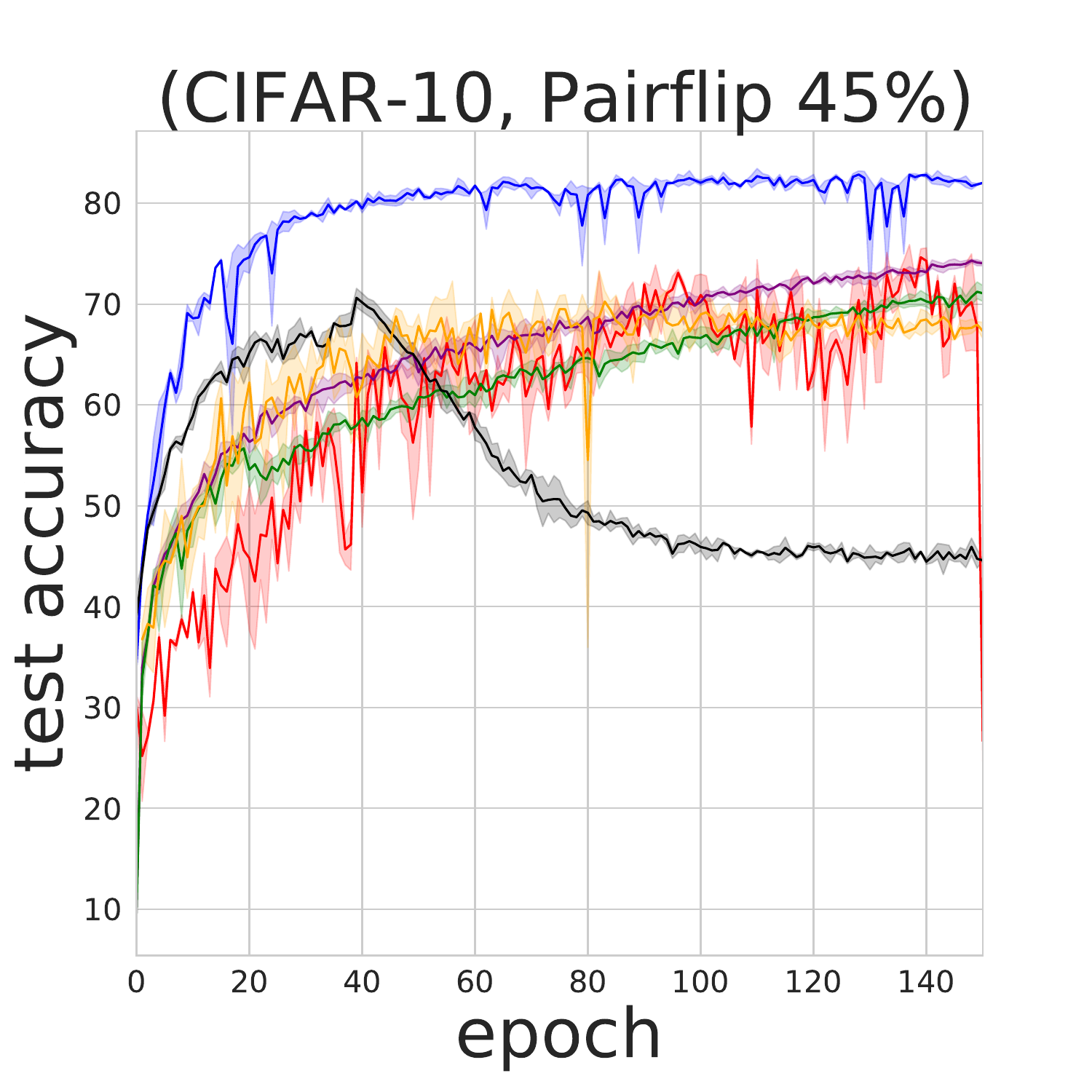} & 
      \includegraphics[width=0.3\textwidth]{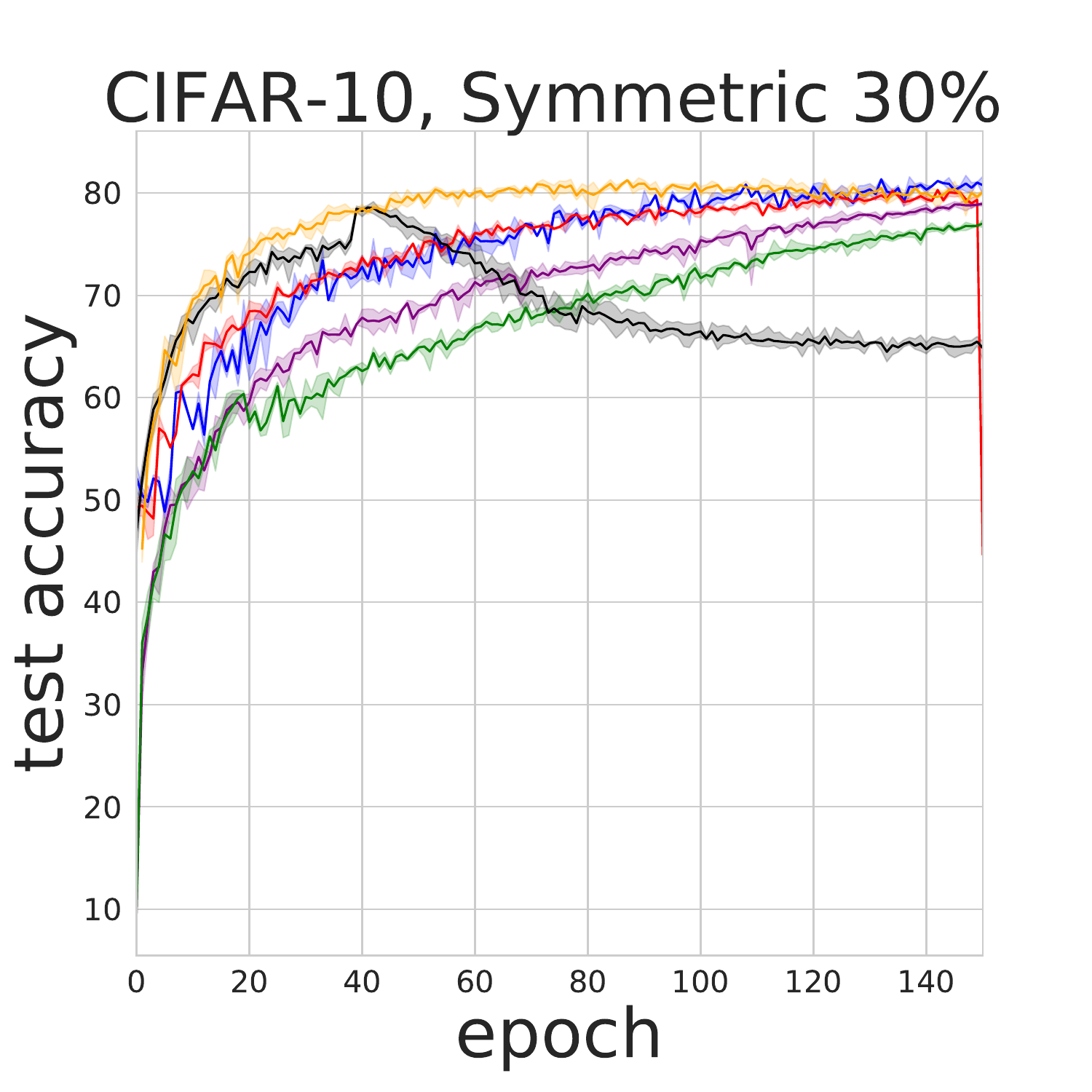} & 
      \includegraphics[width=0.3\textwidth]{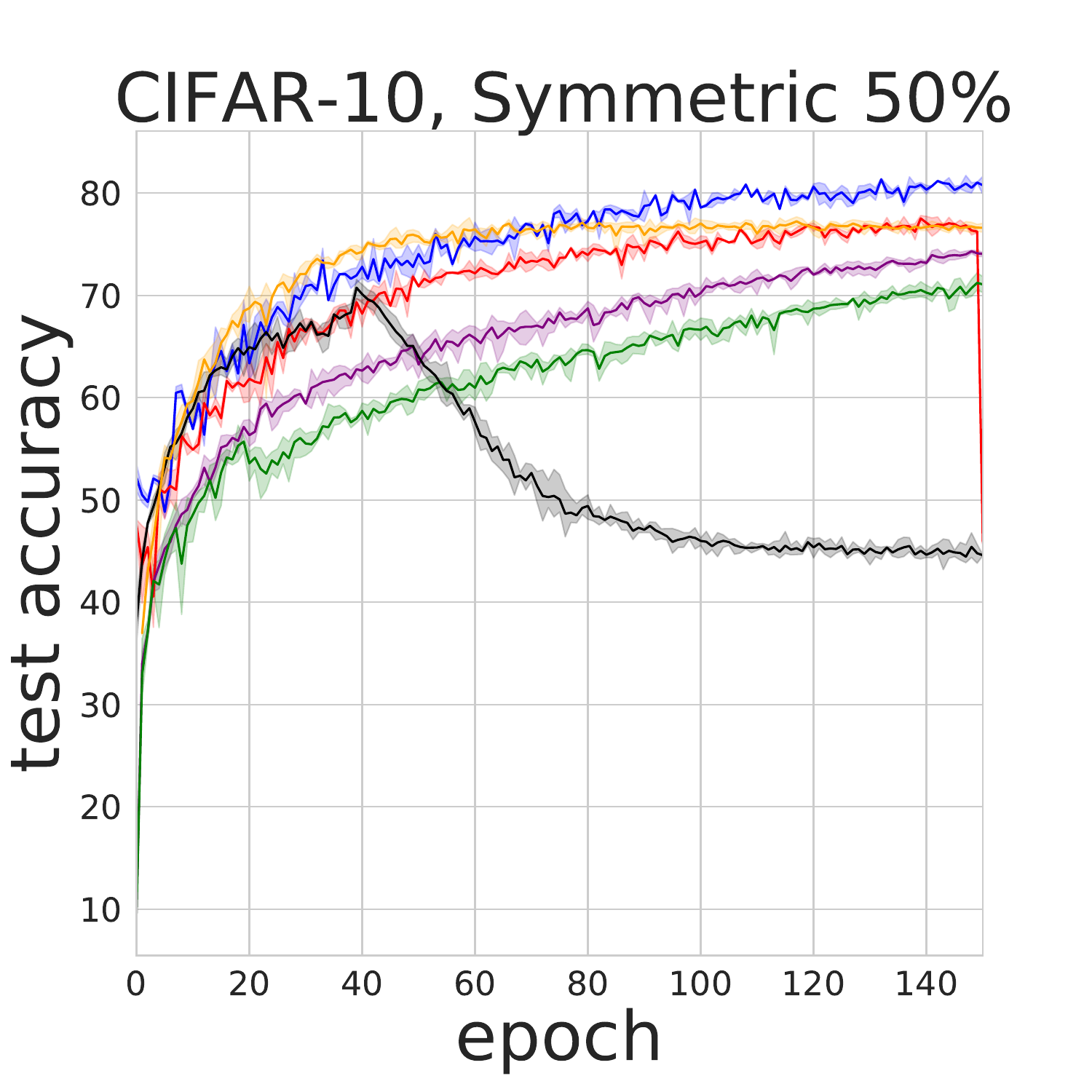} \\
    \end{tabular}
    \caption{Test accuracy ($\%$) vs. epochs on CIFAR-10.}
    \label{fig:cifar10_performance}

    \begin{tabular}{c c c}
      \includegraphics[width=0.3\textwidth]{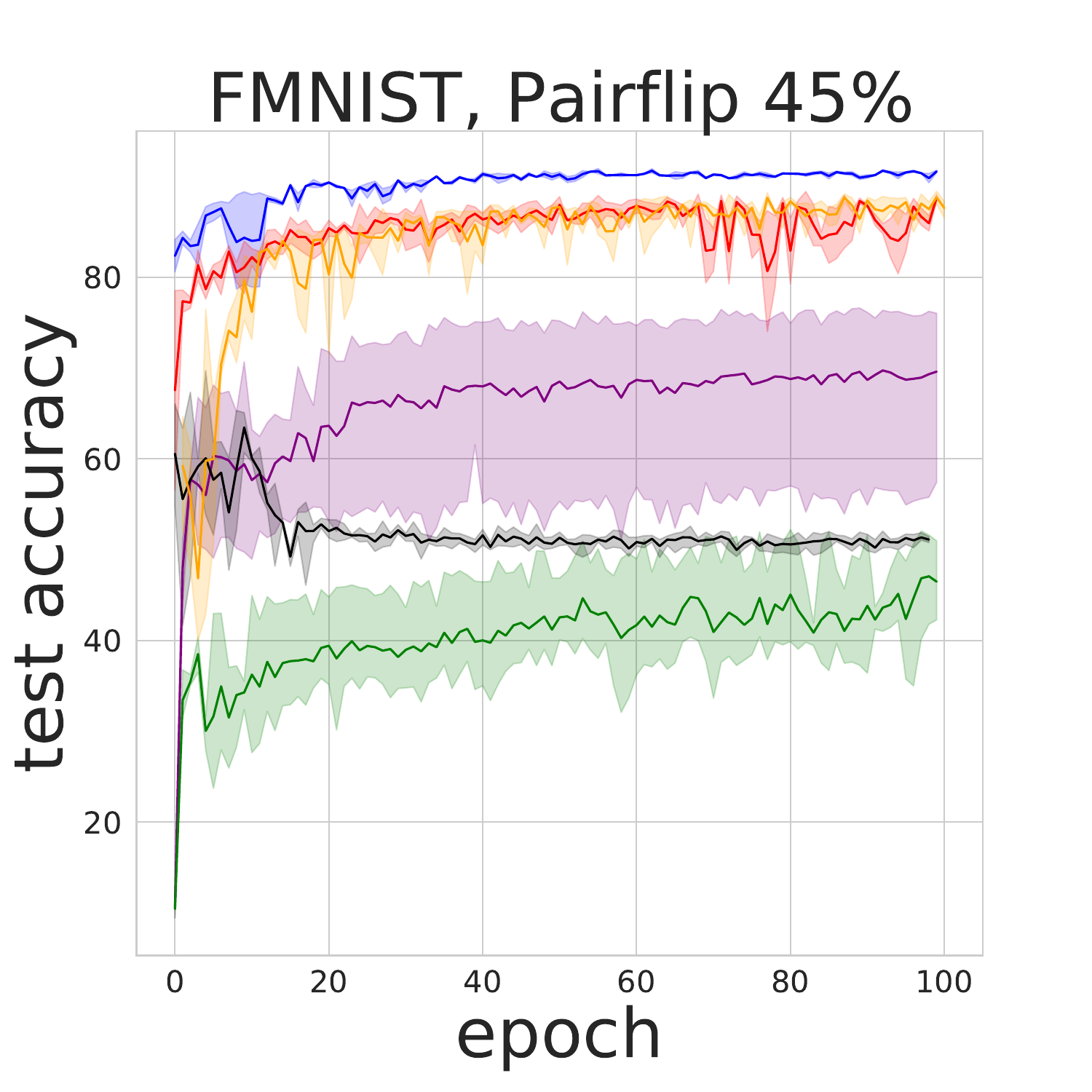} & 
      \includegraphics[width=0.3\textwidth]{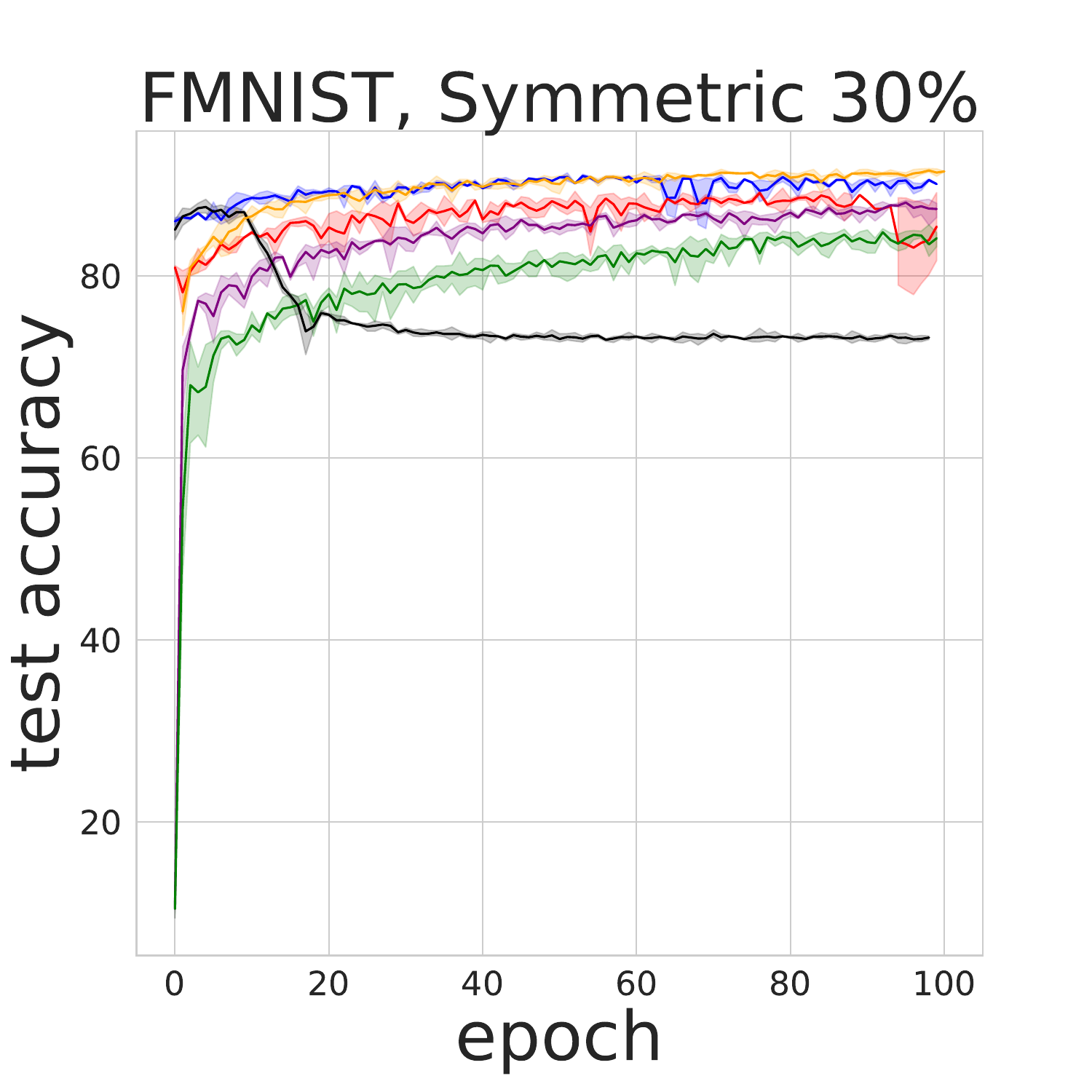} & 
      \includegraphics[width=0.3\textwidth]{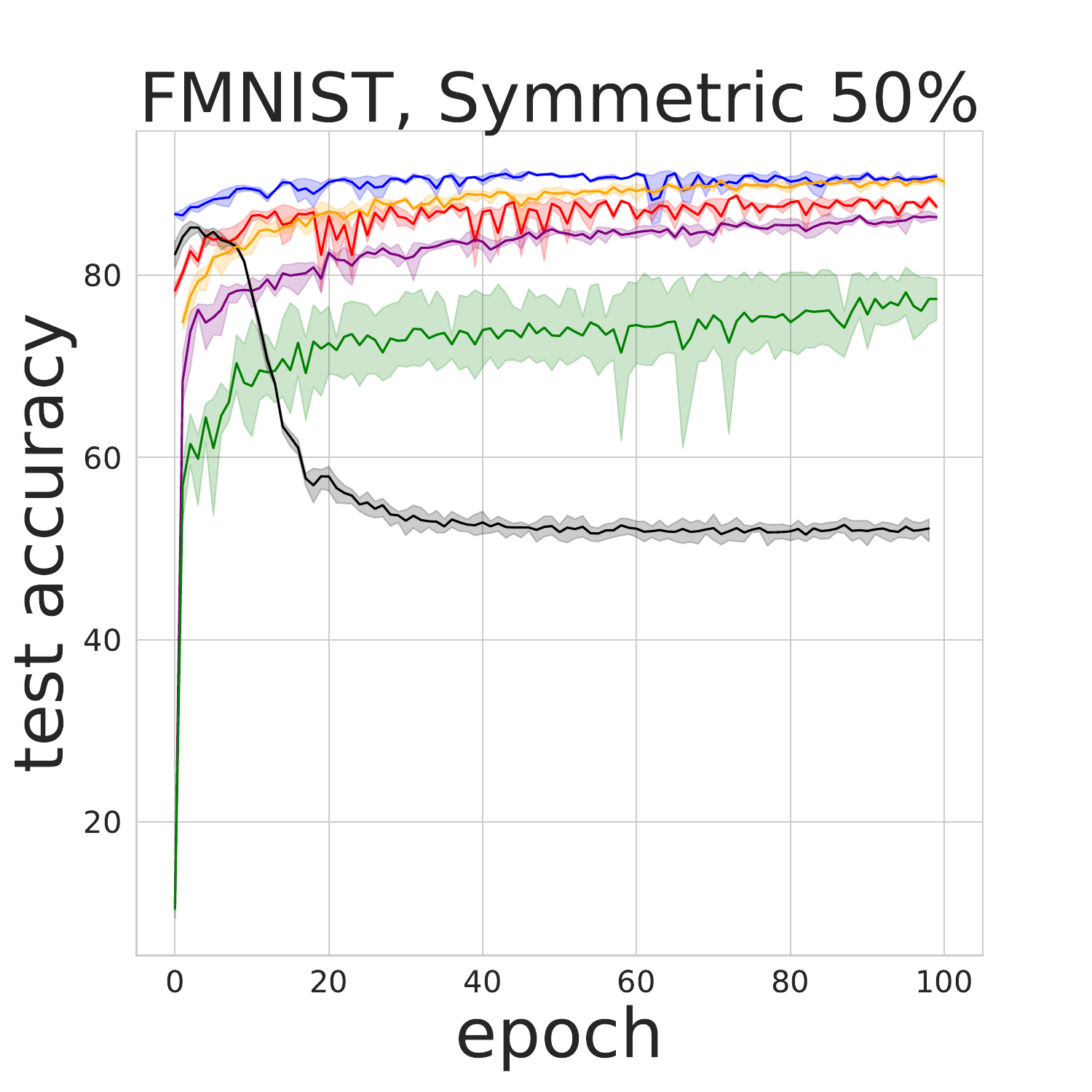} \\
    \end{tabular}
    \caption{Test accuracy ($\%$) vs. epochs on Fashion-MNIST.}
    \label{fig:fmnist_performance}
  \end{figure*}

  In Figure~\ref{fig:mnist_performance}, we highlight the test accuracy vs. number of epochs. We can see clearly that for symmetric noise types, all algorithms gave comparable performance. It can be clearly seen that for symmetric noise, our test accuracy starts to decline a bit. This could be alleviated with an early stopping criterion which was not incorporated in these experiments. For pairflip 45$\%$, the test accuracy starts to increase and stabilize in the later epochs of the experiment and transcends all the baselines.

  \paragraph{CIFAR-10.}\label{sec:CIFAR10_fine_tune}
  For CIFAR-10, we used ResNet-18 as our backbone network for the classifier. The ResNet18 architecture has 18 learnable layers. It starts with an initial convolutional layer that outputs 64 filters, followed by 4 residual blocks where the number of filters increases from 64 to 512. After global average pooling reduces the feature map to 512 dimensions, a custom fully connected layer maps this to the target classes for classification. The annotator network remains unchanged (still has one linear layer that represents the confusion matrices of class $C \times C$ that are stochastic). For this dataset, we fine-tuned the model for an assortment of learning rates, such as $\alpha$ = [0.001, 0.00064, 0.0016, 0.000001, 0.005, 0.008, 0.0016, 0.00064]. We ran the model for 150 epochs; the hyperparameter $\lambda$ for confidence regularizer was slowly increased exponentially again with a rate, r= 1.11. However, the starting value this time is $\lambda$= 3.0517578125e-05. We used a standard batch size, BS=128. We used the standard augmentations of a random crop of size 32 $\times$ 32 and horizontal random flipping. These are the standard augmentations that have been used across all the baselines that we have evaluated. The remaining settings remain the same as described in MNIST above.
  
  Figure~\ref{fig:cifar10_performance} shows the illustrative results of test accuracy vs. number of epochs. In all three plots, it can be clearly seen that our algorithm performs at par with the other algorithms, but the performance gets robustly superior in the extreme noise type of pairflip 45$\%$. This shows that our method is particularly robust against harder noise as it is able to make confident predictions. 
  
  \paragraph{Fashion-MNIST.} We kept the same settings of CIFAR-10, such as the ResNet-18 model and batch size of 128 for the Fashion-MNIST dataset. The model was again fine-tuned for the same set of hyperparameters. However, the starting value of $\lambda$= 6.103515625e-05, and it was increased exponentially with a rate of r=1.12. We also retained the same set of augmentations that we used in the CIFAR-10 dataset.
  
  Figure~\ref{fig:fmnist_performance} gives an illustrative result of test accuracy vs. the number of epochs on the Fashion-MNIST dataset. It showcases the test performance of our algorithm in comparison with other baselines. We can see that for all noise instances, our algorithm performs at par with the high-achieving method like JoCoR. We perform considerably better against sample selection methods like Co-teaching and Co-teaching+, as well as against another method like CDR in the instance of pairflip 45$\%$. 

\begin{figure*}
  \begin{minipage}{\textwidth}  
    \centering
     \begin{minipage}{1.0\textwidth}
      \centering
        \includegraphics[width=1.0\textwidth]{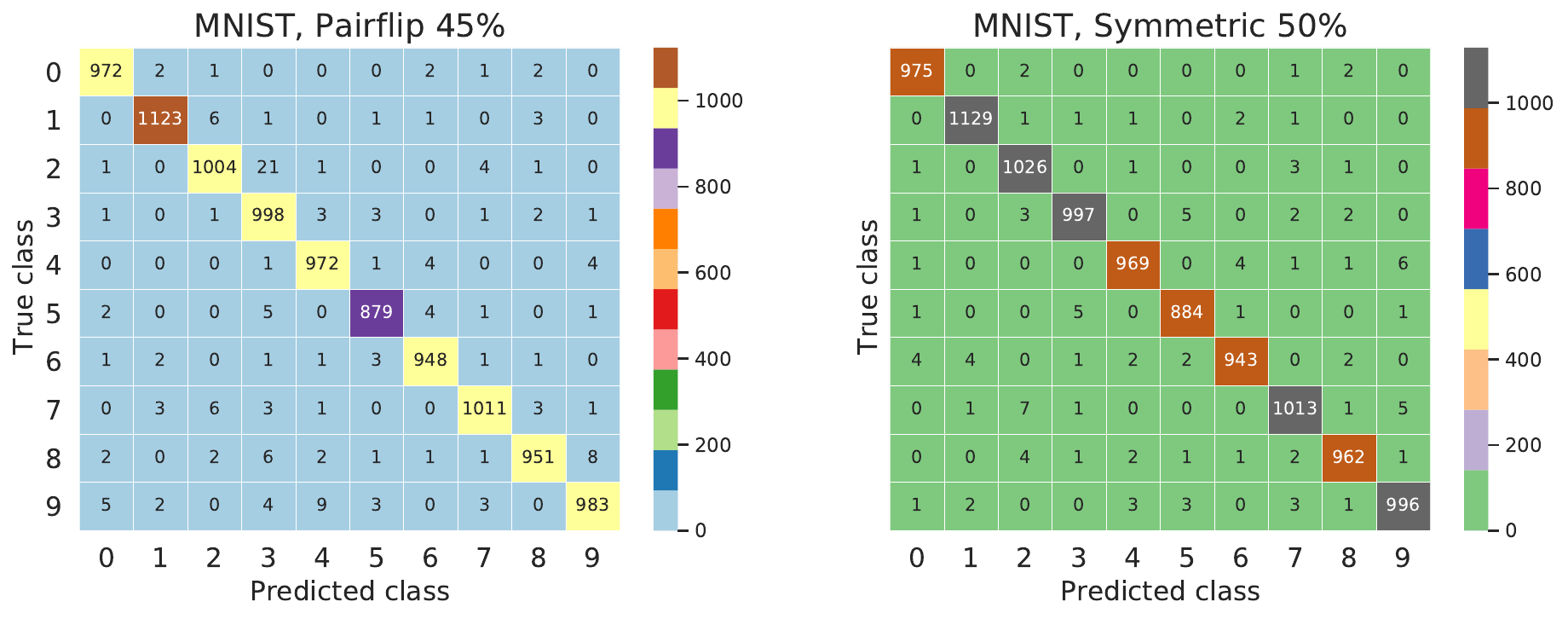}
        \footnotesize (a)
  \end{minipage}
     
     \begin{minipage}{1.0\textwidth}
      \centering
        \includegraphics[width=1.0\textwidth]{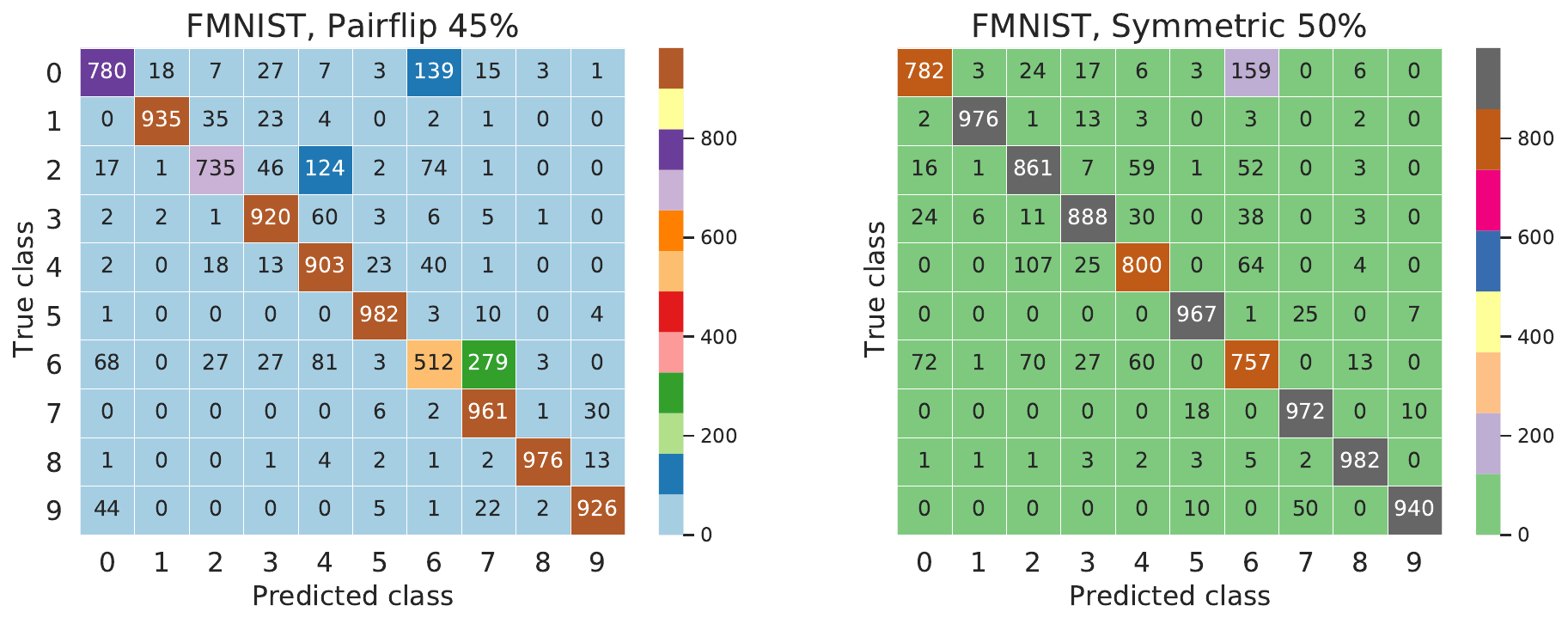}
        \footnotesize (b)
     \end{minipage}
   
     \begin{minipage}{1.0\textwidth}
      \centering
        \includegraphics[width=1.0\textwidth]{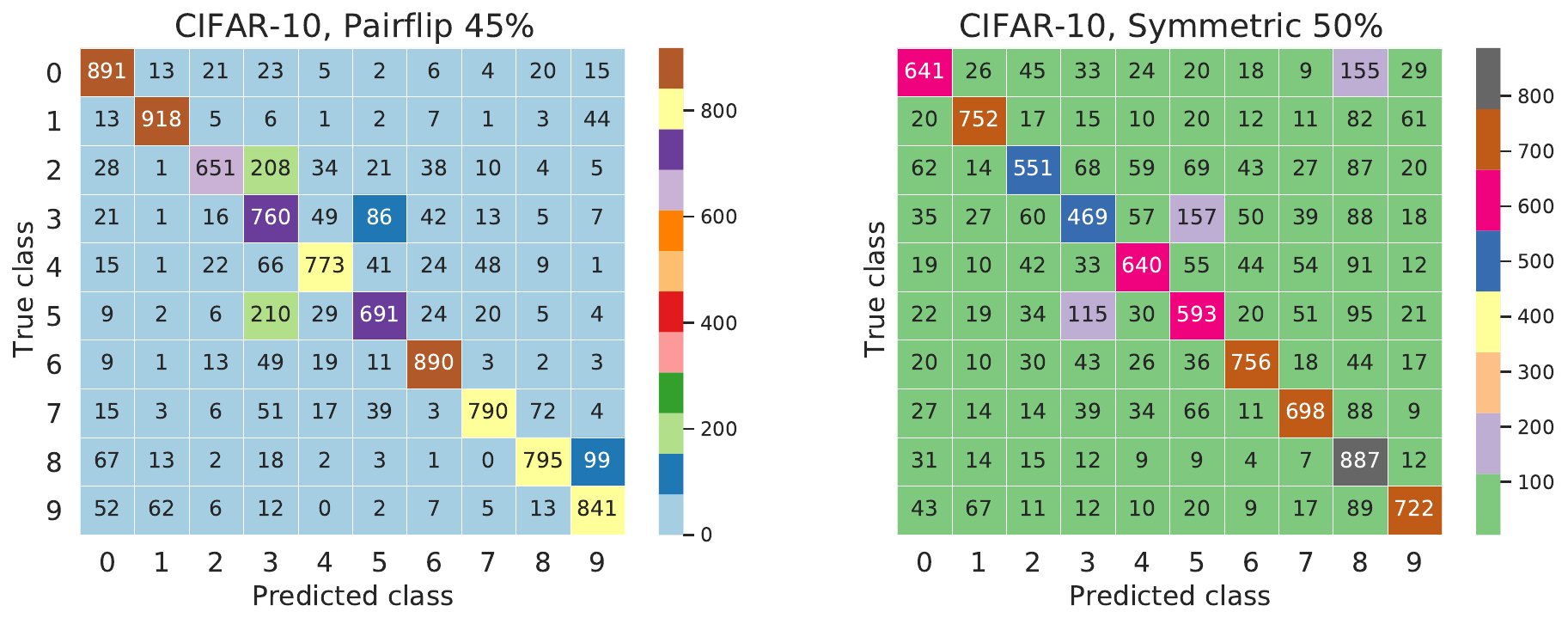}
        \footnotesize (c)
     \end{minipage}
  \end{minipage}
  \vskip10pt
  \caption{Confusion matrices of true class and predicted class for our algorithm for CIFAR-10, MNIST and Fashion-MNIST datasets.}
  \label{fig:confusion matrices}
  \end{figure*}

  In addition, Figure~\ref{fig:confusion matrices} highlights the confusion matrices of the true class and the predicted class by the classifier network of our algorithm. We show the confusion matrices plots for two extreme noise types pairflip 45$\%$ and symmetric 50$\%$ for all the datasets used. It is clearly seen that the confusion matrices are diagonally dominant thus highlighting the robust performance of our method. 

  \paragraph{CIFAR-10N.} We kept the same backbone architecture and fine-tune settings as CIFAR-10 in Section~\ref{sec:CIFAR10_fine_tune}.

\subsubsection{Additional experimental results}
\label{app:additional_classification}
  In our earlier experiments, we kept the same type of noise and noise levels across all the number of annotators in the annotator network. This is usually not representative of the noise in the real-world data, as it is possible that each annotator would be independent in the way it is confused about labeling and annotating the data (subject to their own biases). Therefore, we confuse each annotator with different types and levels of noise. Table~\ref{tab:diff_anno} shows the test accuracy of the classifier network on CIFAR-10, Fashion-MNIST, and MNIST datasets for different types of noise for each annotator. We achieved comparable results with an accuracy of 84.12$\%$, 91.12$\%$, and 98.97$\%$ for CIFAR-10, Fashion-MNIST, and MNIST respectively. Notably, the accuracy of the classifier network remains at par even with using high-level noise, such as pairflip 45$\%$ and symmetric 50$\%$ for two of the three annotators. 

  \begin{table*}
  \caption{Test accuracy (\%) with three different annotators (Annotator1: Pairflip 45\%, Annotator2: Symmetric 20\%, Annotator3: Symmetric 50\%) representing different noise types and noise levels on CIFAR-10, Fashion-MNIST and MNIST datasets.}\label{tab:diff_anno}
  \centering
    \begin{tabular}{c c c}
    \toprule
      CIFAR-10 & Fashion-MNIST & MNIST  \\     
      \midrule
      84.12 &  91.62 &  98.97\\
      $\pm{0.34}$  & $\pm{0.23}$ & $\pm{0.02}$ \\
    \bottomrule 
    \end{tabular} 
  \end{table*}

  \subsubsection{Curated MNIST Dataset} \label{sec:curated_appendix}
  In addition to symmetric and pairflip noises described in Section~\ref{sec:datasets_classification}, we also consider asymmetric and pairflip with permutation noises. In the latter, the ordered label categories were first permuted randomly and labels of two adjacent categories after permutation were swapped based on a preset ratio. Asymmetric noise is a block matrix transformation, where a portion of original labels are retained and the remainder is uniformly reassigned to the closest four categories. 
  
\begin{figure*} 
  \centering
  \small 
  \resizebox{.9\textwidth}{!}{
    \begin{tabular}{c c c}
      \toprule

      Original & Thin & Thick \\
      \midrule

      \includegraphics[width=0.25\textwidth]{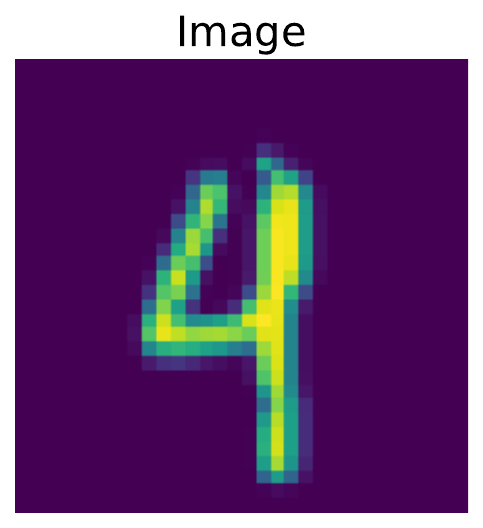} &  
      \includegraphics[width=0.25\textwidth]{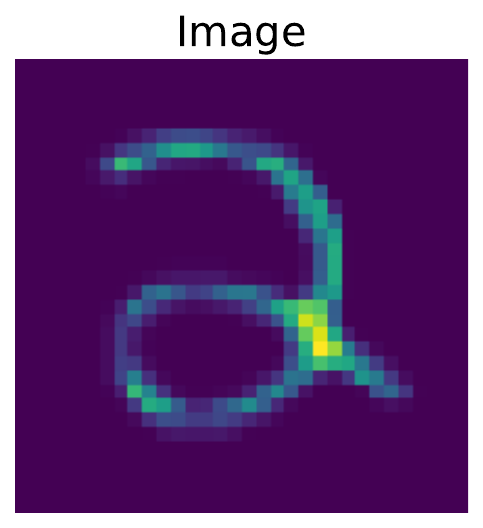}  & 
      \includegraphics[width=0.25\textwidth]{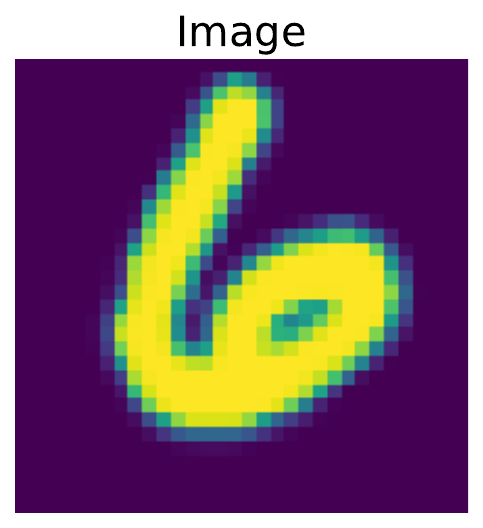}  \\ \hdashline
    
      \includegraphics[width=0.3\textwidth]{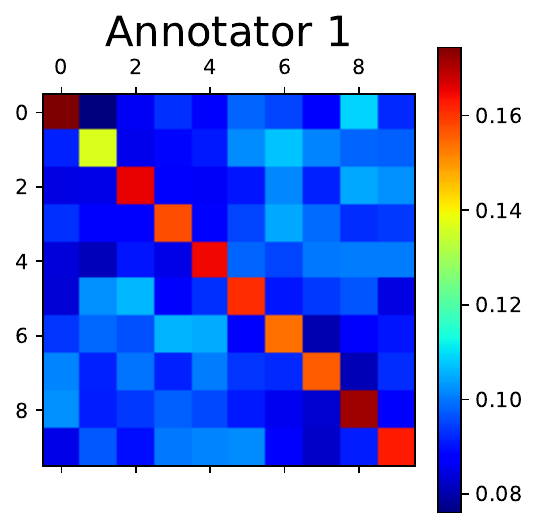} & \includegraphics[width=0.3\textwidth]{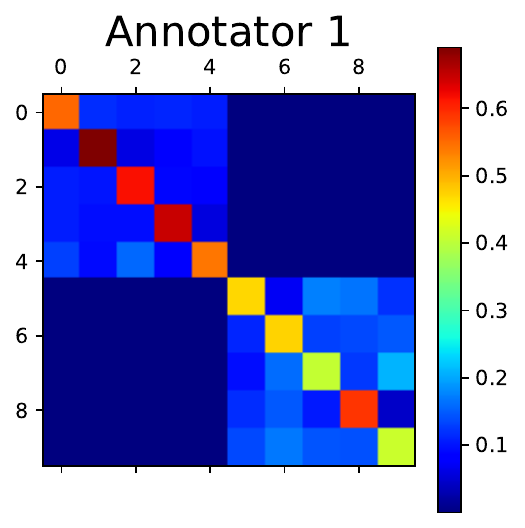} & \includegraphics[width=0.3\textwidth]{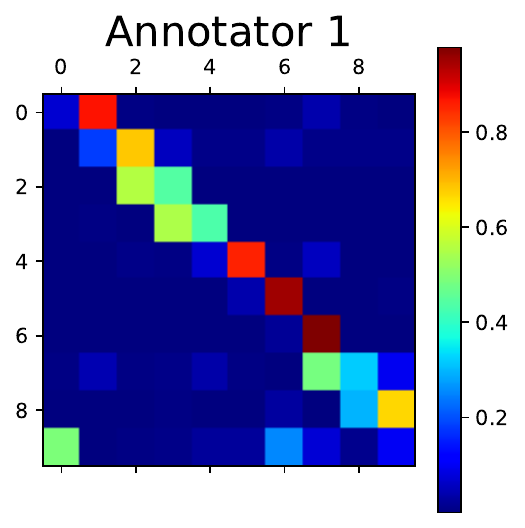}  \\ \hdashline
    
      \includegraphics[width=0.3\textwidth]{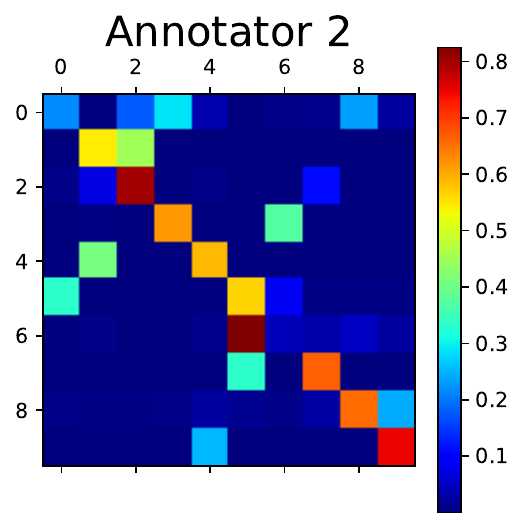}  & \includegraphics[width=0.3\textwidth]{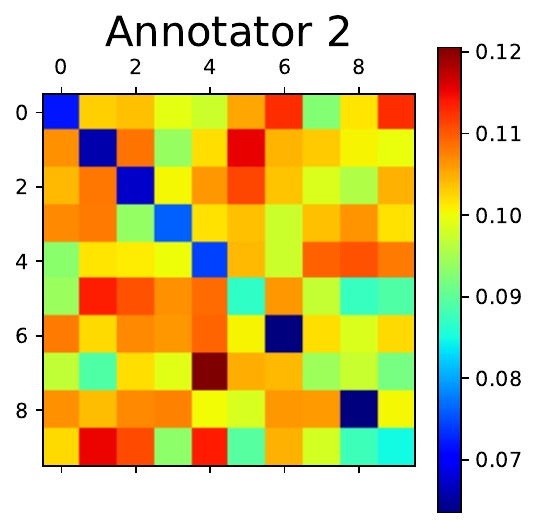} & \includegraphics[width=0.3\textwidth]{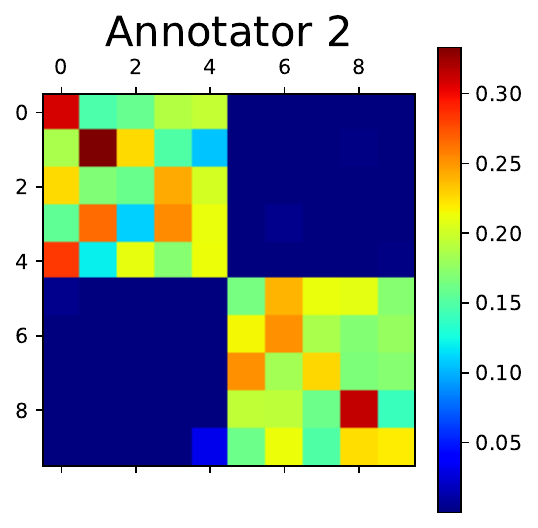}  \\ \hdashline
    
      \includegraphics[width=0.3\textwidth]{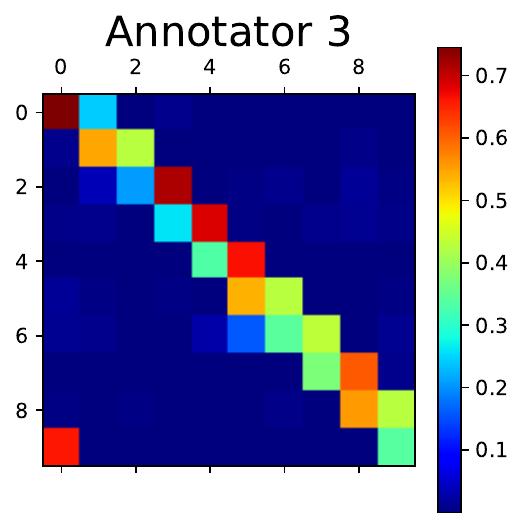}  & \includegraphics[width=0.3\textwidth]{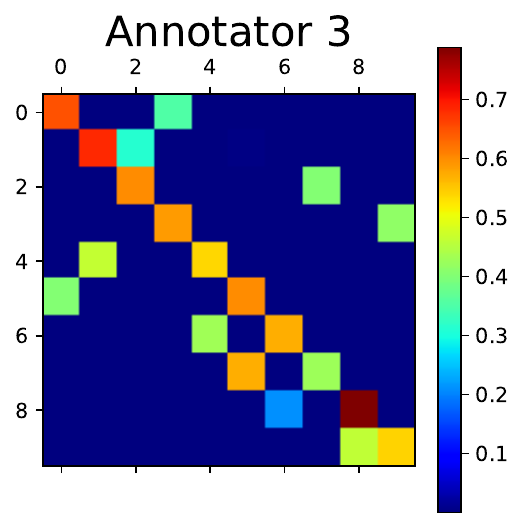} & \includegraphics[width=0.3\textwidth]{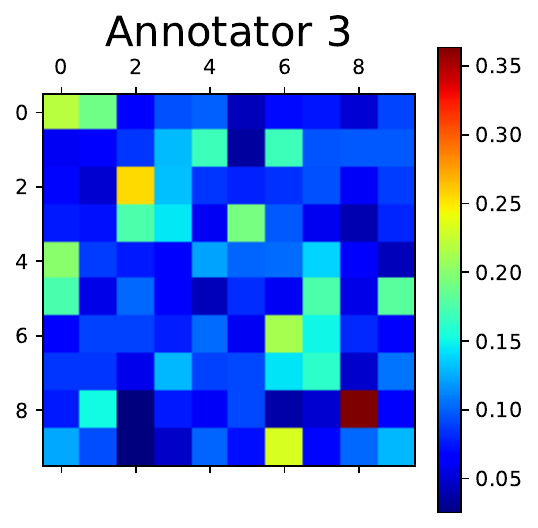}  \\ 
    \hline
    \end{tabular}
    }
    \vspace*{-2pt}
  \caption{Learned Annotators' confusion for different image styles using our approach with the confidence regularizer ($\lambda$=0.01, m=2) on  \textsl{MNIST} dataset.}\label{tab:SMs}
  \end{figure*}

\begin{figure*} 
  \centering
  \small 
    \begin{tabular}{c c c c}
      \toprule
      Image & Ground Truth & Our ($\lambda$=0.01, m=2) & Our ($\lambda$=0) \\
      \midrule
      \includegraphics[width=0.2\textwidth]{figures/Aya_Image_Original.pdf} & \includegraphics[width=0.23\textwidth]{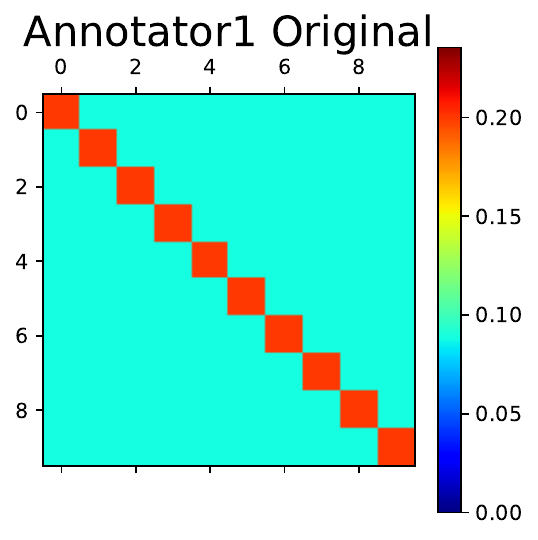} &  \includegraphics[width=0.23\textwidth]{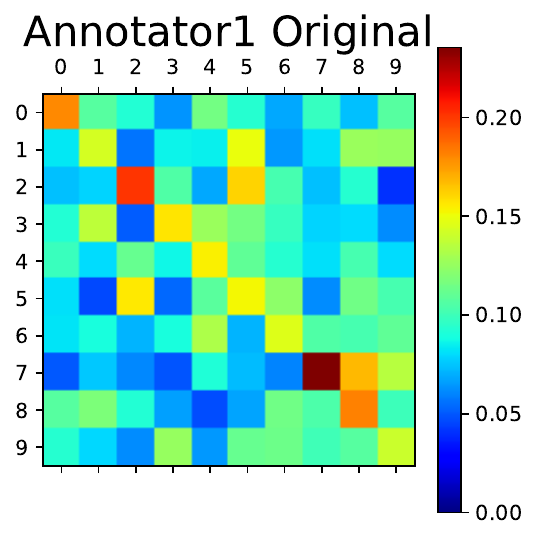}  & \includegraphics[width=0.23\textwidth]{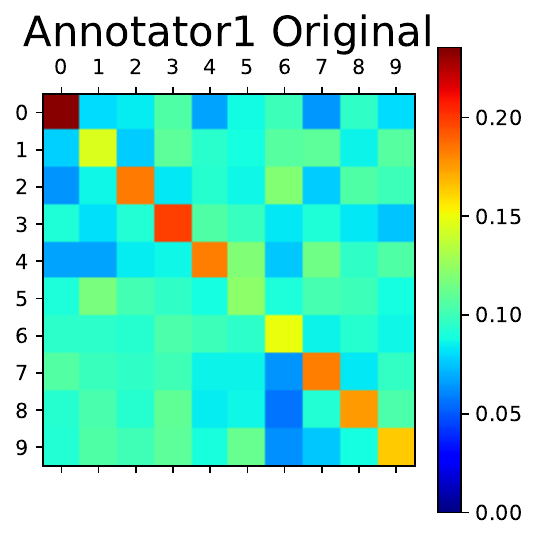}  \\ \hdashline
    
      \includegraphics[width=0.2\textwidth]{figures/Aya_Image_Thick.pdf} & \includegraphics[width=0.23\textwidth]{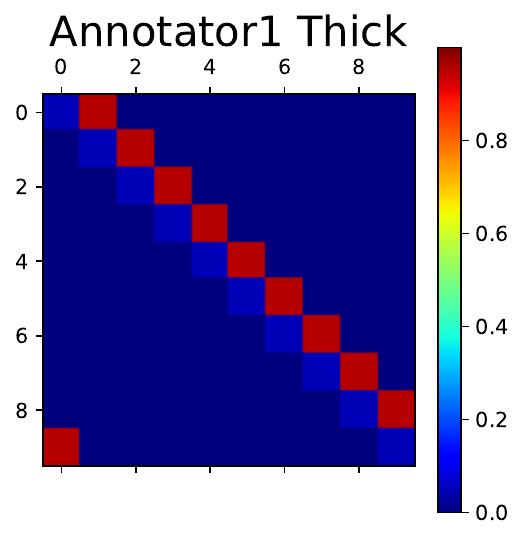} &  \includegraphics[width=0.23\textwidth]{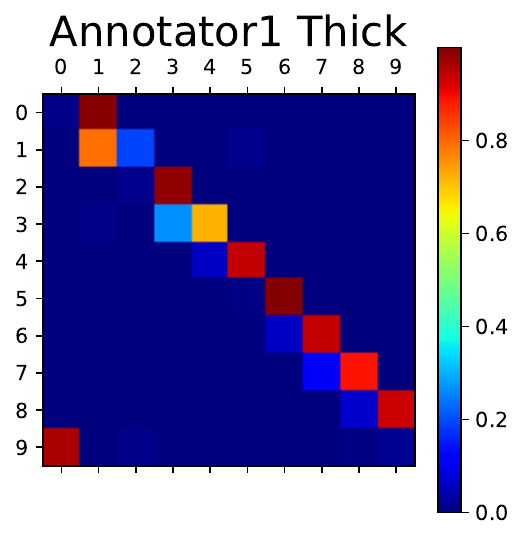}  & \includegraphics[width=0.23\textwidth]{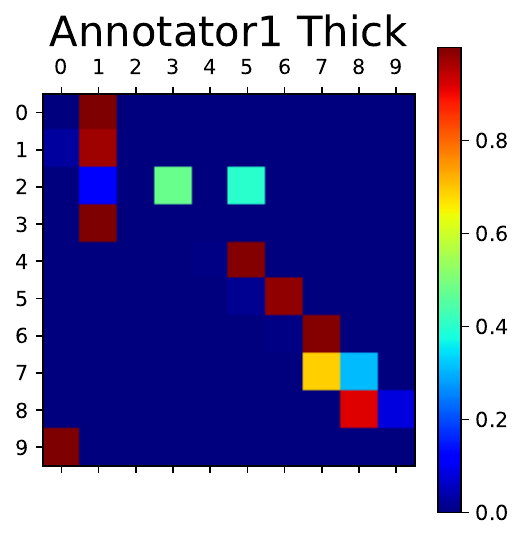}  \\ \hdashline
    
      \includegraphics[width=0.2\textwidth]{figures/Aya_Image_Thin.pdf} & \includegraphics[width=0.23\textwidth]{figures/Aya_SM_ann1_thin.pdf} &  \includegraphics[width=0.23\textwidth]{figures/Aya_annotator1_Thin_with_Reg.pdf}  & \includegraphics[width=0.23\textwidth]{figures/Aya_annotator1_Thin_without_Reg.pdf}  \\ 
      \hline
    \end{tabular}
  \caption{Ground truth and predicted confusion for Annotator 1 using different models: our approach with confidence regularizer ($\lambda=0.01$, m=2) and without it ($\lambda=0$) on  \textsl{Curated MNIST}  dataset.}\label{tab:SMs1}
  \end{figure*}

\begin{figure*} 
  \centering
  \small 
    \begin{tabular}{c c c c}
      \toprule
      Image & Ground Truth & Our ($\lambda$=0.01, m=2) & Our ($\lambda$=0) \\
      \midrule
      \includegraphics[width=0.2\textwidth]{figures/Aya_Image_Original.pdf} & \includegraphics[width=0.23\textwidth]{figures/Aya_SM_ann2_gt.pdf} &  \includegraphics[width=0.23\textwidth]{figures/Aya_annotator2_Original_with_Reg.pdf}  & \includegraphics[width=0.23\textwidth]{figures/Aya_annotator2_Original_without_Reg.pdf}  \\ \hdashline
    
      \includegraphics[width=0.2\textwidth]{figures/Aya_Image_Thick.pdf} & \includegraphics[width=0.23\textwidth]{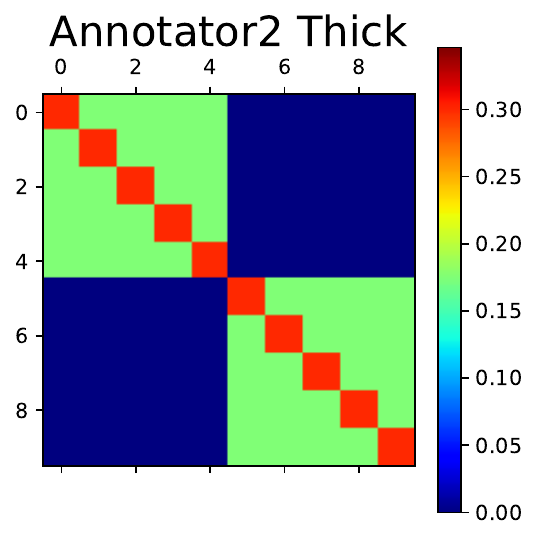} &  \includegraphics[width=0.23\textwidth]{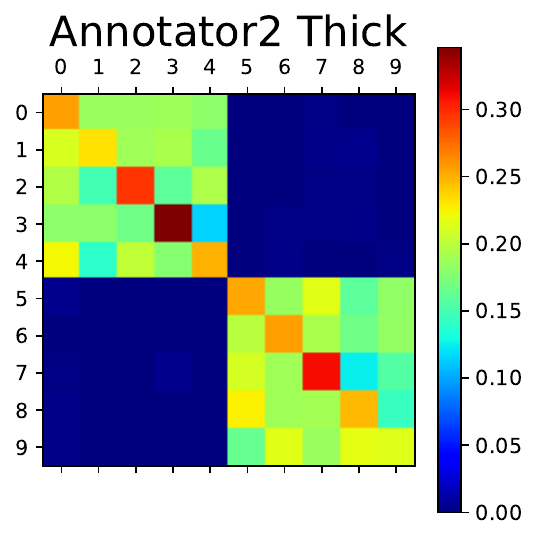}  & \includegraphics[width=0.23\textwidth]{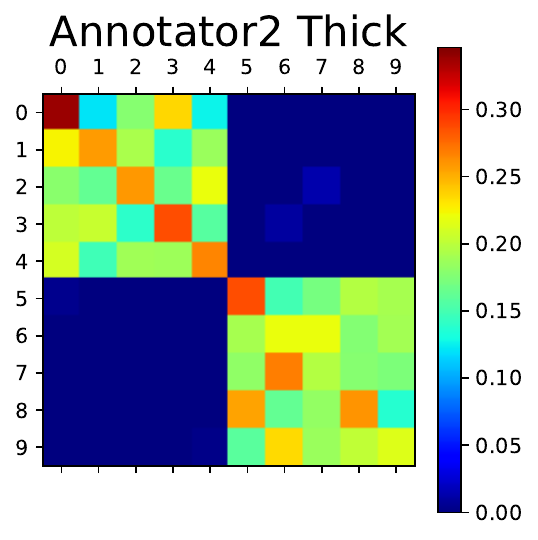}  \\ \hdashline
    
      \includegraphics[width=0.2\textwidth]{figures/Aya_Image_Thin.pdf} & \includegraphics[width=0.23\textwidth]{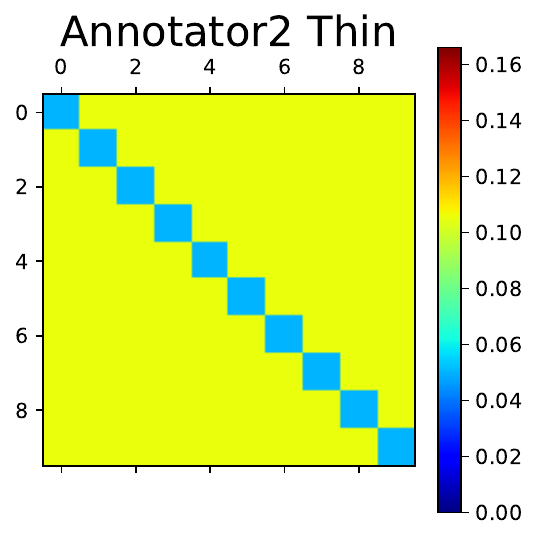} &  \includegraphics[width=0.23\textwidth]{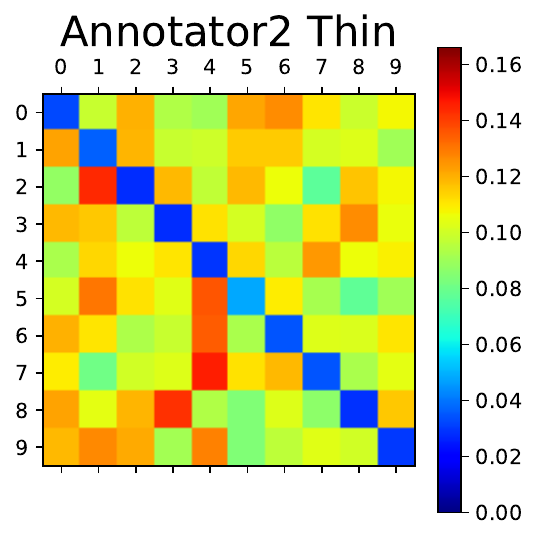}  & \includegraphics[width=0.23\textwidth]{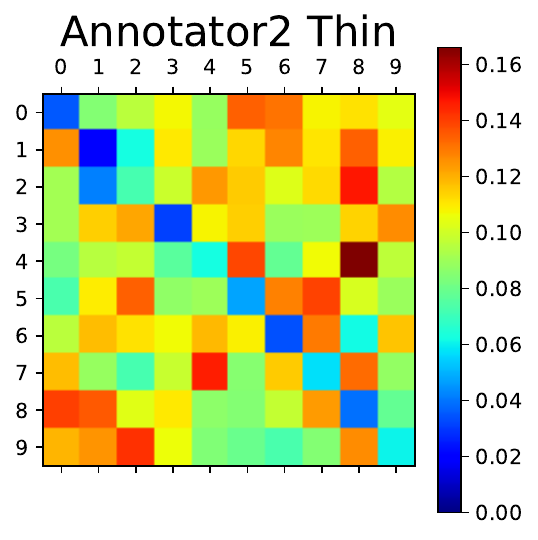}  \\ 
      \hline
    \end{tabular}
  \caption{Grount truth and predicted confusion for Annotator 2 using different models: our approach with confidence regularizer ($\lambda=0.01$, m=2) and without it ($\lambda=0$) on  \textsl{Curated MNIST} dataset.}\label{tab:SMs2}
  \end{figure*}

\begin{figure*} 
  \centering
  \small 
    \begin{tabular}{c c c c}
      \toprule
      Image & Ground Truth & Our ($\lambda$=0.01, m=2) & Our ($\lambda$=0) \\
      \midrule
      \includegraphics[width=0.2\textwidth]{figures/Aya_Image_Original.pdf} & \includegraphics[width=0.23\textwidth]{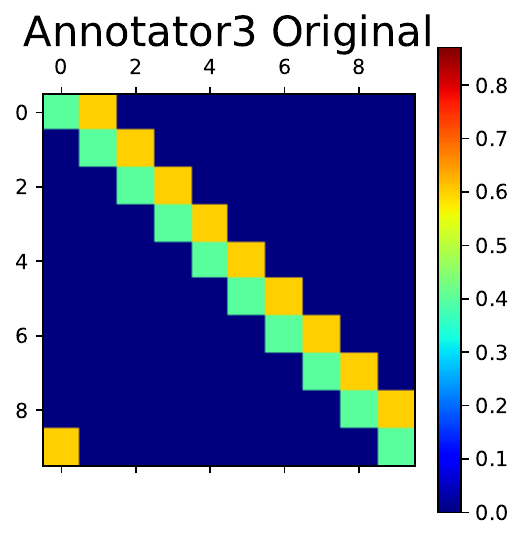} &  \includegraphics[width=0.23\textwidth]{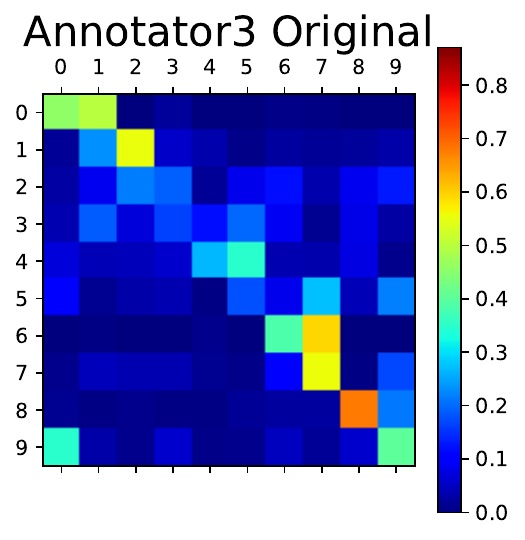}  & \includegraphics[width=0.23\textwidth]{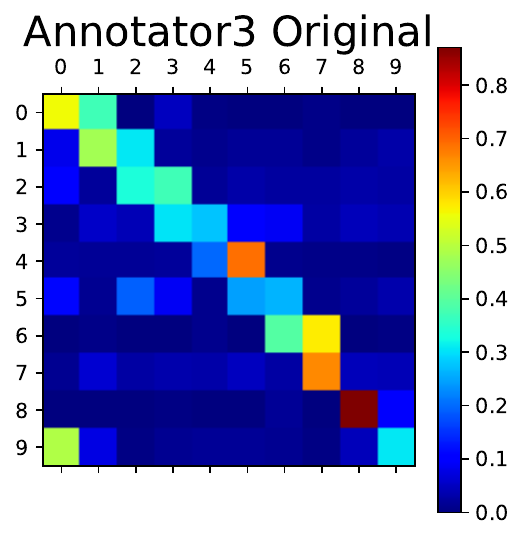}  \\ \hdashline
    
      \includegraphics[width=0.2\textwidth]{figures/Aya_Image_Thick.pdf} & \includegraphics[width=0.23\textwidth]{figures/Aya_SM_ann3_thick.pdf} &  \includegraphics[width=0.23\textwidth]{figures/Aya_annotator3_Thick_with_Reg.pdf}  & \includegraphics[width=0.23\textwidth]{figures/Aya_annotator3_Thick_without_Reg.pdf}  \\ \hdashline
    
      \includegraphics[width=0.2\textwidth]{figures/Aya_Image_Thin.pdf} & \includegraphics[width=0.23\textwidth]{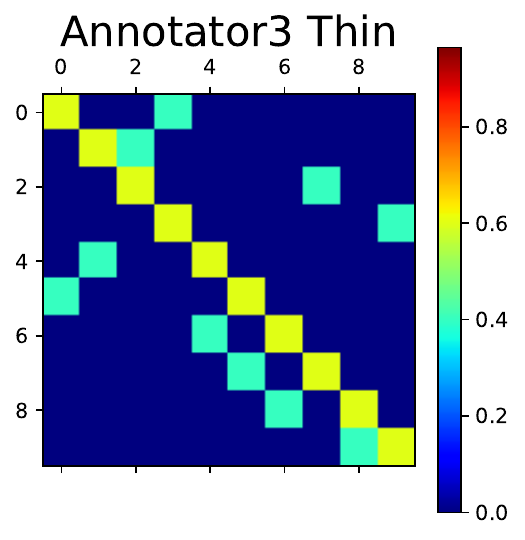} &  \includegraphics[width=0.23\textwidth]{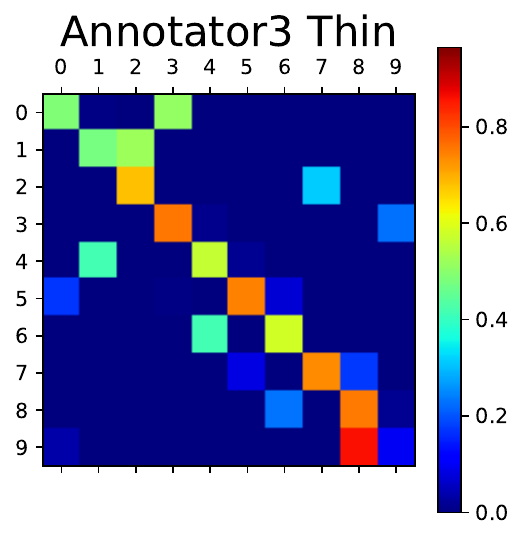}  & \includegraphics[width=0.23\textwidth]{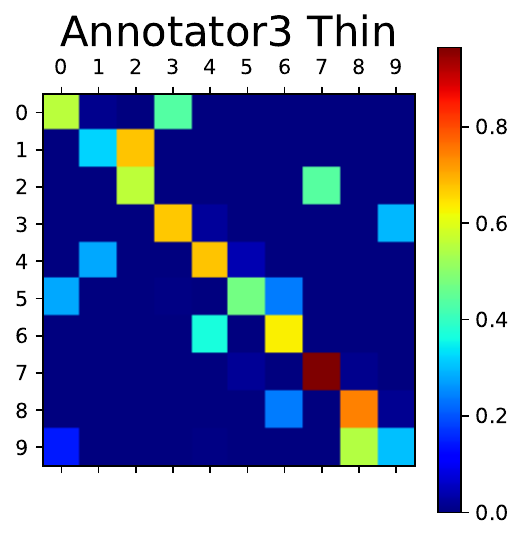}  \\
      \hline
    \end{tabular}
  \caption{Ground truth and predicted confusion for Annotator 3 using different models: our approach with confidence regularizer ($\lambda=0.01$, m=2) and without it ($\lambda=0$) on \textsl{Curated MNIST} dataset.}\label{tab:SMs3}
  \end{figure*}

  In Figure~\ref{tab:SMs} we demonstrate annotators' confusion using our algorithm on the Curated MNIST dataset that showcases different image styles of Original, Thin, and Thick. The strength of the confidence regularizer, $\lambda$=0.01, is increased by the multiplicative scalar m=2 every epoch. Figures~\ref{tab:SMs1}, \ref{tab:SMs2} and~\ref{tab:SMs3} highlight the original and predicted confusion of annotator 1, annotator 2, and annotator 3 using our approach with the regularizer and the non-regularized approach (that is, when $\lambda$=0).

\end{document}